\documentclass[lettersize,journal]{IEEEtran}
\usepackage{amsmath,amsfonts}
\usepackage{algorithmic}
\usepackage{algorithm}
\usepackage{array}
\usepackage{subcaption}
\usepackage[caption=false,font=normalsize,labelfont=sf,textfont=sf]{subfig}
\usepackage{textcomp}
\usepackage{stfloats}
\usepackage{url}
\usepackage{verbatim}
\usepackage{graphicx}
\usepackage{cite}
\usepackage{amssymb}
\usepackage{booktabs}
\usepackage{multirow}
\usepackage{subfig}
\usepackage{tablefootnote}
\usepackage[dvipsnames]{xcolor}
\newtheorem{theorem}{Theorem}

\newtheorem{definition}{Definition}

\usepackage{tabularx}
\usepackage{booktabs}
\usepackage{tikz}
\usetikzlibrary{tikzmark, positioning, arrows.meta}

\usepackage[capitalize]{cleveref}
\crefname{section}{Sec.}{Secs.}
\Crefname{section}{Section}{Sections}
\Crefname{table}{Table}{Tables}
\crefname{table}{Tab.}{Tabs.}

\begin{document}

\title{NTK-Guided Few-Shot Class Incremental Learning}

\author{Jingren~Liu, Zhong~Ji, \textit{Senior Member, IEEE}, Yanwei~Pang, \textit{Senior Member, IEEE}, YunLong~Yu
\thanks{This work was supported by the National Key Research and Development Program of China (Grant No. 2022ZD0160403), and the National Natural Science Foundation of China (NSFC) under Grant 62176178 (Corresponding author: Zhong~Ji).}
\thanks{Zhong Ji and Yanwei Pang are with the School of Electrical and Information Engineering, Tianjin Key Laboratory of Brain-Inspired Intelligence Technology, Tianjin University, Tianjin 300072, China, and also with the Shanghai Artificial Intelligence Laboratory, Shanghai 200232, China (e-mail: jizhong@tju.edu.cn; pyw@tju.edu.cn).}
\thanks{Jingren Liu is with the School of Electrical and Information Engineering, Tianjin Key Laboratory of Brain-Inspired Intelligence Technology, Tianjin University, Tianjin 300072, China (e-mail: jrl0219@tju.edu.cn).}
\thanks{YunLong Yu is with the College of Information Science and Electronic Engineering, Zhejiang University, Hangzhou, 310027, China. (e-mail: yuyunlong@zju.edu.cn).}
}

\markboth{Journal of \LaTeX\ Class Files,~Vol.~14, No.~8, August~2021}%
{Shell \MakeLowercase{\textit{et al.}}: A Sample Article Using IEEEtran.cls for IEEE Journals}


\maketitle

\begin{abstract} 
The proliferation of Few-Shot Class Incremental Learning (FSCIL) methodologies has highlighted the critical challenge of maintaining robust anti-amnesia capabilities in FSCIL learners. In this paper, we present a novel conceptualization of anti-amnesia in terms of mathematical generalization, leveraging the Neural Tangent Kernel (NTK) perspective. Our method focuses on two key aspects: ensuring optimal NTK convergence and minimizing NTK-related generalization loss, which serve as the theoretical foundation for cross-task generalization.
To achieve global NTK convergence, we introduce a principled meta-learning mechanism that guides optimization within an expanded network architecture. Concurrently, to reduce the NTK-related generalization loss, we systematically optimize its constituent factors. Specifically, we initiate self-supervised pre-training on the base session to enhance NTK-related generalization potential. These self-supervised weights are then carefully refined through curricular alignment, followed by the application of dual NTK regularization tailored specifically for both convolutional and linear layers.
Through the combined effects of these measures, our network acquires robust NTK properties, ensuring optimal convergence and stability of the NTK matrix and minimizing the NTK-related generalization loss, significantly enhancing its theoretical generalization. On popular FSCIL benchmark datasets, our NTK-FSCIL surpasses contemporary state-of-the-art approaches, elevating end-session accuracy by 2.9\% to 9.3\%.
\end{abstract}

\section{Introduction}
\label{sec:intro}
Few-shot class-incremental learning (FSCIL) extends class-incremental learning (CIL) by facilitating the continuous learning of incremental classes with only a few data samples. Drawing inspiration from established CIL approaches \cite{rebuffi2017icarl,yan2021dynamically,douillard2022dytox}, contemporary FSCIL methods \cite{zhang2021few,zhou2022few,yoon2023soft,yang2023neural} primarily emphasize safeguarding the pre-established knowledge in every session, specifically addressing the catastrophic forgetting problem. In contrast to CIL, which undergoes training in every session, the FSCIL learner is exclusively trained in the base session. Parameter updates for the FSCIL learner do not occur during subsequent incremental sessions. Consequently, the effectiveness of the subsequent incremental sessions relies heavily on the generalization capabilities developed during the training of the base session. Despite its significance, the emphasis on model generalization has been somewhat overlooked in existing FSCIL literature. To address this gap, we introduce the concept of the ``Neural Tangent Kernel (NTK)" and highlight its pivotal role in understanding the generalization properties of neural networks in the FSCIL context.

\begin{definition}[Neural Tangent Kernel]\label{NTK_Definition}
For a typical ConvNet with randomly initialized parameters $\theta_0$, NTK is given by $\Phi_{0}(x, x') = \left\langle \frac{\partial f(\theta_{0}, x)}{\partial \theta_{0}}, \frac{\partial f(\theta_{0}, x')}{\partial \theta_{0}}\right\rangle$, and it converges to a fixed kernel as the width $l$ approaches infinity
and the different inputs $x$ and $x'$ undergo constant alteration,
\begin{equation}\label{NTK_Defination}
    \begin{aligned}
        \Phi(\cdot) = \lim_{l\rightarrow \infty} \Phi_{0}(\cdot),
    \end{aligned}
\end{equation}
where $f$ denotes the neural network, $\frac{\partial f(\theta, \cdot)}{\partial \theta}$ is the neural network jacobian \cite{novak2022fast}, $\Phi(\cdot)$ signifies the fully converged NTK matrix and $l$ indicates the model's width. 
\end{definition}

A fundamental insight from the NTK corpus posits that \emph{\textbf{increasing the network's width tends to facilitate the NTK's convergence to a stable matrix during optimization, thereby bolstering generalization capabilities}}. This phenomenon suggests that post-optimization, the outputs of a network with infinite width become robust to variations in parameters and inputs, effectively capturing the inherent structures of the input data to produce reliable and consistent outputs for both familiar and novel data. Such a mechanism significantly mitigates the risk of overfitting, enhancing the model's aptitude for generalizing to unseen data and advancing it towards an ideal state of generalization.

As emphasized in \cref{NTK_Definition}, it's noted that incrementally broadening the network width simplifies the path to NTK convergence, diminishing the reliance on manual tuning during the NTK optimization process. However, it's imperative to acknowledge that even before achieving a theoretically infinite width, manual interventions may be necessary, often incurring considerable computational demands as delineated in preceding research. For instance, the work by \cite{Cagnetta2022WhatCB} underscores the computational intensity of computing the NTK for randomized data through ConvNets, requiring up to 200 gigabytes of GPU memory, thus posing a substantial challenge for a wide array of applications. Moreover, while the concept of an infinitely wide network is conceptually appealing, its practical application remains constrained by real-world limitations.

Therefore, incorporating NTK into FSCIL presents the challenge of ensuring that a finite-width network exhibits NTK properties akin to those of an infinitely wide network, even with limited optimization resources. Ideally, an infinitely wide network would demonstrate both excellent NTK convergence \cite{jacot2018neural,wang2022global,caron2023over} and a generalization loss approaching zero after optimization \cite{du2018gradient,wang2022global,zou2018stochastic,allen2019convergence}. Our goal is to replicate these properties within the constraints of a finite-width network.

Guided by these insights, our work focuses on addressing two critical issues—\textbf{NTK Convergence} and \textbf{NTK Generalization}—and seamlessly integrating them within the FSCIL framework. These methods are approached through the lens of the equivalence between meta-learning optimization and NTK convergence from \cref{thm:ml_equates_ntk} and \cref{thm:global-convergence}, as well as through the NTK-related generalization loss in \cref{Task_Specific_Generalization}.

To achieve NTK convergence in our expanded network during optimization, we theoretically align NTK convergence with meta-learning optimization. This alignment is based on the understanding that both meta-loss and meta-outputs are intrinsically linked to the NTK matrix and labels. Ensuring effective meta-learning performance in FSCIL is pivotal for attaining optimal NTK convergence. Consequently, we propose a bespoke meta-learning strategy for the FSCIL task, aiming to ensure that initiating meta-training without supervised pre-training leads to robust foundational performance.

Simultaneously, to ensure optimal NTK dynamics and minimal generalization loss, we incorporate self-supervised pre-training weights, curricular alignment, and dual NTK regularization, in line with \cite{wei2022more} and the FSCIL guidelines. These strategies are chosen for their effectiveness in \cref{NTK_Defination} and \cref{Task_Specific_Generalization}. In the regularization segment, to reduce computational overhead, we decompose the ConvNet and apply spectral regularization in convolutional layers \cite{wei2022more}. For the additional linear layers, we directly compute the NTK matrix and impose a spectral constraint to its eigenvalues, compressing their distribution to render it more consolidated.

In a nutshell, our contributions are threefold:
\begin{itemize}
    \item To the best of our knowledge, our work is the first to integrate NTK with FSCIL, advancing model generalization through robust theoretical foundations.
    \item To establish foundational NTK convergence during FSCIL optimization, we theoretically align NTK convergence with meta-learning and propose a bespoke meta-learning schema specifically tailored for FSCIL, initializing meta-training without supervised pre-training.
    \item Furthermore, to harvest optimal NTK dynamic outputs and the smallest NTK-related generalization loss, drawing inspiration from \cite{richards2021stability, wei2022more, bombari2023stability, taheri2023generalization}, we employ self-supervised pre-training, curricular alignment, and dual NTK regularization to achieve state-of-the-art results both theoretically and empirically.
\end{itemize}

\section{Related Works}
\textbf{Neural Tangent Kernel.} Initially proposed in \cite{neal1996priors}, it reveals the relationship between the convergence of infinitely wide neural networks and model generalization. Building on this pioneering work, subsequent studies \cite{le2007continuous,hazan2015steps} extend these findings to finite-width neural networks. Recent works \cite{du2018gradient,wang2022global,zou2018stochastic,allen2019convergence} on highly over-parameterized neural networks show that gradient descent can achieve zero training error even for finite but sufficiently wide CNNs, revealing promising generalization potential. For example, \cite{du2018gradient,du2019width} elucidate the correlation between network width and generalization, demonstrating that over-parameterization and random initialization enable models to converge linearly to the global minimum. Bombari \textit{et al.} \cite{bombari2023stability} discuss model stability, feature alignment, and their relation to generalization error. Lee \textit{et al.} \cite{lee2020finite,novak2022fast} propose a fast method for quantifying finite-width neural tangent kernels, demonstrating that optimizing NTK can enhance model generalization. Moreover, \cite{xiang2022tkil} addresses class imbalance and knowledge retention in incremental learning by introducing a Gradients Tangent Kernel loss, based on NTK considerations. Diverging from these methods, our work seeks to further integrate FSCIL with NTK dynamics and improve generalization capabilities specifically quantified by NTK.

\textbf{Few-shot Class-Incremental Learning.} It is proposed to address CIL in few-shot context. In general, the current FSCIL approaches \cite{tao2020few,shi2021overcoming,hersche2022constrained} principally are alleviating catastrophic forgetting in incremental sessions. For example, TOPIC \cite{tao2020few} is the pioneer, which leverages neural gas structure to solidify knowledge. CEC \cite{zhang2021few} employs episodic training via a context-propagated graph model to enhance adaptability. Moreover, MCNet \cite{ji2023memorizing} introduces ensemble learning to complement different memorized knowledge. Building upon these foundations, subsequent methods \cite{yoon2023soft,yang2023neural,zhou2022forward,peng2022few} have advanced via the class extension techniques, which is to preemptively prevent forgetting in future classes. For example, considered from forward compatibility, Zhou \textit{et al.} \cite{zhou2022forward} assign virtual prototypes to preserve space for incremental classes. ALICE \cite{peng2022few} extends the virtual classes to limit and facilitates well-clustered features via sophisticate-designed loss. Compared to them, our NTK-FSCIL focuses more on the model generalization and alleviates catastrophic forgetting from the side.

\section{NTK Theoretical Foundation in FSCIL}
\subsection{FSCIL Problem Formulation}
In a dynamic training environment featuring $N$ supervised tasks, $\Upsilon_{t=1}^{N}$, the essence of FSCIL is the incremental learning of these tasks without overlap. Specifically, the class sets $\Upsilon_{i}$ and $\Upsilon_{j}$ are disjoint for all $i \neq j$. The base session (session 0) provides an extensive dataset, while subsequent incremental sessions offer only a limited number of samples. Each task $\Upsilon_{t}$ is configured in the \textit{n}-way \textit{m}-shot \textit{k}-query setting for meta-training, represented as:
$$\Upsilon_{t}=(\mathbf{X}_{i}, \mathbf{X}_{j}, \mathbf{Y}_{i}, \mathbf{Y}_{j}),$$
The pair $[\mathbf{X}_{i}, \mathbf{Y}_{i}] \in \mathbb R^{m\times h \times w \times c}, \mathbb R^{m\times p}$ signifies $m$ support samples along with their corresponding labels. Here, $h$, $w$ and $c$ denote the image's dimensional indices, while $p$ is the dimension of the one-hot encoded labels. Similarly, the pair $[\mathbf{X}_{j}, \mathbf{Y}_{j}] \in \mathbb R^{k \times h \times w \times c}, \mathbb R^{k \times p}$ represents $k$ query samples associated with their labels. Moreover, $[X^{t}_{i}, Y^{t}_{i}]$ and $[X^{t}_{j}, Y^{t}_{j}]$ are all support and query samples in task $t$.

\subsection{NTK Dynamics in FSCIL}
The NTK theory has become a potent analytical tool for elucidating the behavior of deep neural networks, especially in the context of their training dynamics and generalization capabilities \cite{jacot2018neural, lee2019wide}. As the width of a neural network approaches infinity, the training process under gradient descent becomes increasingly deterministic and can be described using kernel methods. This section delves into the theoretical underpinnings of NTK and its implications for FSCIL.

At the heart of NTK theory lies the analysis of neural network architectures. For convolutional networks, the forward pass can be described by the following equations:
\begin{equation}
    u_l = \frac{\sigma_w \ast (W_l \ast h_{l-1})}{\sqrt{Z_{l-1}}} + \sigma_b b_l, \quad h_l = \text{ReLU}(u_l),
\end{equation}
where \(W_l\) is a \(Z_l \times Z_{l-1}\) filter kernel, and \(b_l\) is a bias vector. Here, \(\ast\) denotes the convolution operation, and \(Z_l\) represents the number of filters in the \(l\)-th layer, which is a measure of the layer's width. The depth of the network is denoted by \(L\).

\textbf{\textit{Remark:}} While the analysis above focuses on convolutional architectures, it is essential to recognize that the NTK framework is broadly applicable across a wide range of architectural paradigms, including ResNets and Transformers \cite{yang2020tensor}. The primary differences among these architectures manifest in the structural nuances of their respective NTK matrices.

In the infinite-width regime, where \(Z_l \rightarrow \infty\) for all hidden layers, given a finite number of samples and network depth, gradient descent is theoretically guaranteed to converge to a global minimum \cite{jacot2018neural, lee2019wide, du2018gradient}. This property has profound implications for the analysis and design of neural networks \cite{arora2019fine, chizat2019lazy, daniely2017sgd, allen2019convergence, neyshabur2018towards}. In the context of ConvNets, the model width \(Z_l\) refers to the number of filters in each convolutional layer. A wider network corresponds to a larger number of filters, which increases the model's generalization capacity. Therefore, in FSCIL's ConvNet architectures, the width expansion can be realized through the modulation of the in\_planes and planes parameters within ConvBlock structures.

In the context of FSCIL, the NTK regime provides a predictive model for any input \(x\), which is expressed as:
\begin{equation}\label{NTK_Dynamics}
    f(x) = f_0^*(x) + \Phi(x, X)(\Phi(X, X)+\lambda I)^{-1}(Y - f_0^*(X)),
\end{equation}
where \(f_0^*(x)\) represents the model's output under initial parameterization, \(\Phi\) denotes the NTK matrix for the base session, \(x\) represents an individual sample, and \(X\) embodies the aggregated dataset specific to the base session.

The application of NTK theory to FSCIL offers several advantages. It provides a theoretical framework for understanding how neural networks generalize from a limited number of examples in new classes while maintaining performance on previously learned classes. Moreover, the NTK perspective allows researchers to analyze the trade-offs between model capacity, training dynamics, and generalization performance in the incremental learning setting. As the field of NTK theory continues to evolve, it promises to yield further insights into the behavior of neural networks in various learning paradigms, including FSCIL. Future research directions may include developing NTK-inspired regularization techniques \cite{baratin2021implicit,li2019exponential,novak2018bayesian}, designing architecture-aware optimization algorithms \cite{fort2019deep,wang2022global}, and leveraging NTK analysis for more efficient few-shot and incremental learning strategies \cite{wang2022global,canatar2021spectral,doan2021theoretical, chai2009generalization, canatar2021spectral, bennani2020generalisation}.

\subsection{Generalization and NTK Analysis}
To mathematically capture the essence of model generalization in FSCIL, we adapt \cref{thm:ml_equates_ntk} and \cref{thm:global-convergence} from \cite{wang2022global}, alongside the \cref{Task_Specific_Generalization} from \cite{canatar2021spectral}.

\begin{theorem}[NTK-related Meta-Learning Output]\label{thm:ml_equates_ntk}
    Assuming negligible learning rates $\eta$ and $\lambda$, and given a network width $l$ approaching infinity, the meta-outputs $F_t$ for inputs $\mathbf{X}_j \in \mathbf{X}_j^t$, in relation to the training pairs $(\mathbf{X}_i, \mathbf{Y}_i) \in (\mathbf{X}_i^t, \mathbf{Y}_i^t)$, are highly likely to converge to a state describable by the NTK after optimization:
    \begin{equation}
    \begin{split}
        & F_t(\mathbf{X}_j, \mathbf{X}_i, \mathbf{Y}_i) = G_\Phi^\tau(\mathbf{X}_j, \mathbf{X}_i, \mathbf{Y}_i)+ \Phi((\mathbf{X}_j, \mathbf{X}_i), \\
        & \quad \quad (\mathbf{X}_j^t, \mathbf{X}_i^t))T^\eta_\Phi(t) (\mathbf{Y}_j^t - G^\tau_\Phi(\mathbf{X}_j^t, \mathbf{X}_i^t, \mathbf{Y}_i^t)), 
    \end{split}
    \end{equation}
    \begin{equation}
        \widetilde{T}^{\lambda}_\Phi (\cdot,\tau) = \Phi(\cdot,\cdot)^{-1}(I-e^{-\lambda \Phi(\cdot,\cdot) \tau}),
    \end{equation}
    \begin{equation}
        G_\Phi^{\tau}(\mathbf{X}_j, \mathbf{X}_i, \mathbf{Y}_i) = \Phi(\mathbf{X}_j, \mathbf{X}_i) \widetilde{T}^\lambda_\Phi(\mathbf{X}_i, \tau)\mathbf{Y}_i,
    \end{equation}
\end{theorem}
where \(\Phi((\mathbf{X}_j, \mathbf{X}_i), (\mathbf{X}_j^t, \mathbf{X}_i^t))\) is a kernel function.

\begin{theorem}[NTK-related Meta-Learning Convergence]\label{thm:global-convergence}
Define NTK matrix $\Phi = \lim_{l\rightarrow \infty} \frac{1}{l}J(\theta_0)J(\theta_0)^T$ and learning rate $\eta_0=\frac{2}{\sigma_{min}(\Phi) + \sigma_{max}(\Phi)}$. In a randomly initialized ConvNet of progressive width $l$, for any infinitesimal positive $\delta > 0$, there exist positive real numbers R and $\lambda_{0}$ such that an upper bound on the training loss is assuredly maintained with a probability exceeding $1 - \delta$.
\begin{align}
    \mathcal L(\theta_t) &= \frac{1}{2}\|F_{\theta_t}(\mathbf{X}_{i}, \mathbf{X}_{j}, \mathbf{Y}_{i}) - \mathbf{Y}_{j} \|_2^2 \nonumber\\
    &\leq \left(1 - \frac {\eta_0 \sigma_{min}(\Phi)}{3}\right)^{2t} \frac{R^2}{2}.  
    \label{eq:convergence-loss}
\end{align}
\end{theorem}

\begin{theorem}[NTK-related Generalization] \label{Task_Specific_Generalization}
In the realm of FSCIL, for the base session, there exists an optimal function \(f^*(x)\), fundamentally influenced by the spectral characteristics of the NTK. The expected generalization loss \(L_D(f^*)\) is dictated by these spectral properties and can be quantified as:
{\small
\begin{equation}
    L_D(f^*) = \frac{1}{1-\varepsilon} \sum_{i=0}^\infty \lambda_i w_i^{*2} \left (\frac{\beta}{\beta+ N \lambda_i} \right)^2  + \frac{\varepsilon}{1-\varepsilon} \sigma^2,
\end{equation}
}
where \(\beta\) represents a numerically determined parameter, \(\varepsilon\) is a function of the eigenvalues \(\lambda_i\) of the NTK as well as the sample size \(N\), and \(w_i^*\) symbolize the optimal weights corresponding to the deconstructed NTK in RKHS. The \(\beta\) and \(\varepsilon\) satisfy the relations outlined in the subsequent equations:
{\small
\begin{align}
    \sum_{i=0}^\infty \frac{\lambda_i}{\beta+N \lambda_i} = 1, \qquad \sum_{i=0}^\infty  \frac{N \lambda_i^2}{(\beta+N \lambda_i)^2} = \varepsilon.
\end{align}
}
\end{theorem}

From the insights provided in \cref{NTK_Dynamics}, it becomes clear that the pivotal factors influencing the optimization process during the base session are the initial weights \(f_0^*(x)\) and the alignment of logits \(Y-f_0^*(X)\). To address these critical elements, we adopt self-supervised pre-training alongside curricular alignment as strategies for augmentation.

Ensuring optimal NTK convergence during the dynamic optimization phase necessitates establishing a mathematical equivalence between NTK convergence and meta-learning optimization within the context of FSCIL. Drawing upon \cref{thm:global-convergence}, we ascertain that the meta-outputs and meta-loss are significantly influenced by the eigenvalue distribution of the NTK. This realization underscores the importance of fostering meta-learning optimization to facilitate NTK convergence, as highlighted in seminal works by \cite{jacot2018neural,wang2022global,caron2023over}. Consequently, Subsec.~\ref{meta-learning} is dedicated to elucidating strategies through which the FSCIL learner can surmount prevailing obstacles and achieve convergence within the meta-learning paradigm.

To minimize the generalization loss towards zero within the Reproducing Kernel Hilbert Space (RKHS), it is recognized that an optimal generalization loss depends on both the sample quantity and a stable eigenvalue spectrum \cite{nguyen2021tight,murray2022characterizing}. The formula \(\sum_{i=0}^\infty \lambda_i w_i^{*2} \left (\frac{\beta}{\beta+ N \lambda_i} \right)^2\) underscores that an eigenvalue's \(\lambda_i\) impact on generalization loss is adjusted by its multiplication with the corresponding optimal weight \(w_i^*\). This setup implies that larger eigenvalues, when coupled with larger weights, can significantly contribute to increased loss, whereas the influence of smaller eigenvalues is comparably minimal. Achieving uniformity in the NTK's eigenvalues — thereby avoiding extremes — ensures a more equitable contribution across all eigenvalues, mitigating any disproportionate effects on the total generalization loss. This balance is crucial for overall loss reduction.

In general, enhancing the number of base class samples and carefully moderating NTK eigenvalues are pivotal strategies for lowering generalization loss. We adopt the strategy of enlarging virtual class samples alongside dual NTK regularization to achieve this goal. These strategic measures collectively fortify the FSCIL learner's generalization efficiency, securing top-tier performance both theoretically and practically during the base session optimization.

\section{Meta-Learning Convergence in FSCIL}\label{meta-learning}
As delineated in \cref{thm:ml_equates_ntk} and \cref{thm:global-convergence}, it is evident that the NTK governs the trajectory of meta-outputs during meta-training, with the NTK matrix's eigenvalue distribution shaping the meta-loss for backpropagation optimization. Thus, when optimizing meta-loss through gradient descent, the NTK is bound to achieve global convergence. At this point, NTK convergence is equated to meta-learning optimization.

Grounded in the established equivalence between NTK convergence and meta-learning optimization, we endeavor to steer FSCIL's meta-learning towards global optimal convergence, thereby ensuring the NTK's convergence to its optimal region. However, many existing FSCIL methods \cite{zhang2021few,zhou2022forward,zhou2022few,chi2022metafscil} argue that initiating meta-learning without supervised
pre-training often fails to converge to the global optimum, leading to suboptimal FSCIL performance. To find the reasons of this performance degradation, we conduct a comprehensive study of the engineering implementations of existing methods. From this study, we identify two primary causes: (1) the inflexibility in logit combinations, and (2) the oversight of scenarios similar to incremental sessions during training.

In response to the aforementioned two reasons, we reformulate the classification loss, dividing it into two distinct facets as depicted in Eq.~\ref{best_loss}. The first component integrates the mixup mechanism with a superior margin-based loss (elaborated in Subsec.~\ref{alignment}), addressing the rigid class-level combinations during meta-training. Simultaneously, the second component introduces an unsupervised few-shot loss to simulate testing operations and enhance adaptability to scenarios similar to incremental sessions during training.
\begin{equation}\label{best_loss}
\mathcal L_{cls} = 
\begin{tikzpicture}[remember picture, baseline=(ClassificationLoss.base)]
\node[draw=red, fill=red!40, thick, rectangle, inner sep=2pt] (ClassificationLoss) {$\mathcal{L}_{logits}$};
\end{tikzpicture}
+ \gamma \,
\begin{tikzpicture}[remember picture, baseline=(SmallSampleLoss.base)]
\node[draw=blue, fill=blue!40, thick, rectangle, inner sep=2pt] (SmallSampleLoss) {$\mathcal{L}_{embeddings}$};
\end{tikzpicture}
\end{equation}
\begin{tikzpicture}[remember picture, overlay]
\path[draw=red, line width=0.8pt, -{Latex[length=2mm]}] (ClassificationLoss.south) -- ++(0,-0.15cm) -- ++(-1.4cm,0) node[midway, below, font=\footnotesize] {Margin-based Loss};
\path[draw=blue, line width=0.8pt, -{Latex[length=2mm]}] (SmallSampleLoss.south) -- ++(0,-0.15cm) -- ++(1.4cm,0) node[midway, below, font=\footnotesize] {Adaptability Loss};
\end{tikzpicture}

To mitigate the constraints of rigid class-level combinations during meta-training, inspired by \cite{zhou2022forward, peng2022few}, we deploy a mixup strategy\footnote{See \cref{supp:mix_ups} for an in-depth examination of various mixup strategies and their impact on FSCIL performance.}. This mixup strategy, by virtually expanding class boundaries, facilitates the generation of highly diverse logits for virtual and real classes. Such an expansion is crucial for overcoming the rigid optimization inherent in scenarios with limited logit variability.

In prior meta-learning paradigms \cite{zhang2021few,chi2022metafscil}, supervised pre-training and meta-learning refinement are typically performed across the entire class domain, generating associated logits. However, these methods often overlook the embedding-based classification method specific to each meta-task subspace, designed to simulate testing operations in incremental sessions. To address this, we deploy an unsupervised few-shot loss to better simulate and adapt to incremental session scenarios. Diverging from traditional methods, we reconfigure class indices in each batch to define sub-domain labels $\mathbf{y^{*}}$, and utilize embeddings-derived prototypes to generate pseudo-labels $\mathbf{pl}$.
\begin{equation}\label{pesudo_labels}
   \mathbf{pl} = \frac{{\mathbf{X}_j \cdot \Omega_{n}(\frac{1}{m} \sum_{i=1}^{m} \mathbf{X}_i)}}      {{\|\mathbf{X}_j\|_2 \cdot \|\Omega_{n}(\frac{1}{m} \sum_{i=1}^{m} \mathbf{X}_i)\|_2}},
\end{equation}
where $n$ and $m$ denote their quantities in the $n$-way, $m$-shot scenario, and $\Omega_{n}$ is the operation converging $n$ prototypes.

After generating pseudo-labels in the sub-domain, we employ the classic cross-entropy loss to formulate the adaptability loss, denoted as $\mathcal{L}_{embeddings} = CE(\mathbf{pl}, \mathbf{y}^{*})$. 

Ultimately, by employing the mixup strategy to diversify class-level logit combinations and incorporating an adaptability loss to simulate incremental scenarios, our bespoke meta-learning strategy can directly start meta-training without supervised
pre-training, outperforming the conventional multi-step paradigm in performance and efficiency. Additionally, assisted by meta-learning, our NTK matrix theoretically reaches the global optimal convergence.

\section{Reducing Generalization Loss in FSCIL}\label{ntk_stability}
Drawing inspiration from \cref{Task_Specific_Generalization} and \cite{wei2019regularization, wei2022more, bombari2023stability, taheri2023generalization}, we consider NTK-related dynamics and generalization loss. In this literature, a more stable eigenvalue distribution of NTK matrix indicates a better generalization state, emphasizing the importance of real-time monitoring and constraining of the NTK's eigenvalue distribution. Moreover, following the ideas from \cref{NTK_Dynamics} and \cite{wei2022more, bombari2023stability}, we understand that utilizing pre-trained models to initialize the model weights, combined with better logit-label alignment, will effectively optimize the NTK dynamic outputs, enhancing the model generalization capabilities. Thus, in this section, our exploration will focus on three facets: \textbf{initialization}, \textbf{alignment} and \textbf{regularization}.

\begin{figure}[t]
\centering
\includegraphics[width=0.4\textwidth]{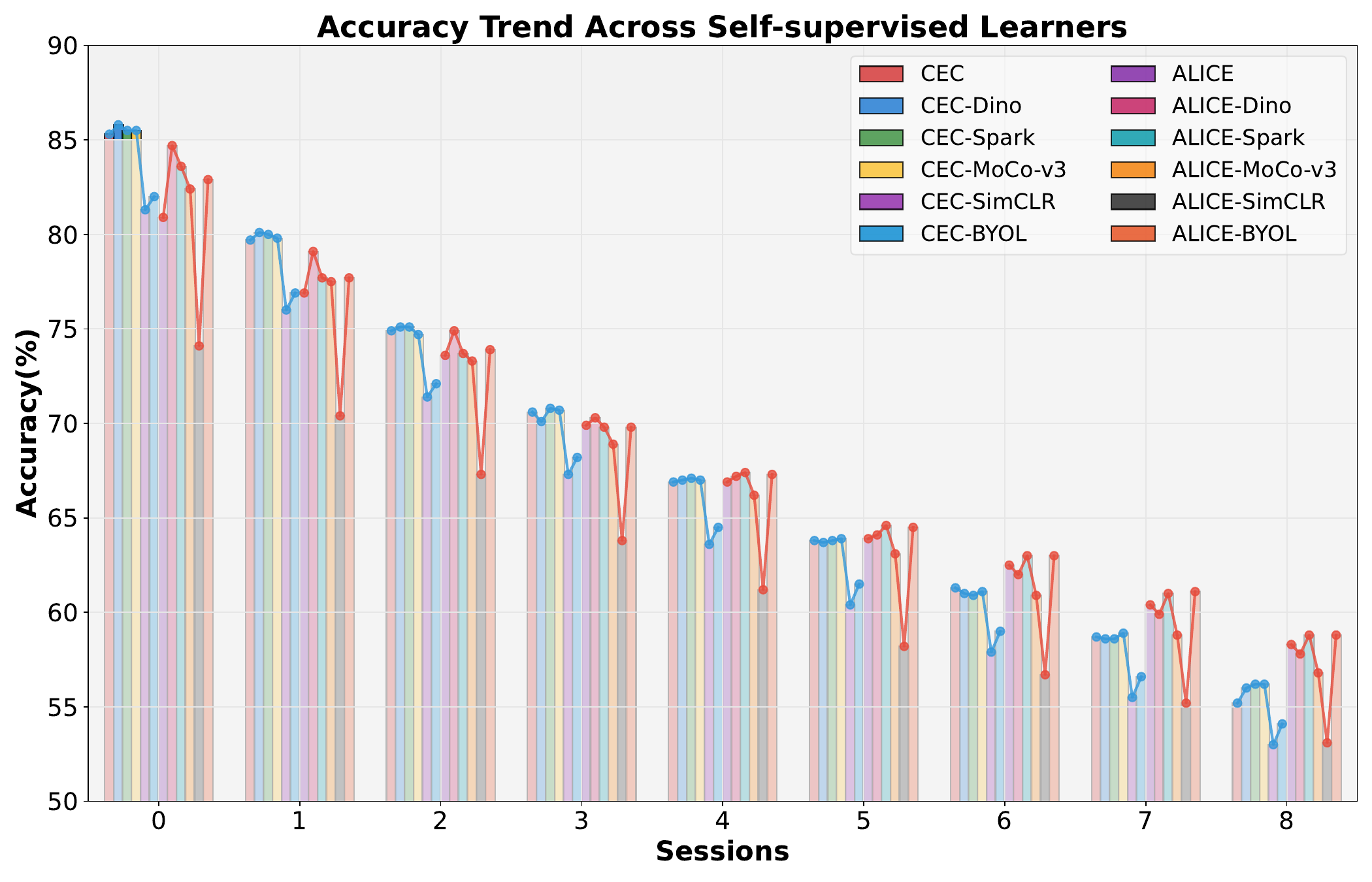}
\caption{The difference in FSCIL performance amongst various self-supervised learners, utilizing ResNet-18$\times$2 on CIFAR100.}
\label{fig:self-supervised}
\vspace{-3ex}
\end{figure}

\subsection{Initialize Weights: Self-Supervised Pre-Training}
\label{pretrain}
As highlighted in \cref{NTK_Dynamics} and \cite{wei2022more}, efficient pre-training methods significantly enhance model generalization \cite{liu2023few} and stabilize eigenvalue decay in NTK matrix. Yet, in FSCIL, using larger datasets for pre-training is often discouraged due to the risk of excessively enhancing the FSCIL learner's generalization capabilities, which could potentially overshadow the challenges inherent in continual and few-shot scenarios. Adhering to FSCIL guidelines, we opt for self-supervised pre-training on the base session for effective initialization. To further investigate the impact of various self-supervised methods on FSCIL, we conduct experiments using two mainstream frameworks, CEC \cite{zhang2021few} and ALICE \cite{peng2022few}, and apply weights trained on the base session from self-supervised methods such as SwAV \cite{caron2020unsupervised}, Dino \cite{caron2021emerging}, SimCLR \cite{chen2020simple}, SimSiam \cite{chen2021exploring}, MAE \cite{he2022masked}, MoCo-v3 \cite{chen2021mocov3}, BYOL \cite{grill2020bootstrap} and SparK \cite{tian2023designing}. The details and results are meticulously presented in \cref{supp:self-supervised pre-training} and \cref{fig:self-supervised}.

As evidenced in \cref{fig:self-supervised}, the generative SparK framework \cite{tian2023designing} outperforms its contrastive learning-based counterparts. When the FSCIL learner is initialized with SparK's weights, we observe notable improvements in incremental session accuracies, alongside maintained base session accuracy, which is indicative of enhanced generalization potential. Furthermore, as detailed in \cref{supp:NTK_properties}, the NTK's eigenvalue distribution with SparK initialization demonstrates greater stability. Therefore, due to its excellent performance in FSCIL task and NTK properties, SparK emerges as our preferred choice.

Through further analysis of existing self-supervised methods, we can better understand why Spark excels as a self-supervised pre-training method in FSCIL. Spark employs sparse and hierarchical mask modeling for self-supervised learning in convolutional networks, effectively capturing local and global features and enhancing adaptability to new categories. Compared to methods like SimCLR \cite{chen2020simple} and BYOL \cite{grill2020bootstrap}, Spark demonstrates greater robustness and generalization in feature representation. SimCLR relies on extensive data augmentation, while BYOL improves performance through network interactions. However, these methods often face challenges with feature drift and memory interference when introducing new categories. In contrast, Spark effectively filters irrelevant features while preserving original information, resulting in superior feature extraction and adaptability. Its hierarchical masking strategy enables progressive learning of more abstract features, further improving FSCIL performance.

\subsection{Better Alignment: Margin-based Loss}\label{alignment}
In \cref{meta-learning}, our mixup strategy addresses the rigidity in class-level combinations, fostering enriched interactions for the FSCIL learner across various virtual classes, thereby increasing the number of base class samples \(N\) and reducing the generalization loss \(L_D(f^*)\) as described in \cref{Task_Specific_Generalization}. Yet, this strategy also introduces certain side effects, namely, making traditional cross-entropy loss less effective in scenarios with a large number of classes \cite{zhou2022forward, peng2022few}. However, as highlighted in \cref{NTK_Dynamics}, alignment plays a crucial role in NTK dynamics and model generalization. Hence, to achieve more effective logit-label alignment, we replace cross-entropy loss with a margin-based loss \cite{peng2022few, zou2022marginbased}. The fundamental structure of our margin-based softmax loss is outlined as follows:
\begin{equation}
\label{eq:general_mining_softmax}
{
\mathcal{L}_{logits}=-\log\frac{e^{sP(\cos\theta_{y_i})}}
{e^{s{P(\cos\theta_{y_i})}}+\sum^{n}_{j\neq y_{i}}
{e^{sN(t,\cos\theta_{j})}}},
}
\end{equation}
Here, $\theta_{y_{i}}$ and $\theta_j$ represent the cosine similarities for the $i$-th and $j$-th class prototypes. $P(\cdot)$ and $N(\cdot)$ adjust positive and negative cosine similarities, respectively.

Furthermore, since the mixup mechanism's inherent randomness in creating virtual classes, it can easily lead to an imbalance between virtual and real classes, which presents a significant challenge for margin-based loss. To counteract this, we adopt curriculum learning concepts \cite{huang2020curricularface}, which involve adaptive alterations to $P(\cdot)$ and $N(\cdot)$, as follows:
\begin{equation}
\label{eq:curriculum_1}
N(\ast) =
\begin{cases}
\cos \theta_j, & \mbox{$\mathbf{\varepsilon} \geq 0$} \\
\cos \theta_j (t + \cos \theta_j), &\mbox{$\mathbf{\varepsilon}<0$},
\end{cases}
\end{equation}
\begin{equation}
\label{eq:curriculum_2}
P(\ast) =
\begin{cases}
\cos \theta_{y_i} \cos m -\cos^{2}\theta_{y_i} \sin m, & \mbox{$\mathbf{\varepsilon} \geq 0$} \\
\cos (\theta_{y_i} + m) + m \sin m, &\mbox{$\mathbf{\varepsilon}<0$},
\end{cases}
\end{equation}
where $\mathbf{\varepsilon} = \cos(\theta_i+m) - \cos\theta_j$ embodies the threshold, $t$ is the mean logits updated via EMA mechanism over each batch, and $m$ designates a predetermined, constant margin.

Through this refined design, training begins by deemphasizing negative samples, setting the initial value of $t$ to zero, while simultaneously accentuating positive samples through the incorporation of $m$. As training progresses and reaches the desired performance, indicated by $\mathbf{\varepsilon} \geq 0$, $t$ is incrementally increased to shift the model's focus towards hard samples in positive classes. When certain classes achieve the target performance, we activate a suppression mechanism for highly accurate positive samples, thereby redirecting focus to less accurate ones. This recalibration ensures balanced performance across all class domains.

Ultimately, this adaptive optimization not only addresses the class weight imbalance but also significantly optimizes NTK dynamic outputs, as detailed in \cref{supp:NTK_properties} and \ref{supp:classification_loss}.

\subsection{Refine NTK eigenvalues: Dual NTK Regularization} 
\label{regularization} 
After ensuring the NTK dynamics reach an optimal state, according to \cref{Task_Specific_Generalization}, we need to further refine the eigenvalue distribution to avoid excessively large or small NTK eigenvalues. However, as mentioned earlier, the computational burden associated with the NTK approximation in ConvNets challenges the limits of contemporary hardware. To address this, we adopt a compromise strategy, implementing dual NTK regularization that is specifically tailored to convolutional and linear layers. This strategy delicately refines the NTK matrix, both indirectly and directly, by intervening in and tightening the NTK's eigenvalue distribution. The details of the associated composite losses are provided below:
\begin{equation} \label{reg_loss}
\mathcal L_{reg} = 
\begin{tikzpicture}[remember picture, baseline=(ConvRegLoss.base)]
\node[draw=ForestGreen, fill=ForestGreen!40, thick, rectangle, inner sep=2pt] (ConvRegLoss) {$\mathcal{L}_{conv\_reg}$};
\end{tikzpicture}
+ 
\begin{tikzpicture}[remember picture, baseline=(LinearRegLoss.base)]
\node[draw=BurntOrange, fill=BurntOrange!40, thick, rectangle, inner sep=2pt] (LinearRegLoss) {$\mathcal{L}_{lin\_reg}$};
\end{tikzpicture}
\end{equation}
\begin{tikzpicture}[remember picture, overlay]
\path[draw=ForestGreen, line width=0.8pt, -{Latex[length=2mm]}] (ConvRegLoss.south) -- ++(0,-0.2cm) -- ++(-2.8cm,0) node[midway, below, font=\footnotesize] {Convolutional Spectral Regularization};
\path[draw=BurntOrange, line width=0.8pt, -{Latex[length=2mm]}] (LinearRegLoss.south) -- ++(0,-0.2cm) -- ++(1.4cm,0) node[midway, below, font=\footnotesize] {Linear NTK Regularization};
\end{tikzpicture}

To ensure a slower eigenvalue decay in convolutional layers, we implement spectral regularization on their weights \cite{wei2022more,wei2019regularization}. Furthermore, to reduce the computational overhead associated with large-scale weight calculations, we employ Singular Value Decomposition (SVD) to decompose the weights. The detailed formulation is as follows:
\begin{equation}
\label{eq:linear_ntk3}
\mathcal{L}_{conv\_reg} = \alpha \cdot \sum_{c \in f} \frac{\varphi_{\text{{max}}}^{c}}{\varphi_{\text{{min}}}^{c}},
\end{equation}
where $\alpha$ is a hyperparameter, $\varphi$ denotes the SVD function, and $c$ is each convolutional weight in the ConvNet.

Conversely, for the additional linear layers, simplification is not necessary. Inspired by self-supervised linear evaluation studies \cite{caron2021emerging, he2022masked, chen2021mocov3, tian2023designing}, we directly compute their NTK matrix and impose constraints on the eigenvalue distribution. Moreover, acknowledging the NTK's inherent property— the optimal NTK stabilizes into a constant matrix irrespective of inputs after optimization—we concentrate on refining the stability of eigenvalue fluctuations in each batch, aiming to maintain relative consistency during optimization. Consequently, we devise a regularization loss to control the variance of these NTK eigenvalues:
\begin{align}
\mathcal{L}_{lin\_reg} &= \beta \cdot \frac{\sigma_{max}(\Phi_{linear})-\sigma_{min}(\Phi_{linear})}{\sigma^{'}_{max}(\Phi_{linear})-\sigma^{'}_{min}(\Phi_{linear}) + \epsilon},
\end{align}
where \(\Phi_{linear}\) denotes the NTK matrix of the linear layer, \(\beta\) represents the hyperparameter governing the NTK eigenvalue fluctuation, while \(\sigma(\cdot)\) and \(\sigma(\cdot)^{'}\) refer to the eigenvalues of the current and preceding batches, respectively. Additionally, \(\epsilon\) is introduced as a perturbation term to avoid a zero denominator.

\textbf{Optimization}: Overall, our NTK-FSCIL adeptly navigates NTK convergence and NTK-related generalization loss through a three-stage process: SparK-based self-supervised pre-training on the base session, fully supervised FSCIL optimization on the base session and NCM testing on incremental sessions. During the crucial FSCIL optimization phase, the FSCIL learner is optimized through meta-learning, utilizing the sum of \cref{best_loss} and \cref{reg_loss}, ensuring optimal NTK convergence, NTK dynamics and minimal generalization loss.

\section{Experiments}
In the experimental section, we commence with a detailed overview of the utilized datasets, as delineated in \cref{datasets}. This foundational step sets the stage for a series of experiments conducted on the most prominent FSCIL methods, meticulously validating the rationale and efficacy of each component. These components include the impact of convolutional network width on FSCIL performance as explored in \cref{wider convnets}, the influence of self-supervised pre-training (i.e., the initialization \(f_0^*(x)\)) on FSCIL performance detailed in \cref{supp:self-supervised pre-training}, the effect of logit-label alignment (i.e., \(Y-f_0^*(X)\)) on FSCIL performance as discussed in \cref{supp:classification_loss}, and the impact of expanding virtual classes, corresponding to \(N\) in \cref{Task_Specific_Generalization}, on outcomes as examined in \cref{supp:mix_ups}.

Subsequent to these foundational experiments, we conduct comparative and analytical evaluations between our comprehensive approach and the latest methodologies in the field, supplemented by detailed ablation studies. This analytical journey culminates in a quantitative assessment of the eigenvalue distribution within the NTK matrix, alongside the visualization of embeddings through t-SNE plots, vividly illustrating NTK characteristics and model generalization capabilities.

\subsection{Experimental Setup} \label{datasets}
\textbf{Datasets and Splits:} Our method is evaluated on multiple datasets: CIFAR100 \cite{krizhevsky2009learning} (consisting of 60,000 images across 100 classes), CUB200-2011 \cite{wah2011caltech} (featuring 200 bird species classes), \textit{mini}ImageNet \cite{ILSVRC15} (a subset of ImageNet-1K with 100 classes at reduced resolution) and ImageNet100 \cite{wu2019large} (comprising 100 classes selected from ImageNet-1K). For CIFAR100, \textit{mini}ImageNet, and ImageNet100, we partition the classes into two sets: 60 base classes and 40 incremental classes. The latter are further divided into eight 5-way 5-shot tasks. In the case of CUB200, we have 100 base classes and 100 incremental classes, organized into ten 10-way 5-shot tasks. Our training splits align with those used in previous work \cite{tao2020few}, except for ImageNet100, which follows the splits introduced in \cite{zhou2022forward}. More detailed content is shown in \cref{supp:datasets}.

\begin{table}[!t]
    \centering
    \setlength{\tabcolsep}{1.6mm}
    \renewcommand{\arraystretch}{0.95}
    \footnotesize
    \resizebox{0.84\linewidth}{!}{
        \begin{tabular}{ccccccc}
            \toprule
            Dataset & $\mathcal{C}^{base}$ & $\mathcal{N}_{base}$ & $\mathcal{C}^{inc}$ & \#Inc. & Shots & Resolution \\
            \midrule
            CIFAR100~\cite{krizhevsky2009learning} & 60 & 30000 & 40 & 8 & 5 & 32×32 \\
            \emph{mini}ImageNet~\cite{ILSVRC15} & 60 & 30000 & 40 & 8 & 5 & 84×84 \\
            CUB200~\cite{wah2011caltech} & 100 & 3000 & 100 & 10 & 5 & 224×224 \\
            ImageNet100 \cite{wu2019large} & 60 & 77352 & 40 & 8 & 5 & 112×112 \\
            \bottomrule
        \end{tabular}
    }
\caption{Statistics of benchmarks datasets. $\mathcal{C}^{base}$: number of classes in base session. $\mathcal{C}^{inc}$: total number of classes in incremental sessions. \#Inc.: number of incremental sessions. Shots: training shots for incremental sessions. $\mathcal{N}_{base}$: number of samples in base session.}
\label{supp:datasets}
\end{table}

\begin{figure}[t]
\centering
\includegraphics[width=0.4\textwidth]{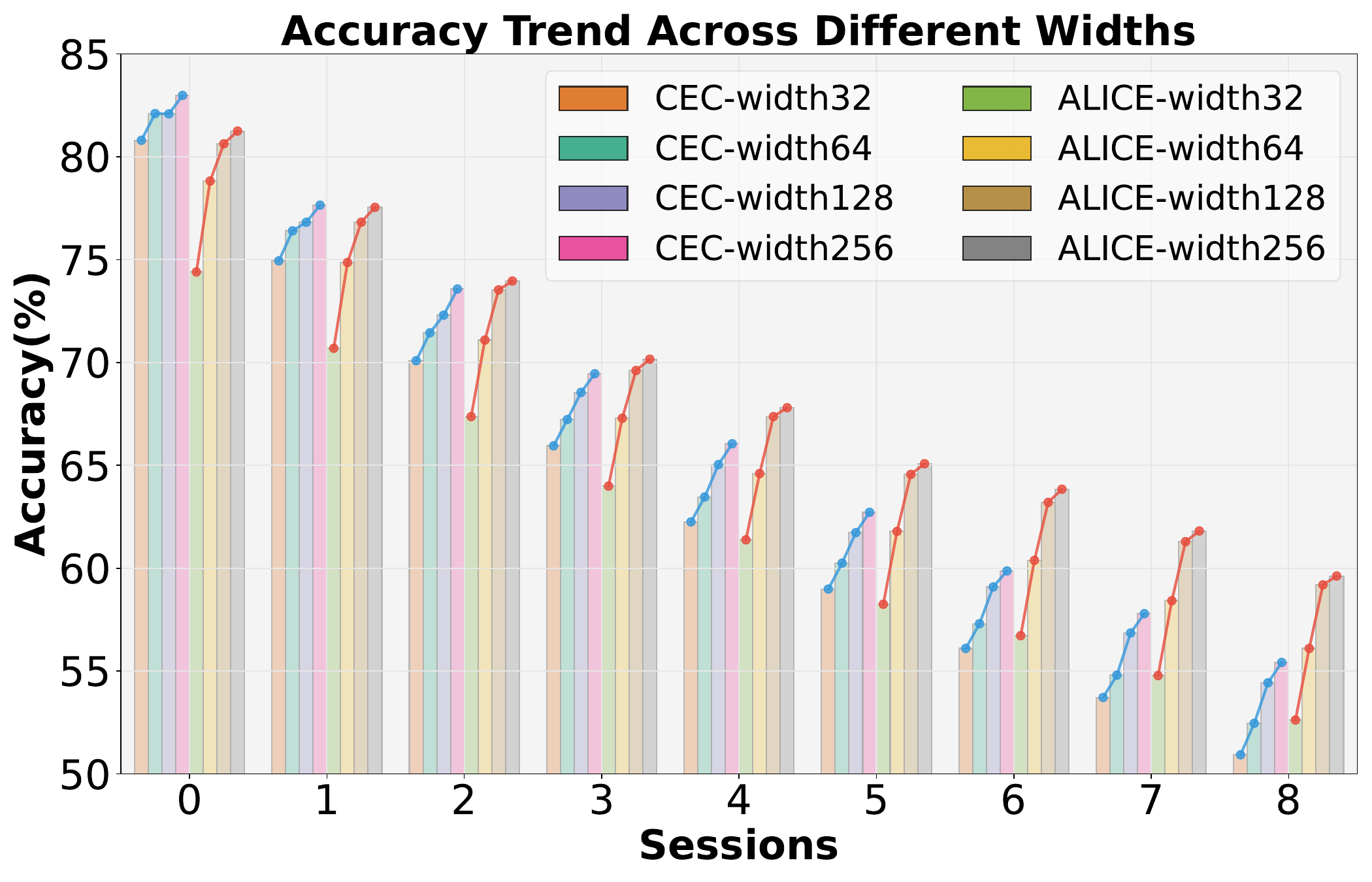}
\caption{The FSCIL performance on CIFAR100 across different widths in ResNet-18, employing the CEC \cite{zhang2021few} and ALICE \cite{peng2022few}.}
\label{fig:width_extension}
\vspace{-3ex}
\end{figure}

\subsection{Impacts of Network Width in FSCIL}\label{wider convnets}
In this subsection, we assess how network width influences FSCIL performance. Specifically, we select two methods, i.e., CEC \cite{zhang2021few} and ALICE \cite{peng2022few}, for evaluation by gradually widening the backbone. As shown in \cref{fig:width_extension}, an expansion in network width leads to a gradual enhancement in accuracy across sessions, providing additional evidence that widening the network improves its generalization. However, excessively increasing the width of the network poses significant challenges to optimization. Hence, unless stated otherwise, our default backbone is ResNet-18$\times$2 (`width128' in \cref{fig:width_extension}), with doubled convolutional block widths.

Next, we aim to clearly demonstrate the impact of varying the width of individual convolutional layer on FSCIL performance. Using a ConvNet with doubled width as our baseline, we experimentally modify the width of specific layers and observe the resulting changes in FSCIL performance. These experiments are conducted using various ResNet architectures on the CIFAR100 dataset to ensure efficient and informative outcomes. Table~\ref{supp:res18_layer_width} provides a detailed analysis of the effects of varying the width of each convolutional layer on FSCIL performance. The data suggests a direct correlation: \textbf{as the width of these layers increases, FSCIL efficacy consistently improves}. This enhancement is observed not only when the entire ConvNet is widened but also when individual convolutional layer is selectively widened.

\begin{table*}[htbp]
    \centering
    \setlength{\tabcolsep}{2.8mm}
    \renewcommand{\arraystretch}{1.0}
    \resizebox{0.8\linewidth}{!}{
        \begin{tabular}{cccccc}
            \toprule
            ResNet-18 & CEC & ALICE & ResNet-18 & CEC & ALICE \\
            \midrule
            32$\Rightarrow$256$\Rightarrow$512$\Rightarrow$1024 & 81.2$\rightarrow$53.9 & 79.4$\rightarrow$58.0 & 128$\Rightarrow$64$\Rightarrow$512$\Rightarrow$1024 & 81.1$\rightarrow$53.1 & 78.8$\rightarrow$57.3 \\ 
            64$\Rightarrow$256$\Rightarrow$512$\Rightarrow$1024 & 81.6$\rightarrow$53.6 & 80.0$\rightarrow$58.0 & 128$\Rightarrow$128$\Rightarrow$512$\Rightarrow$1024 & 81.8$\rightarrow$54.2 &  79.6$\rightarrow$58.2\\ 
            \textbf{128$\Rightarrow$256$\Rightarrow$512$\Rightarrow$1024} & \textbf{82.1$\rightarrow$54.4} & \textbf{80.6$\rightarrow$58.2} & \textbf{128$\Rightarrow$256$\Rightarrow$512$\Rightarrow$1024} & \textbf{82.1$\rightarrow$54.4} & \textbf{80.6$\rightarrow$58.2} \\ 
            256$\Rightarrow$256$\Rightarrow$512$\Rightarrow$1024 & 82.3$\rightarrow$54.7 & 80.7$\rightarrow$58.2 & 128$\Rightarrow$512$\Rightarrow$512$\Rightarrow$1024 & 83.0$\rightarrow$54.5 & 80.6$\rightarrow$58.6\\ 
            \toprule
            ResNet-18 & CEC & ALICE & ResNet-18 & CEC & ALICE \\
            \midrule
            128$\Rightarrow$256$\Rightarrow$128$\Rightarrow$1024 & 81.3$\rightarrow$52.7 & 78.8$\rightarrow$57.2 & 128$\Rightarrow$256$\Rightarrow$512$\Rightarrow$256 & 84.7$\rightarrow$54.9 & 79.7$\rightarrow$57.4 \\ 
            128$\Rightarrow$256$\Rightarrow$256$\Rightarrow$1024 & 81.6$\rightarrow$53.4 & 79.4$\rightarrow$57.7 & 128$\Rightarrow$256$\Rightarrow$512$\Rightarrow$512 & 83.8$\rightarrow$54.7 & 80.6$\rightarrow$58.1 \\ 
            \textbf{128$\Rightarrow$256$\Rightarrow$512$\Rightarrow$1024} & \textbf{82.1$\rightarrow$54.4} & \textbf{80.6$\rightarrow$58.2} &  \textbf{128$\Rightarrow$256$\Rightarrow$512$\Rightarrow$1024} & \textbf{82.1$\rightarrow$54.4} & \textbf{80.6$\rightarrow$58.2}  \\ 
            128$\Rightarrow$256$\Rightarrow$1024$\Rightarrow$1024 & 83.0$\rightarrow$54.9 & 80.4$\rightarrow$58.9 & 128$\Rightarrow$256$\Rightarrow$512$\Rightarrow$2048 & 80.6$\rightarrow$52.3 & 80.3$\rightarrow$58.7 \\
            \bottomrule
        \end{tabular}
        }
\caption{The FSCIL performance on \textbf{CIFAR100} using ResNet-18 with varying widths of convolutional layers is presented. The baseline is indicated in bold, and the performance is measured solely by the accuracies in the base and final sessions.}
\label{supp:res18_layer_width}
\end{table*}

\begin{figure*}
\centering
    \begin{subfigure}[b]{0.44\columnwidth}
        \centering
        \includegraphics[width=\columnwidth]{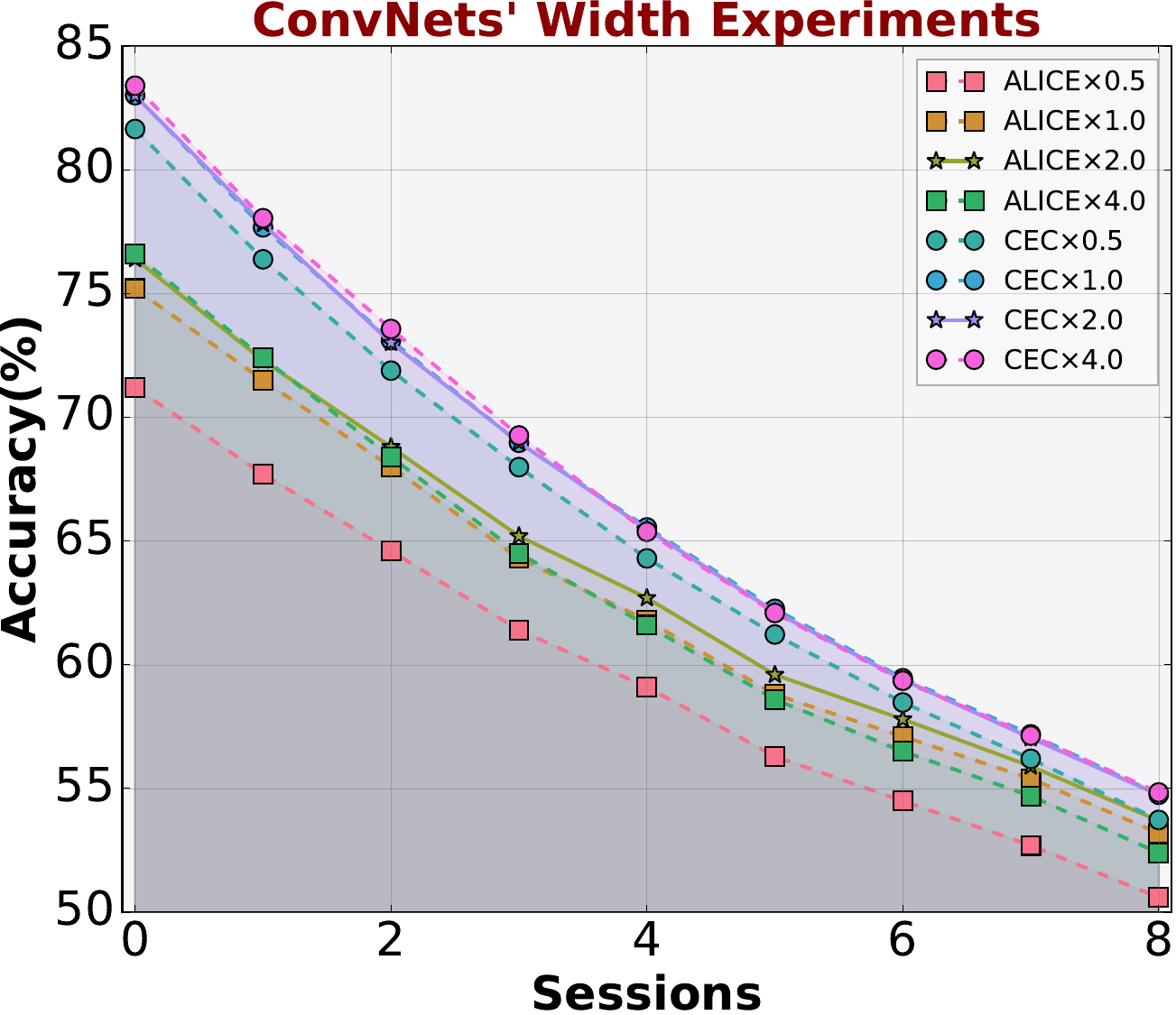}
        \caption{\footnotesize ResNet-12}
        \label{supp:resnet12_width}
    \end{subfigure}
    \hspace{2ex}
    \begin{subfigure}[b]{0.44\columnwidth}
        \centering
        \includegraphics[width=\columnwidth]{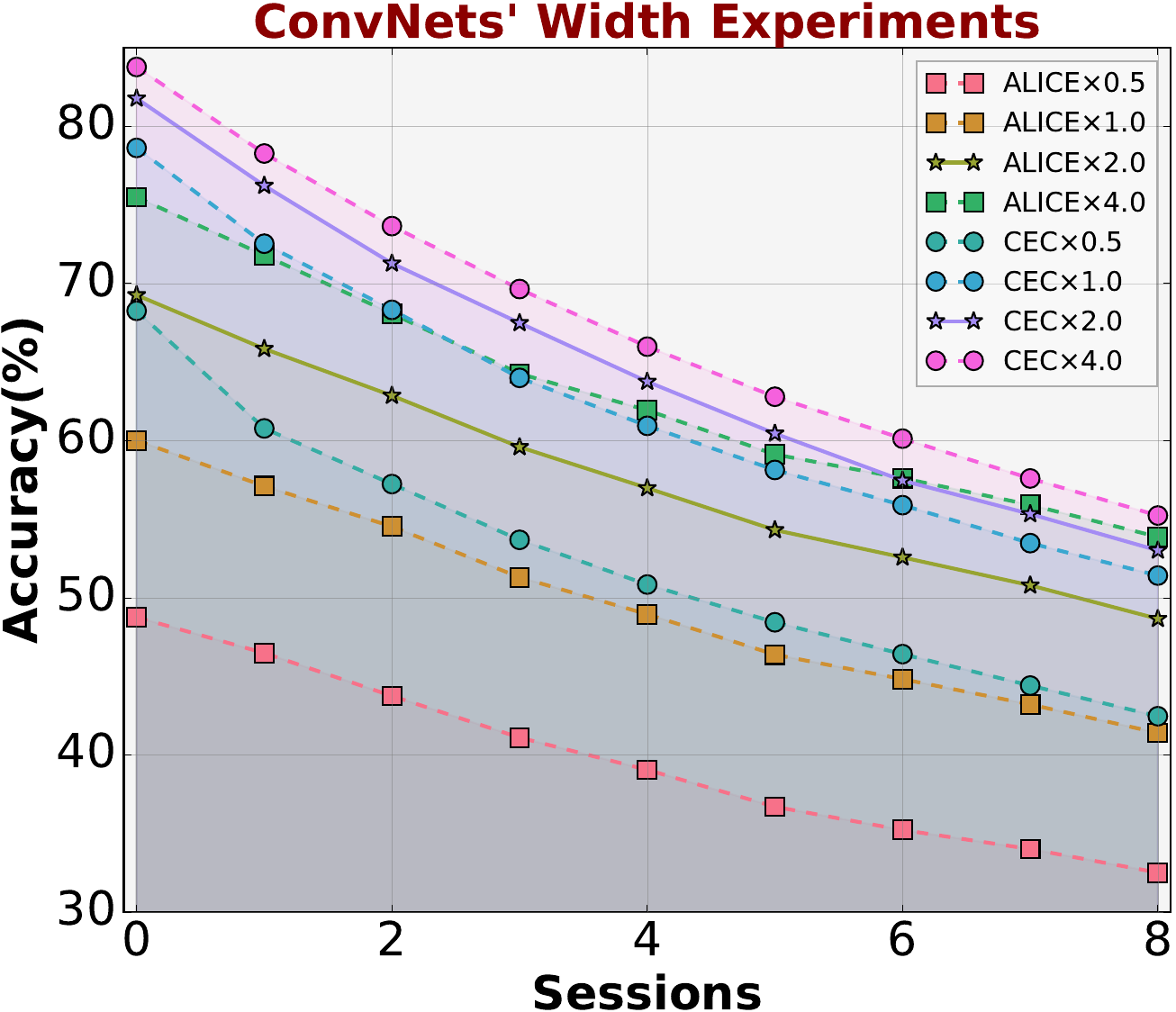}
        \caption{\footnotesize ResNet-20}
        \label{supp:resnet20_width}
    \end{subfigure}
    \hspace{2ex}
    \begin{subfigure}[b]{0.44\columnwidth}
        \centering
        \includegraphics[width=\columnwidth]{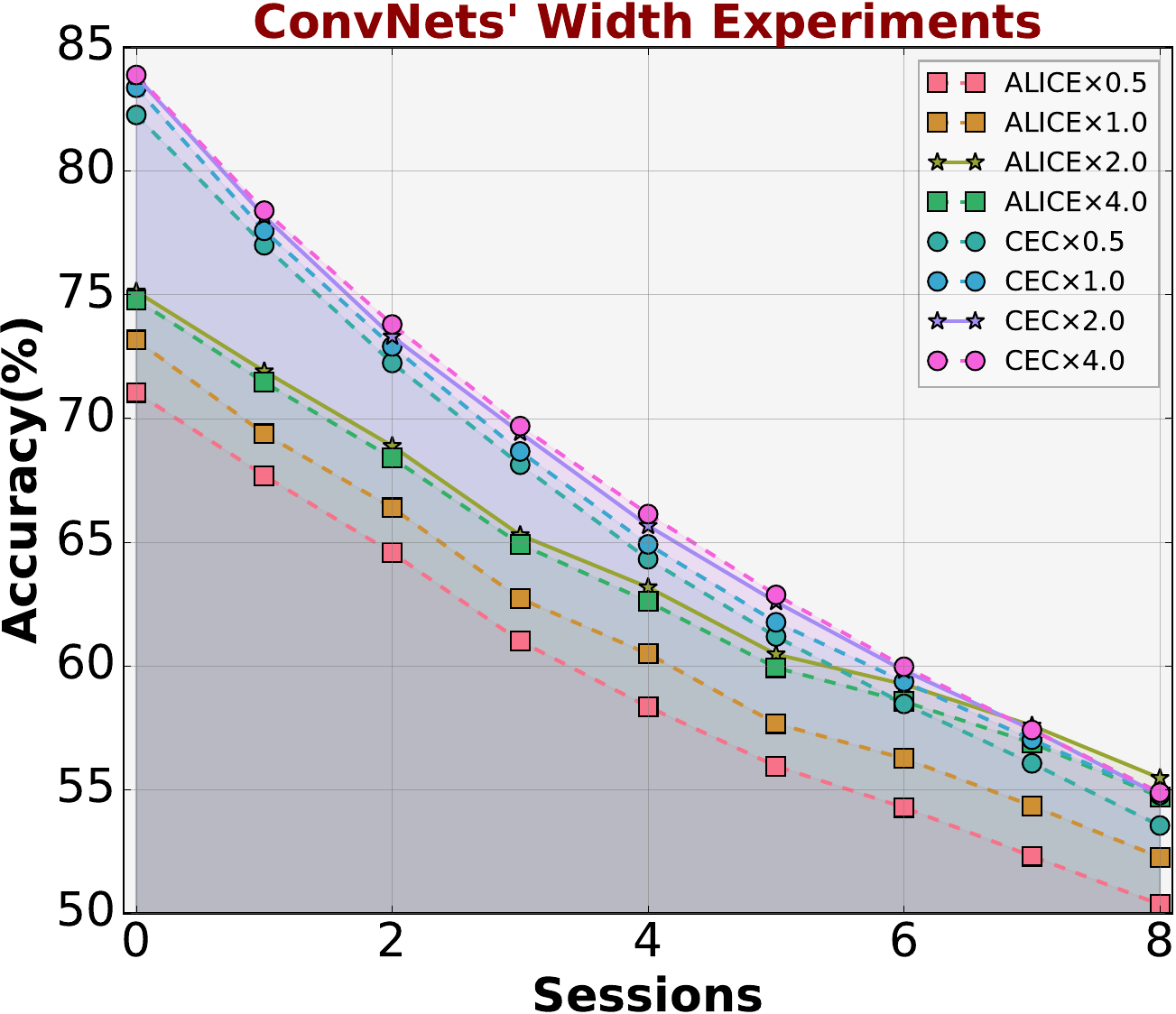}
        \caption{\footnotesize ResNet-34}
        \label{supp:resnet34_width}
    \end{subfigure}
    \hspace{2ex}
    \begin{subfigure}[b]{0.44\columnwidth}
        \centering
        \includegraphics[width=\columnwidth]{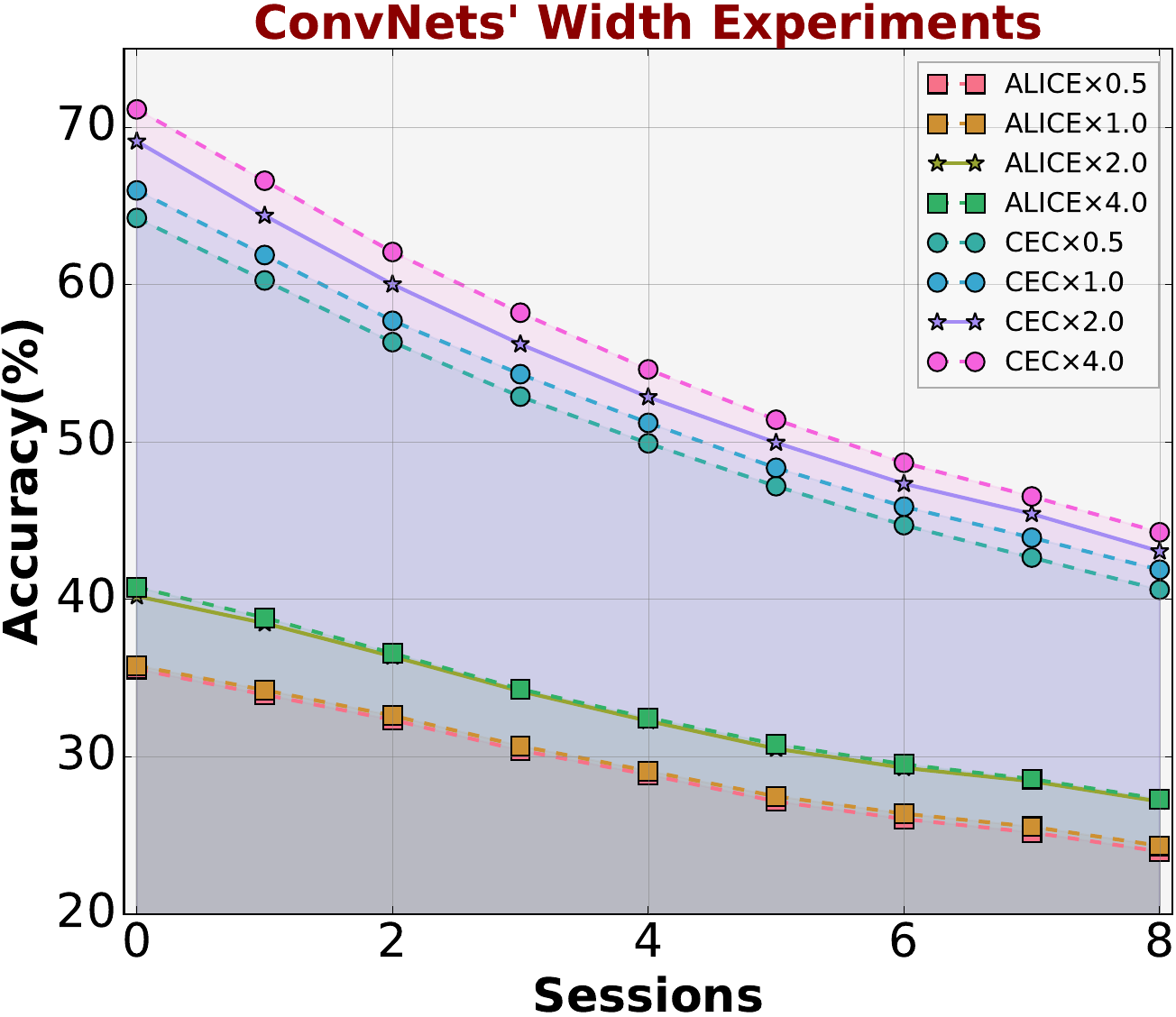}
        \caption{\footnotesize ConvNext-tiny}
        \label{supp:convnext_width}
    \end{subfigure}
\caption{This figure elucidates the FSCIL results of four different ConvNets (a.k.a. ResNet-12, ResNet-20, ResNet-34 and ConvNext), grounded on CEC and ALICE, following different width expansions on \textbf{CIFAR100}.}
\label{supp:width_experiments}
\end{figure*}

Subsequently, we extend this width modification approach to other members of the ResNet family, including ResNet-12, ResNet-20, ResNet-34 and ConvNext-tiny, conducting similar experiments on CIFAR100 as done with ResNet-18. The detailed outcomes of these experiments are presented in \cref{supp:width_experiments}. These results consistently demonstrate that, irrespective of the specific CNN architecture, the generalization capabilities in incremental sessions are positively influenced by the width of the convolutional layers. The wider these layers, the better the performance in terms of generalization and FSCIL. These findings corroborate the NTK theory's proposition regarding the relationship between a network's width and its generalization ability, even within different CNN architectures.

\subsection{Exploring Self-supervised Pre-training}\label{supp:self-supervised pre-training}
In this subsection, we explicate and scrutinize the related experimental specifics and outcomes associated with the self-supervised pre-training component.

\textbf{CEC:} In assessing CEC's performance, we exclude components such as the continually evolved classifier and the fine-tuning post pre-training, opting instead for the `autoaugment' image transformation. The performance analysis, as detailed in Table~\ref{cec self-supervised backbones}, involves six distinct self-supervised methodologies, excluding SimSiam \cite{chen2021exploring} and SwAV \cite{caron2020unsupervised} due to non-convergence on CIFAR100. The results indicate that self-supervised pre-training on base classes does not consistently enhance FSCIL performance for ConvNets in CEC. Only DINO \cite{caron2021emerging}, SparK \cite{tian2023designing} and MoCo-v3 \cite{chen2021mocov3} demonstrate positive results, whereas simCLR \cite{chen2020simple} and BYOL \cite{grill2020bootstrap} show negative effects. This disparity may stem from the reliance of some self-supervised algorithms on contrastive learning and hand-crafted view transformations, which can lead to representation collapse in ConvNets, especially given CIFAR100's small image size. Such issues seem less prevalent in generative methods like masked image modeling.

\textbf{ALICE:} ALICE undergoes similar adaptations to CEC, including the removal of the 128-dimensional fully-connected layer from the backbone, reduction in mix-up volume, abandonment of the dual-branch training model, and discontinuation of balanced testing.
Furthermore, ALICE's outcomes, as detailed in Table~\ref{alice self-supervised backbones}, align with those observed in CEC. The performance of contrastive learning methodologies in FSCIL is notably inconsistent, with only BYOL showing slight improvements. In contrast, generative strategies based on masked image modeling, particularly SparK, consistently perform well. This supports our earlier hypothesis and suggests the reliability of these strategies from a different perspective.

\begin{table*}[!t]
    \centering
    \setlength{\tabcolsep}{2.8mm}
    \renewcommand{\arraystretch}{1.0}
    \resizebox{0.8\linewidth}{!}{
        \begin{tabular}{cccccccccccc}
            \toprule
            \multirow{2}{*}{Method} & \multicolumn{10}{c}{Acc. in each session (\%) ↑} & \multicolumn{1}{l}{\multirow{2}{*}{PD $\downarrow$}} \\ \cline{2-11} & Backbone & 0 & 1 & 2 & 3 & 4 & 5 & 6 & 7 & 8 
            & \multicolumn{1}{l}{} \\ \hline
            $\textbf{CEC}$ \cite{zhang2021few} & ResNet-20 & 73.1 & 68.9 & 65.3 & 61.2 & 58.1 & 55.6 & 53.2 & 51.3 & 49.1 & 24.0 \\
            $\textbf{CEC}^*$ \cite{zhang2021few} & ResNet-20 & 79.0 & 73.1 & 68.8 & 64.8 & 61.4 & 58.3 & 56.0 & 53.8 & 51.7 & 27.3 \\
            $\textbf{CEC}^*$ \cite{zhang2021few} & ResNet-18$\times$2 & 85.3 & 79.7 & \textcolor[rgb]{0,0,1}{\textbf{74.9}} & 70.6 & 66.9 & \textcolor[rgb]{0,0,1}{\textbf{63.8}} & \textcolor[rgb]{1,0,0}{\textbf{61.3}} & \textcolor[rgb]{0,0,1}{\textbf{58.7}} & 55.2 & 30.1 \\
            $\textbf{CEC}^*$ \cite{zhang2021few} & ViT-tiny & 54.9 & 47.6 & 44.7 & 42.2 & 39.8 & 37.7 & 36.1 & 35.0 & 33.5 & \textcolor[rgb]{1,0,0}{\textbf{21.4}} \\
            $\textbf{CEC}^*$ \cite{zhang2021few} & ViT-small & 58.0 & 50.1 & 47.4 & 44.5 & 42.2 & 40.2 & 38.3 & 37.0 & 35.5 & 22.5 \\ \hline
            \textbf{DINO} \cite{caron2021emerging} & ResNet-18$\times$2 & \textcolor[rgb]{1,0,0}{\textbf{85.8}} & \textcolor[rgb]{1,0,0}{\textbf{80.1}} & \textcolor[rgb]{1,0,0}{\textbf{75.1}} & 70.1 & \textcolor[rgb]{0,0,1}{\textbf{67.0}} & 63.7 & 61.0 & 58.6 & \textcolor[rgb]{0,0,1}{\textbf{56.0}} & 29.8 \\
            \textbf{DINO} \cite{caron2021emerging} & ViT-tiny & 75.0 & 69.8 & 65.2 & 61.1 & 57.8 & 54.6 & 51.8 & 49.6 & 47.4 & \textcolor[rgb]{0,0,1}{\textbf{22.4}} \\
            \textbf{DINO} \cite{caron2021emerging} & ViT-small & 67.8 & 63.0 & 58.8 & 55.1 & 51.9 & 49.1 & 46.6 & 44.5 & 42.6 & 25.2 \\ \hline
            \textbf{MAE} \cite{he2022masked} & ViT-tiny & 78.6 & 73.8 & 69.1 & 64.9 & 61.3 & 58.3 & 55.4 & 53.0 & 50.4 & 28.2 \\
            \textbf{MAE} \cite{he2022masked} & ViT-small & 78.6 & 73.2 & 68.6 & 64.5 & 61.0 & 57.9 & 55.2 & 52.7 & 50.3 & 28.3 \\ \hline
            \textbf{SparK} \cite{tian2023designing} & ResNet-18$\times$2 & \textcolor[rgb]{0,0,1}{\textbf{85.5}} & \textcolor[rgb]{0,0,1}{\textbf{80.0}} & \textcolor[rgb]{1,0,0}{\textbf{75.1}} & \textcolor[rgb]{1,0,0}{\textbf{70.8}} & \textcolor[rgb]{1,0,0}{\textbf{67.1}} & \textcolor[rgb]{0,0,1}{\textbf{63.8}} & 60.9 & 58.6 & \textcolor[rgb]{1,0,0}{\textbf{56.2}} & 29.3 \\ \hline
            \textbf{MoCo-v3} \cite{chen2021mocov3} & ResNet-18$\times$2 & \textcolor[rgb]{0,0,1}{\textbf{85.5}} & 79.8 & 74.7 & \textcolor[rgb]{0,0,1}{\textbf{70.7}} & 67.0 & \textcolor[rgb]{1,0,0}{\textbf{63.9}} & \textcolor[rgb]{0,0,1}{\textbf{61.1}} & \textcolor[rgb]{1,0,0}{\textbf{58.9}} & \textcolor[rgb]{1,0,0}{\textbf{56.2}} & 29.3 \\
            \textbf{MoCo-v3} \cite{chen2021mocov3} & ViT-tiny & 69.4 & 64.8 & 60.9 & 57.3 & 54.0 & 51.3 & 48.8 & 46.5 & 51.7 & 28.1 \\
            \textbf{MoCo-v3} \cite{chen2021mocov3} & ViT-small & 74.1 & 69.6 & 65.9 & 62.0 & 58.8 & 55.9 & 53.4 & 51.3 & 49.2 & 24.9 \\ \hline
            \textbf{simCLR} \cite{chen2020simple} & ResNet-18$\times$2 & 81.3 & 76.0 & 71.4 & 67.3 & 63.6 & 60.4 & 57.9 & 55.5 & 53.0 & 25.8 \\
            \textbf{simCLR} \cite{chen2020simple} & ViT-tiny & 78.5 & 73.5 & 68.7 & 64.8 & 61.1 & 57.9 & 55.0 & 52.6 & 50.2 & 28.3 \\
            \textbf{simCLR} \cite{chen2020simple} & ViT-small & 78.4 & 72.8 & 68.2 & 63.9 & 60.3 & 57.2 & 54.5 & 52.1 & 49.8 & 28.6 \\ \hline
            \textbf{BYOL} \cite{grill2020bootstrap} & ResNet-18$\times$2 & 82.0 & 76.9 & 72.1 & 68.2 & 64.5 & 61.5 & 59.0 & 56.6 & 54.1 & 27.9 \\
            \textbf{BYOL} \cite{grill2020bootstrap} & ViT-tiny & 66.2 & 59.6 & 56.1 & 52.7 & 50.1 & 47.8 & 45.6 & 44.0 & 42.3 & 23.9 \\
            \textbf{BYOL} \cite{grill2020bootstrap} & ViT-small & 66.1 & 60.0 & 56.5 & 53.1 & 50.3 & 47.8 & 45.5 & 43.4 & 41.7 & 24.4 \\
            \bottomrule
        \end{tabular}
    }
\caption{The performance gleaned from various self-supervised learners on \textbf{CIFAR100}, anchored on $\textbf{CEC}$ and the modified $\textbf{CEC}^*$. The red-bolded segments represent the optimal results, while the blue-bolded segments represent suboptimal results.}
\label{cec self-supervised backbones}
\end{table*}

\begin{table*}[!t]
    \centering
    \setlength{\tabcolsep}{2.8mm}
    \renewcommand{\arraystretch}{1.0}
    \resizebox{0.8\linewidth}{!}{
        \begin{tabular}{cccccccccccc}
            \toprule
            \multirow{2}{*}{Method} & \multicolumn{10}{c}{Acc. in each session (\%) ↑} & \multicolumn{1}{l}{\multirow{2}{*}{PD $\downarrow$}} \\ \cline{2-11} & Backbone & 0 & 1 & 2 & 3 & 4 & 5 & 6 & 7 & 8 
            & \multicolumn{1}{l}{} \\ \hline
            $\textbf{ALICE}$ \cite{peng2022few} & ResNet-18 & 79.0 & 70.5 & 67.1 & 63.4 & 61.2 & 59.2 & 58.1 & 56.3 & 54.1 & 24.9 \\
            $\textbf{ALICE}^*$ \cite{peng2022few} & ResNet-18 & 78.8 & 75.2 & 72.3 & 68.7 & 65.5 & 62.1 & 60.7 & 58.3 & 56.6 & 22.2 \\
            $\textbf{ALICE}^*$ \cite{zhang2021few} & ResNet-18$\times$2 & 80.9 & 76.9 & 73.6 & \textcolor[rgb]{0,0,1}{\textbf{69.9}} & 66.9 & 63.9 & \textcolor[rgb]{0,0,1}{\textbf{62.5}} & 60.4 & \textcolor[rgb]{0,0,1}{\textbf{58.3}} & 22.6 \\
            $\textbf{ALICE}^*$ \cite{zhang2021few} & ViT-tiny & 58.7 & 55.4 & 52.3 & 49.0 & 46.8 & 44.4 & 42.4 & 40.8 & 39.1 & \textcolor[rgb]{1,0,0}{\textbf{19.6}} \\
            $\textbf{ALICE}^*$ \cite{zhang2021few} & ViT-small & 62.4 & 58.5 & 55.2 & 51.8 & 49.5 & 46.8 & 44.8 & 43.1 & 41.4 & 21.0 \\ \hline
            \textbf{DINO} \cite{caron2021emerging} & ResNet-18$\times$2 & \textcolor[rgb]{1,0,0}{\textbf{84.7}} & \textcolor[rgb]{1,0,0}{\textbf{79.1}} & \textcolor[rgb]{1,0,0}{\textbf{74.9}} & \textcolor[rgb]{1,0,0}{\textbf{70.3}} & 67.2 & 64.1 & 62.0 & 59.9 & 57.8 & 26.9 \\
            \textbf{DINO} \cite{caron2021emerging} & ViT-tiny & 82.7 & 76.9 & 71.9 & 67.4 & 63.9 & 60.7 & 57.7 & 55.1 & 52.8 & 29.9 \\
            \textbf{DINO} \cite{caron2021emerging} & ViT-small & 71.8 & 67.0 & 62.8 & 58.8 & 55.6 & 52.5 & 50.1 & 48.0 & 45.9 & 25.9 \\ \hline
            \textbf{MAE} \cite{he2022masked} & ViT-tiny & 81.9 & 76.7 & 71.7 & 67.5 & 64.2 & 60.9 & 58.1 & 55.5 & 53.2 & 28.7 \\
            \textbf{MAE} \cite{he2022masked} & ViT-small & 81.5 & 76.4 & 71.3 & 67.1 & 63.5 & 60.3 & 57.4 & 54.9 & 52.7 & 28.8 \\ \hline
            \textbf{SparK} \cite{tian2023designing} & ResNet-18$\times$2 & \textcolor[rgb]{0,0,1}{\textbf{83.6}} & \textcolor[rgb]{0,0,1}{\textbf{77.7}} & 73.7 & 69.8 & \textcolor[rgb]{1,0,0}{\textbf{67.4}} & \textcolor[rgb]{1,0,0}{\textbf{64.6}} & \textcolor[rgb]{1,0,0}{\textbf{63.0}} & \textcolor[rgb]{0,0,1}{\textbf{61.0}} & \textcolor[rgb]{1,0,0}{\textbf{58.8}} & 24.8 \\ \hline
            \textbf{MoCo-v3} \cite{chen2021mocov3} & ResNet-18$\times$2 & 82.4 & 77.5 & 73.3 & 68.9 & 66.2 & 63.1 & 60.9 & 58.8 & 56.8 & 25.6 \\
            \textbf{MoCo-v3} \cite{chen2021mocov3} & ViT-tiny & 78.7 & 73.4 & 68.7 & 64.7 & 61.1 & 57.8 & 55.2 & 52.8 & 50.5 & 28.2 \\
            \textbf{MoCo-v3} \cite{chen2021mocov3} & ViT-small & 78.4 & 73.1 & 68.4 & 64.0 & 60.3 & 57.3 & 54.5 & 52.1 & 49.8 & 28.6 \\ \hline
            \textbf{simCLR} \cite{chen2020simple} & ResNet-18$\times$2 & 74.1 & 70.4 & 67.3 & 63.8 & 61.2 & 58.2 & 56.7 & 55.2 & 53.1 & 21.0 \\
            \textbf{simCLR} \cite{chen2020simple} & ViT-tiny & 80.7 & 75.7 & 70.9 & 66.4 & 62.8 & 59.5 & 56.5 & 53.9 & 51.7 & 29.0 \\
            \textbf{simCLR} \cite{chen2020simple} & ViT-small & 79.8 & 74.8 & 69.9 & 65.6 & 62.1 & 58.8 & 56.0 & 53.5 & 51.3 & 28.5 \\ \hline
            \textbf{BYOL} \cite{grill2020bootstrap} & ResNet-18$\times$2 & 82.9 & \textcolor[rgb]{0,0,1}{\textbf{77.7}} & \textcolor[rgb]{0,0,1}{\textbf{73.9}} & 69.8 & \textcolor[rgb]{0,0,1}{\textbf{67.3}} & \textcolor[rgb]{0,0,1}{\textbf{64.5}} & \textcolor[rgb]{1,0,0}{\textbf{63.0}} & \textcolor[rgb]{1,0,0}{\textbf{61.1}} & \textcolor[rgb]{1,0,0}{\textbf{58.8}} & 24.1 \\
            \textbf{BYOL} \cite{grill2020bootstrap} & ViT-tiny & 56.9 & 53.5 & 50.4 & 47.3 & 45.0 & 42.7 & 40.7 & 39.0 & 37.2 & \textcolor[rgb]{0,0,1}{\textbf{19.7}} \\
            \textbf{BYOL} \cite{grill2020bootstrap} & ViT-small & 55.5 & 52.0 & 48.8 & 45.4 & 43.2 & 40.8 & 38.9 & 37.6 & 35.9 & \textcolor[rgb]{1,0,0}{\textbf{19.6}} \\
            \bottomrule
        \end{tabular}
    }
\caption{The performance from various self-supervised learners on \textbf{CIFAR100}, anchored on $\textbf{ALICE}$ and the modified $\textbf{ALICE}^*$. The red-bolded segments represent the optimal results, while the blue-bolded segments represent suboptimal results.}
\label{alice self-supervised backbones}
\end{table*}

In the spirit of rigorous scientific investigation, we extend our experiments to equivalent-sized Transformer-based models. Traditionally, these models are perceived as less effective than ConvNets on small datasets and few-shot tasks. However, our results indicate that with self-supervised pre-training, Transformers significantly close the performance gap, mirroring the results of the original CEC. This intriguing finding suggests that Transformer-based models may approach or even match the efficiency of ConvNets in few-shot scenarios, presenting an exciting avenue for future research. Unfortunately, for a fair comparison with other FSCIL methods and due to the limitations of Spark self-supervised pre-training, we do not use the ViT architecture for our experiments.

\subsection{Better Logit-Label Alignment} \label{supp:classification_loss}
Drawing from the findings of \cite{peng2022few}, we delve into the performance of various loss functions, placing particular emphasis on those rooted in margin-based losses. More detailed results are presented in Table~\ref{supp:cl_table}. Our empirical analysis uncovers significant performance limitations when solely relying on the traditional cross-entropy loss. Delving deeper, it becomes apparent that margin-based losses possess inherent advantages in handling a vast number of classes in the class extension mechanism. Among these margin-based losses, curricular alignment stands out significantly. This is largely attributed to its incorporation of the curriculum learning philosophy, mirroring the human education process of transitioning from simple to complex. It effectively balances the relationship between easy and hard samples and further alleviates the class imbalance issues introduced by the mixup mechanism. Furthermore, compared to other margin-based losses, our curricular alignment, despite lagging in base session accuracy, excels in incremental session accuracies. This indicates that it effectively learns from virtual classes, potentially reducing performance on real classes to some extent, but in turn, achieving better model generalization.

\begin{table*}[htbp]
    \centering
    \setlength{\tabcolsep}{2.8mm}
    \renewcommand{\arraystretch}{1.0}
    \resizebox{0.75\linewidth}{!}{
        \begin{tabular}{cccccccccc}
            \toprule
            \multirow{2}*{\textbf{Classification Loss}} & \multicolumn{9}{c}{Acc. in each session (\%) ↑}\\
            \cline{2-10} & 0     & 1     & 2     & 3     & 4     & 5     & 6     & 7     & 8 \\
            \hline
            \textbf{CE} & 24.5 & 23.1 & 21.4 & 19.9 & 19.1 & 18.1 & 17.3 & 16.7 & 15.9 \\
            \textbf{ArcFace} & 76.5 & 71.2 & 67.4 & 63.5 & 60.6 & 57.7 & 56.1 & 54.2 & 52.1 \\
            \textbf{CosFace} & 77.0 & 71.9 & 67.1 & 63.2 & 60.8 & 57.9 & 56.2 & 54.3 & 52.2 \\
            \textbf{Curricular} & 75.5 & 71.7 & 68.2 & 64.5 & 61.9 & 58.9 & 57.5 & 55.6 & 53.2 \\
            \bottomrule
        \end{tabular}
    }
\caption{The notations \textbf{CE}, \textbf{ArcFace}, \textbf{CosFace} and \textbf{Curricular} correspond to Cross Entropy Loss, ArcFace Loss in \cite{deng2019arcface}, CosFace Loss in \cite{wang2018cosface} and Curricular Loss respectively. All experiments are implemented on the ALICE \cite{peng2022few} framework.}
\label{supp:cl_table}
\end{table*}

For hyperparameters \(s\) and \(m\), our settings are closely aligned with those used in ALICE \cite{peng2022few}. Specifically, for CIFAR100 and \emph{mini}ImageNet, we set \(s\) to 15.0 and \(m\) to 0.1. For datasets with larger image size and more complex information, such as CUB200 and ImageNet100, we adjust \(s\) to 25.0 while keeping \(m\) at 0.1, to accommodate the need for greater separation between positive and negative classes.

\begin{table*}[htbp]
\centering
\setlength{\tabcolsep}{2.8mm}
\renewcommand{\arraystretch}{1.0}
\resizebox{1.0\linewidth}{!}{
\begin{tabular}{cccccccccc|ccc}
\toprule
\multirow{2}*{\textbf{Methods}} & \multicolumn{9}{c|}{\textbf{Acc. in each session (\%) ↑}} & \multirow{2}{*}{\textbf{Base Acc.}} & \multirow{2}{*}{\textbf{Incremental Acc.}} & \multirow{2}{*}{\textbf{Harmonic Mean}} \\
\cline{2-10}
& 0 & 1 & 2 & 3 & 4 & 5 & 6 & 7 & 8 & & & \\
\midrule
\textbf{Mix-up} \cite{zhang2018mixup} & 84.2 & 79.7 & 76.1 & 71.8 & 69.0 & 66.4 & 64.8 & 62.9 & 60.6 & 84.2 & 37.8 & 50.5 \\
\textbf{CutMix} \cite{yun2019cutmix} & 85.1 & 80.0 & 76.0 & 71.4 & 68.7 & 65.7 & 64.2 & 62.3 & 60.2 & 85.1 & 36.2 & 49.1 \\
\textbf{AugMix} \cite{2020augmix} & 80.8 & 76.8 & 73.2 & 69.2 & 66.3 & 63.1 & 61.7 & 59.9 & 57.8 & 80.8 & 33.3 & 46.0 \\
\textbf{PuzzleMix} \cite{kim2020puzzle} & 84.4 & 79.9 & 76.6 & 72.3 & 69.6 & 66.7 & 65.0 & 63.1 & 61.0 & 84.4 & 37.2 & 50.1 \\
\bottomrule
\end{tabular}
}
\caption{Average accuracy across base and incremental sessions using different mix-up methods, with their harmonic mean.}
\label{supp:mix_up_tables}
\end{table*}

\begin{table*}[!t]
    \centering
    \setlength{\tabcolsep}{2.8mm}
    \renewcommand{\arraystretch}{1.0}
    \resizebox{0.8\linewidth}{!}{
        \begin{tabular}{ccccccccccccccccc}
            \toprule
            \multirow{2}{*}{Method} & \multicolumn{12}{c}{Acc. in each session (\%) ↑} & \multirow{2}{*}{H $\uparrow$} & \multirow{2}{*}{PD $\downarrow$} \\ \cline{2-13}
            & Backbone & Params & FLOPs & 0 & 1 & 2 & 3 & 4 & 5 & 6 & 7 & 8 & & \\ \hline
            iCaRL \cite{rebuffi2017icarl} & ResNet-32 & 0.47M & 0.07G & 64.1 & 53.3 & 41.7 & 34.1 & 27.9 & 25.1 & 20.4 & 15.5 & 13.7 & 39.9 & 50.4 \\
            TOPIC \cite{tao2020few} & ResNet-18 & 11.17M & 0.56G & 64.1 & 55.9 & 47.1 & 45.2 & 40.1 & 36.4 & 34.0 & 31.6 & 29.4 & 49.2 & 34.7 \\
            CEC \cite{zhang2021few} & ResNet-20 & 0.27M & 0.04G & 73.1 & 68.9 & 65.3 & 61.2 & 58.1 & 55.6 & 53.2 & 51.3 & 49.1 & 64.6 & 24.0 \\
            LIMIT \cite{zhou2022few} & ResNet-20 & 0.27M & 0.04G & 73.0 & 70.8 & 67.5 & 63.4 & 60.0 & 56.9 & 54.8 & 52.2 & 49.9 & 65.5 & 23.1 \\
            FACT \cite{zhou2022forward} & ResNet-20 & 0.27M & 0.04G & 74.6 & 72.1 & 67.6 & 63.5 & 61.4 & 58.4 & 56.3 & 54.2 & 52.1 & 66.9 & \textcolor[rgb]{0,0,1}{\textbf{22.5}} \\
            ALICE \cite{peng2022few} & ResNet-18 & 11.17M & 0.56G & 79.0 & 70.5 & 67.1 & 63.4 & 61.2 & 59.2 & 58.1 & 56.3 & 54.1 & 69.0 & 24.9 \\
            GKEAL \cite{zhuang2023gkeal} & ResNet-20 & 0.27M & 0.04G & 74.0 & 70.5 & 67.0 & 63.1 & 60.0 & 57.3 & 55.5 & 53.4 & 51.4 & 66.1 & 22.6 \\
            WaRP \cite{kim2023warping} & ResNet-20 & 0.27M & 0.04G & 80.3 & 75.9 & 71.9 & 67.6 & 64.4 & 61.3 & 59.2 & 57.1 & 54.7 & 71.2 & 25.6 \\
            SoftNet \cite{yoon2023soft} & ResNet-18 & 11.17M & 0.56G & 79.9 & 75.5 & 71.6 & 67.5 & 64.5 & 61.1 & 59.1 & 57.3 & 55.3 & 71.1 & 24.6 \\
            NC-FSCIL \cite{yang2023neural} & ResNet-12 & 12.42M & 0.53G & 82.5 & 76.8 & 73.3 & 69.7 & 66.2 & 62.9 & 61.0 & 59.0 & 56.1 & 73.1 & 26.4 \\
            \hline
            CEC$\dag$ \cite{zhang2021few} & ResNet-18 & 11.17M & 0.56G & 78.3 & 73.3 & 68.8 & 64.9 & 61.9 & 59.1 & 57.1 & 54.6 & 52.5 & 68.9 & 25.8 \\
            CEC$\dag$ \cite{zhang2021few} & ResNet-18$\times$2 & 44.65M & 2.22G & 78.7 & 73.5 & 69.6 & 65.7 & 62.7 & 59.6 & 57.6 & 55.6 & 53.5 & 69.5 & 25.2 \\
            FACT$\dag$ \cite{zhou2022forward} & ResNet-18 & 11.17M & 0.56G & \textcolor[rgb]{0,0,1}{\textbf{83.3}} & 77.5 & 72.8 & 68.9 & 65.1 & 61.8 & 59.3 & 57.1 & 54.5 & 72.7 & 28.8 \\
            FACT$\dag$ \cite{zhou2022forward} & ResNet-18$\times$2 & 44.65M & 2.22G & 83.1 & \textcolor[rgb]{0,0,1}{\textbf{77.7}} & 73.0 & 69.0 & 65.4 & 62.3 & 59.7 & 57.6 & 55.2 & 72.9 & 27.9 \\
            ALICE$\dag$ \cite{peng2022few} & ResNet-18 & 11.17M & 0.56G & 77.1 & 71.0 & 66.9 & 62.8 & 60.3 & 57.9 & 57.1 & 55.6 & 53.6 & 67.9 & 23.5 \\
            ALICE$\dag$ \cite{peng2022few} & ResNet-18$\times$2 & 44.65M & 2.22G & 78.3 & 71.5 & 67.8 & 64.2 & 62.0 & 59.7 & 58.7 & 57.1 & 55.0 & 69.2 & 23.3 \\
            WaRP$\dag$ \cite{kim2023warping} & ResNet-18 & 11.17M & 0.56G & 80.1 & 75.2 & 71.0 & 66.7 & 62.9 & 60.0 & 57.7 & 55.4 & 53.4 & 71.0 & 26.7 \\
            WaRP$\dag$ \cite{kim2023warping} & ResNet-18$\times$2 & 44.65M & 2.22G & 80.8 & 76.2 & 71.7 & 67.7 & 64.2 & 61.1 & 58.6 & 56.5 & 54.5 & 71.3 & 26.3\\
            SoftNet$\dag$ \cite{yoon2023soft} & ResNet-18 & 11.17M & 0.56G & 80.0 & 75.8 & 72.0 & 67.9 & 64.7 & 61.8 & 59.8 & 58.0 & 55.6 & 71.4 & 24.4 \\
            SoftNet$\dag$ \cite{yoon2023soft} & ResNet-18$\times$2 & 44.65M & 2.22G & 81.1 & 76.8 & 73.1 & 68.9 & 65.6 & 62.4 & 60.1 & 58.1 & 55.8 & 72.2 & \textcolor[rgb]{1,0,0}{\textbf{21.0}} \\
            \hline
            \textbf{NTK-FSCIL} & ResNet-18 & 11.17M & 0.56G & 81.8 & 77.3 & \textcolor[rgb]{0,0,1}{\textbf{74.3}} & \textcolor[rgb]{0,0,1}{\textbf{69.8}} & \textcolor[rgb]{0,0,1}{\textbf{67.1}} & \textcolor[rgb]{0,0,1}{\textbf{64.2}} & \textcolor[rgb]{0,0,1}{\textbf{62.9}} & \textcolor[rgb]{0,0,1}{\textbf{61.1}} & \textcolor[rgb]{0,0,1}{\textbf{58.7}} & \textcolor[rgb]{0,0,1}{\textbf{73.6}} & 23.1 \\
            \textbf{NTK-FSCIL*} & ResNet-18$\times$2 & 44.65M & 2.22G & \textcolor[rgb]{1,0,0}{\textbf{84.1}} & \textcolor[rgb]{1,0,0}{\textbf{79.9}} & \textcolor[rgb]{1,0,0}{\textbf{76.9}} & \textcolor[rgb]{1,0,0}{\textbf{72.4}} & \textcolor[rgb]{1,0,0}{\textbf{69.9}} & \textcolor[rgb]{1,0,0}{\textbf{66.8}} & \textcolor[rgb]{1,0,0}{\textbf{65.5}} & \textcolor[rgb]{1,0,0}{\textbf{63.6}} & \textcolor[rgb]{1,0,0}{\textbf{61.3}} & \textcolor[rgb]{1,0,0}{\textbf{76.1}} & 22.8 \\
            \bottomrule
        \end{tabular}
    }
    \caption{Performance evaluation on \textbf{CIFAR100} dataset within the context of a 5-way 5-shot class-incremental learning paradigm. The red-bolded segments represent the optimal results, while the blue-bolded segments represent suboptimal results.}
    \label{tab:CIFAR100}
\end{table*}

\begin{table*}[!t]
    \centering
    \setlength{\tabcolsep}{2.8mm}
    \renewcommand{\arraystretch}{1.0}
    \resizebox{0.8\linewidth}{!}{
        \begin{tabular}{ccccccccccccccccc}
            \toprule
            \multirow{2}{*}{Method} & \multicolumn{12}{c}{Acc. in each session (\%) ↑} & \multirow{2}{*}{H $\uparrow$} & \multirow{2}{*}{PD $\downarrow$} \\ \cline{2-13}
            & Backbone & Params & FLOPs & 0 & 1 & 2 & 3 & 4 & 5 & 6 & 7 & 8 & & \\ \hline
            iCaRL \cite{rebuffi2017icarl} & ResNet-18 & 11.17M & 3.94G & 61.3 & 46.3 & 42.9 & 37.6 & 30.5 & 24.0 & 20.9 & 18.8 & 17.2 & 40.1 & 44.1 \\
            TOPIC \cite{tao2020few} & ResNet-18 & 11.17M & 3.94G & 61.3 & 50.1 & 45.2 & 41.2 & 37.5 & 35.5 & 32.2 & 29.5 & 24.4 & 46.1 & 36.9 \\
            CEC \cite{zhang2021few} & ResNet-18 & 11.17M & 3.94G & 72.0 & 66.8 & 63.0 & 59.4 & 56.7 & 53.7 & 51.2 & 49.2 & 47.6 & 63.0 & 24.4 \\
            LIMIT \cite{zhou2022few} & ResNet-18 & 11.17M & 3.94G & 71.9 & 67.9 & 63.6 & 60.2 & 57.3 & 54.3 & 52.0 & 50.0 & 48.4 & 63.4 & 23.5 \\
            FACT \cite{zhou2022forward} & ResNet-18 & 11.17M & 3.94G & 72.6 & 69.6 & 66.4 & 62.8 & 60.6 & 57.3 & 54.3 & 52.2 & 50.5 & 65.2 & \textcolor[rgb]{0,0,1}{\textbf{22.1}} \\
            ALICE \cite{peng2022few} & ResNet-18 & 11.17M & 3.94G & 80.6 & 70.6 & 67.4 & 64.5 & 62.5 & 60.0 & 57.8 & 56.8 & 55.7 & 70.0 & 24.9 \\
            GKEAL \cite{zhuang2023gkeal} & ResNet-18 & 11.17M & 3.94G & 73.6 & 68.9 & 65.3 & 62.3 & 59.4 & 56.7 & 54.2 & 52.6 & 51.3 & 65.4 & 22.3 \\
            WaRP \cite{kim2023warping} & ResNet-18 & 11.17M & 3.94G & 73.0 & 68.1 & 64.3 & 61.3 & 58.6 & 56.1 & 53.4 & 51.7 & 50.7 & 64.7 & 22.3 \\
            SoftNet \cite{yoon2023soft} & ResNet-18 & 11.17M & 3.94G & 79.4 & 74.3 & 69.9 & 66.2 & 63.4 & 60.8 & 57.6 & 55.7 & 54.3 & 70.1 & 25.1 \\
            NC-FSCIL \cite{yang2023neural} & ResNet-12 & 12.42M & 3.53G & \textcolor[rgb]{0,0,1}{\textbf{84.0}} & 76.8 & 72.0 & 67.8 & 66.4 & 64.0 & \textcolor[rgb]{0,0,1}{\textbf{61.5}} & 59.5 & 58.3 & 73.8 & 25.7 \\ 
            \hline
            CEC$\dag$ \cite{zhang2021few} & ResNet-18 & 11.17M & 3.94G & 83.1 & 77.8 & 73.3 & 69.7 & 66.1 & 62.9 & 60.1 & 58.2 & 57.0 & 73.3 & 26.1 \\
            CEC$\dag$ \cite{zhang2021few} & ResNet-18$\times$2 & 44.65M & 15.69G & 82.8 & \textcolor[rgb]{1,0,0}{\textbf{79.2}} & \textcolor[rgb]{1,0,0}{\textbf{74.8}} & \textcolor[rgb]{1,0,0}{\textbf{71.1}} & \textcolor[rgb]{1,0,0}{\textbf{67.8}} & \textcolor[rgb]{0,0,1}{\textbf{64.2}} & 61.3 & 59.4 & 57.9 & \textcolor[rgb]{1,0,0}{\textbf{74.0}} & 24.9 \\
            FACT$\dag$ \cite{zhou2022forward} & ResNet-18 & 11.17M & 3.94G & 82.6 & 78.2 & 73.9 & 70.1 & 66.6 & 63.2 & 60.1 & 58.2 & 56.6 & 73.3 & 26.0 \\
            FACT$\dag$ \cite{zhou2022forward} & ResNet-18$\times$2 & 44.65M & 15.69G & 82.9 & \textcolor[rgb]{0,0,1}{\textbf{78.8}} & \textcolor[rgb]{0,0,1}{\textbf{74.5}} & \textcolor[rgb]{0,0,1}{\textbf{70.6}} & \textcolor[rgb]{0,0,1}{\textbf{67.5}} & 64.1 & 60.9 & 59.0 & 57.3 & \textcolor[rgb]{0,0,1}{\textbf{73.9}} & 25.6 \\
            ALICE$\dag$ \cite{peng2022few} & ResNet-18 & 11.17M & 3.94G & 81.8 & 71.9 & 68.4 & 64.5 & 61.4 & 58.6 & 56.8 & 55.2 & 53.9 & 70.1 & 27.9 \\
            ALICE$\dag$ \cite{peng2022few} & ResNet-18$\times$2 & 44.65M & 15.69G & \textcolor[rgb]{1,0,0}{\textbf{84.5}} & 75.0 & 71.1 & 66.6 & 63.7 & 60.7 & 58.2 & 56.9 & 55.4 & 72.5 & 29.1 \\
            WaRP$\dag$ \cite{kim2023warping} & ResNet-18 & 11.17M & 3.94G & 82.6 & 77.3 & 73.1 & 69.5 & 67.0 & 63.8 & 60.7 & 58.7 & 57.4 & 73.3 & 25.2 \\
            WaRP$\dag$ \cite{kim2023warping} & ResNet-18$\times$2 & 44.65M & 15.69G & 82.5 & 76.9 & 72.7 & 69.2 & 66.4 & 63.4 & 60.4 & 58.3 & 57.0 & 73.0 & 25.5 \\
            SoftNet$\dag$ \cite{yoon2023soft} & ResNet-18 & 11.17M & 3.94G & 80.4 & 75.4 & 71.3 & 66.9 & 64.0 & 61.0 & 58.4 & 56.2 & 55.0 & 71.0 & 25.4 \\
            SoftNet$\dag$ \cite{yoon2023soft} & ResNet-18$\times$2 & 44.65M & 15.69G & 81.9 & 77.1 & 73.0 & 69.0 & 65.8 & 62.4 & 59.7 & 57.7 & 56.2 & 72.5 & 25.7 \\
            \hline
            \textbf{NTK-FSCIL} & ResNet-18 & 11.17M & 3.94G & 80.0 & 75.5 & 71.6 & 68.6 & 66.4 & 64.0 & 61.3 & \textcolor[rgb]{0,0,1}{\textbf{59.9}} & \textcolor[rgb]{0,0,1}{\textbf{58.8}} & 72.2 & \textcolor[rgb]{1,0,0}{\textbf{21.2}} \\
            \textbf{NTK-FSCIL} & ResNet-18$\times$2 & 44.65M & 15.69G & 82.4 & 77.2 & 73.3 & 70.2 & \textcolor[rgb]{1,0,0}{\textbf{67.8}} & \textcolor[rgb]{1,0,0}{\textbf{65.3}} & \textcolor[rgb]{1,0,0}{\textbf{62.7}} & \textcolor[rgb]{1,0,0}{\textbf{60.9}} & \textcolor[rgb]{1,0,0}{\textbf{60.0}} & \textcolor[rgb]{1,0,0}{\textbf{74.0}} & 22.4 \\
            \bottomrule
        \end{tabular}
    }
\caption{Performance evaluation on \textbf{\emph{mini}ImageNet} dataset within the context of a 5-way 5-shot class-incremental learning paradigm. The red-bolded segments represent the optimal results, while the blue-bolded segments represent suboptimal results.}
\label{tab:mini-ImageNet}
\end{table*}

\begin{table*}[!t]
    \centering
    \setlength{\tabcolsep}{2.8mm}
    \renewcommand{\arraystretch}{1.0}
    \resizebox{0.8\linewidth}{!}{
        \begin{tabular}{ccccccccccccccccc}
            \toprule
            \multirow{2}{*}{Method} & \multicolumn{14}{c}{Acc. in each session (\%) ↑} & \multirow{2}{*}{H $\uparrow$} & \multirow{2}{*}{PD $\downarrow$} \\ \cline{2-15} 
            & Backbone & Params & FLOPs & 0 & 1 & 2 & 3 & 4 & 5 & 6 & 7 & 8 & 9 & 10 & \\ \hline
            iCaRL \cite{rebuffi2017icarl} & ResNet-18 & 11.17M & 27.37G & 68.7 & 52.7 & 48.6 & 44.2 & 36.6 & 29.5 & 27.8 & 26.3 & 24.0 & 23.9 & 21.2 & 45.0 & 47.5 \\
            TOPIC \cite{tao2020few} & ResNet-18 & 11.17M & 27.37G & 68.7 & 62.5 & 54.8 & 50.0 & 45.3 & 41.4 & 38.4 & 35.4 & 32.2 & 28.3 & 26.3 & 51.7 & 42.4 \\
            CEC \cite{zhang2021few} & ResNet-18 & 11.17M & 27.37G & 75.9 & 71.9 & 68.5 & 63.5 & 62.4 & 58.3 & 57.7 & 55.8 & 54.8 & 53.5 & 52.3 & 66.9 & 23.6 \\
            LIMIT \cite{zhou2022few} & ResNet-18 & 11.17M & 27.37G & 76.3 & 74.2 & 72.7 & 69.2 & 68.8 & 65.6 & 63.6 & 62.7 & 61.5 & 60.4 & 58.4 & 70.6 & 17.9 \\
            FACT \cite{zhou2022forward} & ResNet-18 & 11.17M & 27.37G & 75.9 & 73.2 & 70.8 & 66.1 & 65.6 & 62.2 & 61.7 & 59.8 & 58.4 & 57.9 & 56.9 & 69.0 & 19.0 \\
            ALICE \cite{peng2022few} & ResNet-18 & 11.17M & 27.37G & 77.4 & 72.7 & 70.6 & 67.2 & 65.9 & 63.4 & 62.9 & 61.9 & 60.5 & 60.6 & 60.1 & 70.4 & 17.3 \\
            GKEAL \cite{zhuang2023gkeal} & ResNet-18 & 11.17M & 27.37G & 78.9 & 75.6 & 72.3 & 68.6 & 67.2 & 64.3 & 63.0 & 61.9 & 60.2 & 59.2 & 58.7 & 71.3 & 20.2 \\
            WaRP \cite{kim2023warping} & ResNet-18 & 11.17M & 27.37G & 77.7 & 74.2 & 70.8 & 66.9 & 65.0 & 62.6 & 61.4 & 59.9 & 58.0 & 57.8 & 57.0 & 69.8 & 20.7 \\
            SoftNet\cite{yoon2023soft} & ResNet-18 & 11.17M & 27.37G & 78.1 & 74.6 & 71.3 & 67.5 & 65.1 & 62.4 & 60.8 & 59.2 & 57.4 & 57.1 & 56.6 & 69.9 & 21.5 \\
            NC-FSCIL \cite{yang2023neural} & ResNet-18 & 11.17M & 27.37G & 80.5 & 76.0 & 72.3 & 70.3 & 68.2 & 65.2 & 64.4 & 63.3 & 60.7 & 60.0 & 59.4 & 72.5 & 21.1 \\ 
            \hline
            CEC$\dag$ \cite{zhang2021few} & Swin-Small & 50.0M & 8.7G & \textcolor[rgb]{1,0,0}{\textbf{84.2}} & 81.6 & 79.8 & 77.2 & 76.5 & 75.1 & 74.2 & 72.6 & 72.3 & 72.0 & 72.0 & 79.5 & 12.2 \\
            CEC$\dag$ \cite{zhang2021few} & ConvNeXt-Base & 88.6M & 15.4G & 83.3 & 81.2 & 79.9 & 77.3 & \textcolor[rgb]{0,0,1}{\textbf{77.2}} & \textcolor[rgb]{1,0,0}{\textbf{75.7}} & \textcolor[rgb]{1,0,0}{\textbf{74.8}} & \textcolor[rgb]{1,0,0}{\textbf{74.7}} & 73.9 & \textcolor[rgb]{0,0,1}{\textbf{73.9}} & 73.6 & 79.6 & 9.7 \\
            FACT$\dag$ \cite{zhou2022forward} & Swin-Small & 50.0M & 8.7G & 83.4 & 79.4 & 78.4 & \textcolor[rgb]{0,0,1}{\textbf{77.9}} & \textcolor[rgb]{1,0,0}{\textbf{77.7}} & \textcolor[rgb]{1,0,0}{\textbf{75.7}} & \textcolor[rgb]{1,0,0}{\textbf{74.8}} & 74.4 & 73.4 & 72.9 & 72.9 & 79.4 & 10.5 \\
            FACT$\dag$ \cite{zhou2022forward} & ConvNeXt-Base & 88.6M & 15.4G & 83.8 & 80.6 & 79.8 & 77.5 & 76.8 & \textcolor[rgb]{0,0,1}{\textbf{75.2}} & \textcolor[rgb]{1,0,0}{\textbf{74.8}} & 73.8 & 73.7 & 73.7 & 73.4 & 79.7 & 12.4 \\
            ALICE$\dag$ \cite{peng2022few} & Swin-Small & 50.0M & 8.7G & 79.2 & 76.6 & 73.5 & 70.1 & 68.2 & 66.3 & 65.0 & 64.3 & 63.0 & 62.8 & 62.3 & 72.7 & 16.9 \\
            ALICE$\dag$ \cite{peng2022few} & ConvNeXt-Base & 88.6M & 15.4G & 82.7 & 80.9 & 79.8 & 77.7 & 77.1 & 74.9 & \textcolor[rgb]{0,0,1}{\textbf{74.6}} & \textcolor[rgb]{0,0,1}{\textbf{74.6}} & 73.9 & 73.7 & \textcolor[rgb]{0,0,1}{\textbf{73.7}} & 79.3 & \textcolor[rgb]{1,0,0}{\textbf{9.0}} \\
            WaRP$\dag$ \cite{kim2023warping} & Swin-Small & 50.0M & 8.7G & \textcolor[rgb]{0,0,1}{\textbf{84.1}} & 80.8 & 78.7 & 75.2 & 73.9 & 71.8 & 71.0 & 70.7 & 69.1 & 68.9 & 68.6 & 78.1 & 15.5 \\
            WaRP$\dag$ \cite{kim2023warping} & ConvNeXt-Base & 88.6M & 15.4G & \textcolor[rgb]{1,0,0}{\textbf{84.2}} & \textcolor[rgb]{0,0,1}{\textbf{81.7}} & \textcolor[rgb]{1,0,0}{\textbf{80.3}} & \textcolor[rgb]{0,0,1}{\textbf{77.9}} & 76.7 & 75.0 & 74.2 & 74.1 & 72.8 & 72.6 & 72.6 & \textcolor[rgb]{0,0,1}{\textbf{79.8}} & 11.6 \\
            \hline
            \textbf{NTK-FSCIL} & ResNet-18 & 11.17M & 27.37G & 78.1 & 75.0 & 72.0 & 68.6 & 67.1 & 64.6 & 63.6 & 62.7 & 61.1 & 61.0 & 60.4 & 71.3 & 17.9 \\
            \textbf{NTK-FSCIL} & Swin-Small & 50.0M & 8.7G & 84.0 & \textcolor[rgb]{1,0,0}{\textbf{81.8}} & \textcolor[rgb]{0,0,1}{\textbf{80.1}} & 77.4 & 76.6 & 74.6 & \textcolor[rgb]{0,0,1}{\textbf{74.6}} & 74.4 & \textcolor[rgb]{0,0,1}{\textbf{74.0}} & \textcolor[rgb]{0,0,1}{\textbf{73.9}} & 73.6 & \textcolor[rgb]{1,0,0}{\textbf{79.9}} & 10.4 \\
            \textbf{NTK-FSCIL} & ConvNeXt-Base & 88.6M & 15.4G & 83.3 & 81.3 & \textcolor[rgb]{0,0,1}{\textbf{80.1}} & \textcolor[rgb]{1,0,0}{\textbf{78.0}} & \textcolor[rgb]{0,0,1}{\textbf{77.2}} & 74.9 & \textcolor[rgb]{0,0,1}{\textbf{74.6}} & 74.5 & \textcolor[rgb]{1,0,0}{\textbf{74.3}} & \textcolor[rgb]{1,0,0}{\textbf{74.2}} & \textcolor[rgb]{1,0,0}{\textbf{74.0}} & 79.7 & \textcolor[rgb]{0,0,1}{\textbf{9.3}} \\
            \bottomrule
        \end{tabular}
    }
\caption{Performance evaluation on \textbf{CUB-200-2011} dataset within the context of a 10-way 5-shot class-incremental learning paradigm. The red-bolded segments represent the optimal results, while the blue-bolded segments represent suboptimal results.}
\label{tab:cub200}
\end{table*}

\begin{table*}[!t]
    \centering
    \setlength{\tabcolsep}{2.8mm}
    \renewcommand{\arraystretch}{1.0}
    \resizebox{0.8\linewidth}{!}{
        \begin{tabular}{ccccccccccccccc}
            \toprule
            \multirow{2}{*}{Method} & \multicolumn{12}{c}{Acc. in each session (\%) ↑} & \multirow{2}{*}{H $\uparrow$} & \multirow{2}{*}{PD $\downarrow$} \\ \cline{2-13} 
            & Backbone & Params & FLOPs & 0 & 1 & 2 & 3 & 4 & 5 & 6 & 7 & 8 & \\ \hline
            CEC$\dag$ \cite{zhang2021few} & ResNet-18 & 11.17M & 27.37G & 72.2 & 67.4 & 64.0 & 60.7 & 57.6 & 54.7 & 53.2 & 52.5 & 50.4 & 64.1 & 21.8 \\
            CEC$\dag$ \cite{zhang2021few} & ResNet-18$\times$2 & 44.65M & 108.99G & 72.9 & 69.7 & 66.9 & 64.3 & 61.0 & 58.5 & \textcolor[rgb]{0,0,1}{\textbf{57.5}} & \textcolor[rgb]{0,0,1}{\textbf{56.9}} & \textcolor[rgb]{0,0,1}{\textbf{54.9}} & 66.5 & \textcolor[rgb]{1,0,0}{\textbf{18.0}} \\
            ALICE$\dag$ \cite{peng2022few} & ResNet-18 & 11.17M & 27.37G & 71.9 & 67.8 & 64.1 & 61.7 & 58.6 & 56.5 & 55.3 & 54.8 & 52.6 & 64.8 & \textcolor[rgb]{0,0,1}{\textbf{19.3}} \\
            ALICE$\dag$ \cite{peng2022few} & ResNet-18$\times$2 & 44.65M & 108.99G & \textcolor[rgb]{0,0,1}{\textbf{74.6}} & \textcolor[rgb]{0,0,1}{\textbf{70.2}} & \textcolor[rgb]{0,0,1}{\textbf{67.0}} & 63.8 & 60.5 & 57.2 & 55.7 & 55.0 & 52.8 & 66.7 & 21.8 \\
            \hline
            \textbf{NTK-FSCIL} & ResNet-18 & 11.17M & 27.37G & 73.8 & 70.0 & \textcolor[rgb]{0,0,1}{\textbf{67.0}} & \textcolor[rgb]{0,0,1}{\textbf{64.5}} & \textcolor[rgb]{0,0,1}{\textbf{61.1}} & \textcolor[rgb]{0,0,1}{\textbf{58.7}} & 57.1 & 56.4 & 54.3 & \textcolor[rgb]{0,0,1}{\textbf{66.9}} & 19.5 \\
            \textbf{NTK-FSCIL} & ResNet-18×2 & 44.65M & 108.99G & \textcolor[rgb]{1,0,0}{\textbf{75.9}} & \textcolor[rgb]{1,0,0}{\textbf{72.2}} & \textcolor[rgb]{1,0,0}{\textbf{68.9}} & \textcolor[rgb]{1,0,0}{\textbf{66.1}} & \textcolor[rgb]{1,0,0}{\textbf{62.7}} & \textcolor[rgb]{1,0,0}{\textbf{60.3}} & \textcolor[rgb]{1,0,0}{\textbf{58.6}} & \textcolor[rgb]{1,0,0}{\textbf{58.4}} & \textcolor[rgb]{1,0,0}{\textbf{56.2}} & \textcolor[rgb]{1,0,0}{\textbf{68.8}} & 19.7 \\
            \bottomrule
        \end{tabular}
    }
    \caption{Performance evaluation on \textbf{ImageNet100} dataset within the context of a 5-way 5-shot class-incremental learning paradigm. The red-bolded segments represent the optimal results, while the blue-bolded segments represent suboptimal results.}
    \label{tab:ImageNet100}
\end{table*}

\subsection{Exploring Logits Diversity in Meta-Learning}\label{supp:mix_ups}
In FSCIL, the prevailing approaches often leverage the mix-up mechanism for class extension, directly merging two images \cite{peng2022few,yang2023neural}. In this subsection, we delve into the nuanced variations of this mechanism. To investigate the impact of virtual classes generated by different mixup mechanisms on logits diversity, and further explore their effect on model generalization, we employ the harmonic mean for an intuitive representation.
\begin{align} \label{harmonic_mean}
    H = \frac{2 \times Base \ Acc \times Incremental \ Acc}{Base \ Acc + Incremental \ Acc}.
\end{align}

We conduct experiments with various mixup mechanisms on CIFAR100, and the results are displayed in Table~\ref{supp:mix_up_tables}. As shown in Table~\ref{supp:mix_up_tables}, methods that perform mixup operations globally on images produce better virtual samples and more flexible logit combinations, resulting in improved base session accuracy and harmonic mean. However, mixup methods involving image transformations fail to alleviate the rigid logit combinations in meta-learning. This could be due to the small image size of CIFAR100, where overly complex image transformations might irreversibly damage the original information and generate unrealistic virtual samples.

\subsection{Comparison with State-of-the-Art} 
In this subsection, we report the primary results for CIFAR100 in Table~\ref{tab:CIFAR100}, \emph{mini}ImageNet in Table~\ref{tab:mini-ImageNet}, CUB200 in Table~\ref{tab:cub200}, and ImageNet100 in Table~\ref{tab:ImageNet100}. To ensure a comprehensive comparison, we report both the original results from various methods and our re-implementation using open-source code and optimal parameters on identical hardware for ResNet-18 and ResNet $\times$ 2. Our experimental results demonstrate that NTK-FSCIL consistently outperforms competing methods across all datasets. This superiority is evident in three key metrics: (1) end-session accuracy, (2) PD, defined as the disparity between initial and final session accuracies, and (3) harmonic mean ($H$), as shown in Eq.~\ref{harmonic_mean}. A higher end-session accuracy indicates improved generalization across incremental sessions, while a lower PD value suggests enhanced anti-amnesia capability. The harmonic mean provides a comprehensive measure of model performance, demonstrating that generalization improves with increasing network width, rather than merely enhancing overall model expressivity. For instance, our method demonstrates a notable improvement on CIFAR100, achieving an 9.3\% increase in end-session accuracy, an 13.6\% enhancement in PD, and a 4.1\% boost in $H$ compared to the next best method, NC-FSCIL \cite{yang2023neural}. Furthermore, similar outcomes are also observed on other datasets. For both \emph{mini}ImageNet and ImageNet100, our approach on the ResNet-18×2 architecture demonstrates significant improvements: it achieves a 3.5\% gain in end-session accuracy over the second-best-performing CEC \cite{zhang2021few} using the same architecture for \emph{mini}ImageNet, and for ImageNet100, it yields enhancements of 6.0\% in end-session accuracy and 3.1\% in \( H \) over the second-best-performing ALICE \cite{peng2022few} also using the same architecture.

For CUB200, as most methodologies \cite{zhang2021few, peng2022few, yang2023neural} leverage pre-trained ResNet-18, bypassing both the widening and self-supervised modules is necessary to ensure fair comparison. Additionally, we also experiment with and report the results of using Swin Transformer \cite{liu2021swin} and ConvNeXt \cite{liu2022convnet} pre-trained weights for other methods (excluding SoftNet \cite{yoon2023soft}, which does not converge). As shown in Table~\ref{tab:cub200}, these pre-trained weights further enhance these methods. However, even with the same pre-trained weights, our NTK-FSCIL outperforms others. Notably, with pre-trained ConvNeXt-Base, it surpasses the next best method, ALICE \cite{peng2022few}, with improvements of 0.41\% in end-session accuracy and 0.50\% in PD.

\begin{table*}[htbp]
\centering
\setlength{\tabcolsep}{2.8mm}
\renewcommand{\arraystretch}{1.0}
\resizebox{0.9\linewidth}{!}{
    \begin{tabular}{ccccccccccccccc}
    \toprule
 \multirow{2}*{\textbf{ML}} & \multirow{2}*{\textbf{WC}} & \multirow{2}*{\textbf{SSI}} & \multirow{2}{*}{\textbf{LNR}} & \multirow{2}{*}{\textbf{CSR}} & \multicolumn{9}{c}{Acc. in each session (\%) ↑} & \multirow{2}{*}{$\triangle_{last}\uparrow$}\\
\cline{6-14} & & & & & 0     & 1     & 2     & 3     & 4     & 5     & 6     & 7     & 8 \\
    \hline
    \checkmark & & & & & 75.3 & 71.9 & 68.5 & 64.2 & 61.7 & 59.1 & 57.7 & 56.2 & 54.4 & +0.0 \\
    \checkmark & \checkmark & & & & 79.2 & 75.4 & 72.0 & 67.9 & 65.4 & 62.7 & 61.2 & 59.6 & 57.4 & +3.0 \\
    \checkmark & \checkmark & \checkmark & & & 82.9 & 78.7 & 74.5 & 70.5 & 67.8 & 65.1 & 63.7 & 61.8 & 59.0 & +4.6 \\
    \checkmark & \checkmark & \checkmark & \checkmark & & 83.3 & 78.7 & 75.2 & 71.0 & 67.7 & 65.1 & 63.5 & 61.6 & 59.5 & +5.1 \\
    \checkmark & \checkmark & \checkmark & \checkmark & \checkmark & 84.4 & 79.9 & 76.6 & 72.3 & 69.6 & 66.7 & 65.0 & 63.1 & 61.0 & +6.6 \\
    \bottomrule
    \end{tabular}
}
\caption{Ablation studies on \textbf{CIFAR100}. Specifically, \textbf{ML}, \textbf{WC}, \textbf{SSI}, \textbf{LNR}, and \textbf{CSR} represent Meta-Learning in Subsec.\ref{meta-learning}, Wider ConvNets in Subsec.\ref{wider convnets}, Self-Supervised Initialization in Subsec.\ref{pretrain}, Linear NTK Regularization, and Convolutional Spectral Regularization in Subsec.\ref{regularization}, respectively. The $\triangle_{last}\uparrow$ denotes the improvement over the first row's scenario.}
\label{ablation study}
\end{table*}

\begin{table*}[htbp]
\centering
\setlength{\tabcolsep}{2.8mm}
\renewcommand{\arraystretch}{1.0}
\resizebox{0.9\linewidth}{!}{
    \begin{tabular}{cccccccccccccccc}
    \toprule
 \multirow{2}*{\textbf{PW}} & \multirow{2}*{\textbf{ML}} & \multirow{2}*{\textbf{LNR}} & \multirow{2}{*}{\textbf{CSR}} & \multicolumn{11}{c}{Acc. in each session (\%) ↑} & \multirow{2}{*}{$\triangle_{last}\uparrow$}\\
\cline{5-15} & & & & 0     & 1     & 2     & 3     & 4     & 5     & 6     & 7     & 8     & 9     & 10 \\
    \hline
     & & & & 5.8 & 5.4 & 5.1 & 4.8 & 4.4 & 4.1 & 3.8 & 3.5 & 3.4 & 3.2 & 3.1 & +0.0 \\
    \checkmark & & & & 67.1 & 64.0 & 61.0 & 57.7 & 55.8 & 53.6 & 52.0 & 50.3 & 48.5 & 48.4 & 47.8 & +0.0 \\
    \checkmark & \checkmark & & & 77.7 & 74.7 & 72.0 & 68.4 & 66.6 & 64.0 & 62.8 & 61.6 & 59.9 & 59.8 & 59.2 & +11.4 \\
    \checkmark & \checkmark & \checkmark & & 77.8 & 74.6 & 72.0 & 68.4 & 66.6 & 63.9 & 62.8 & 61.8 & 60.5 & 60.4 & 59.9 & +12.1 \\
    \checkmark & \checkmark & \checkmark & \checkmark & 78.1 & 75.0 & 72.0 & 68.6 & 67.1 & 64.6 & 63.6 & 62.7 & 61.1 & 61.0 & 60.4 & +12.6 \\
    \bottomrule
    \end{tabular}
}
\caption{Ablation studies on \textbf{CUB200}. Specifically, \textbf{PW}, \textbf{ML}, \textbf{LNR} and \textbf{CSR} represent Pre-Training Weight, Meta-Learning in \cref{meta-learning}, Linear NTK Regularization and Convolutional Spectral Regularization in \cref{regularization}, respectively. The $\triangle_{last}\uparrow$ denotes the improvement over the second row's scenario.}
\label{ablation study cub}
\end{table*}

\begin{figure*}[t]
    \centering
    \begin{subfigure}[b]{0.54\columnwidth}
        \centering
        \includegraphics[width=\columnwidth]{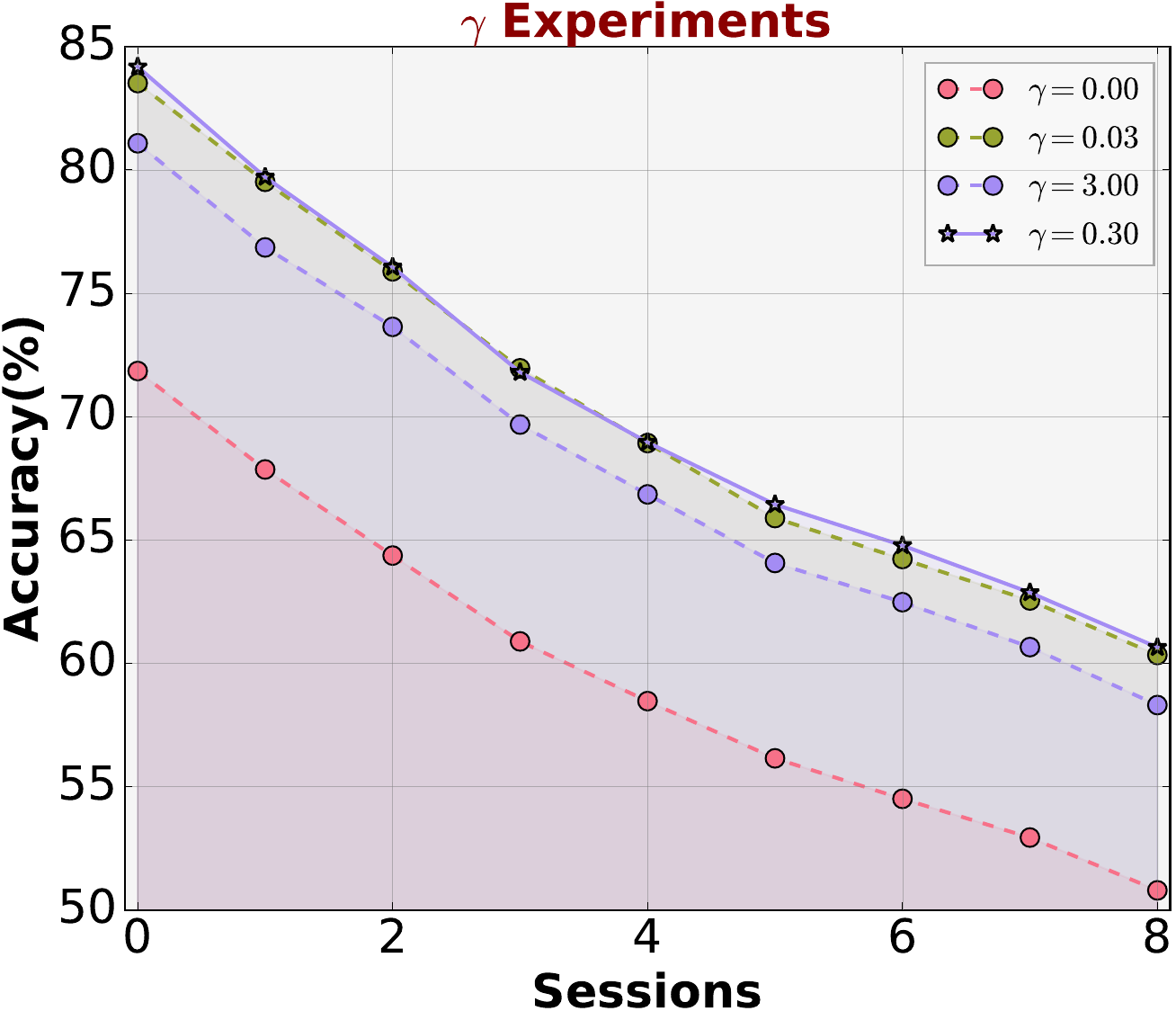}
        \label{supp:meta_cifar100_gamma}
    \end{subfigure}
    \hspace{2ex}
    \begin{subfigure}[b]{0.54\columnwidth}
        \centering
        \includegraphics[width=\columnwidth]{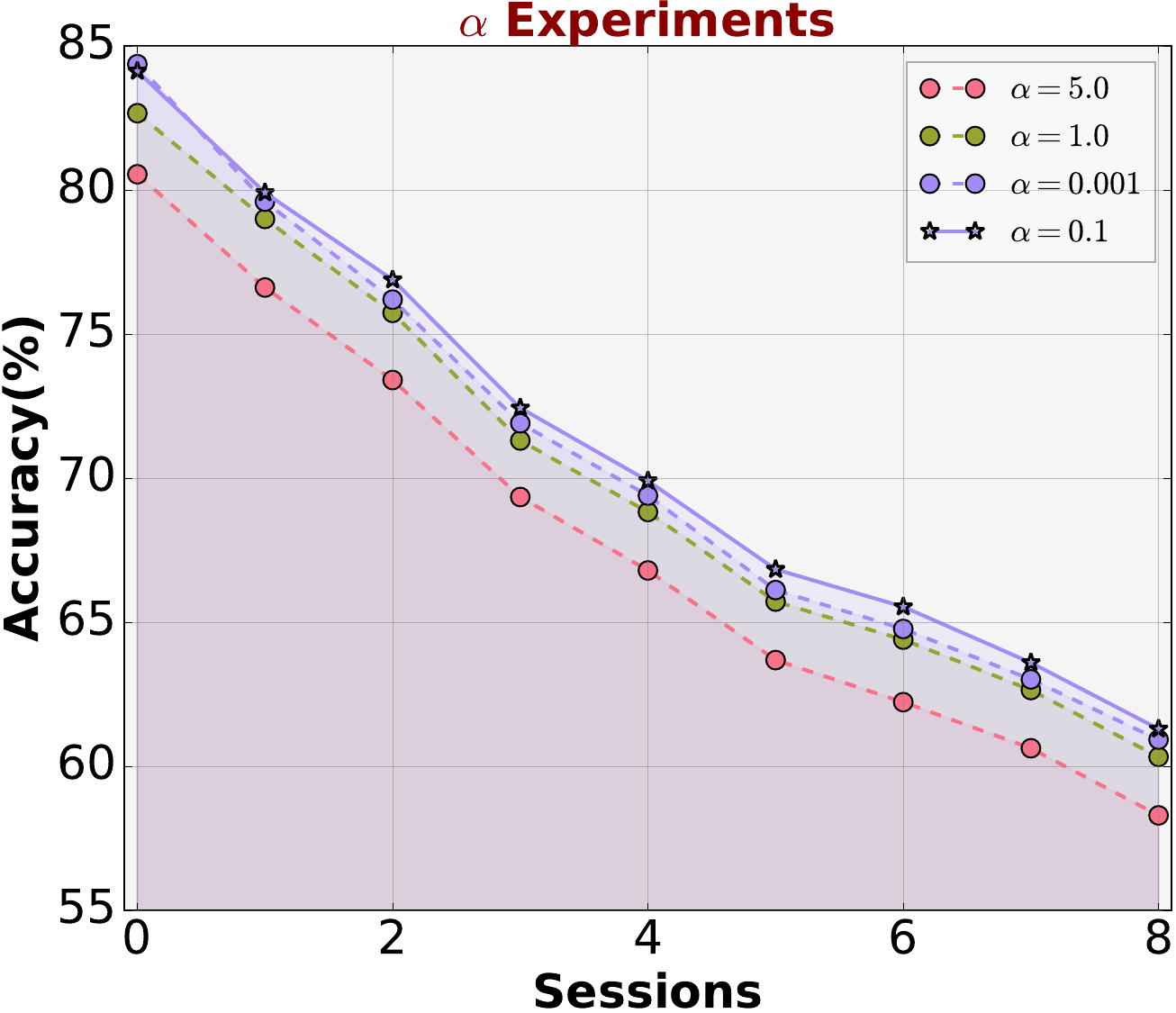}
        \label{supp:meta_cifar100_alpha}
    \end{subfigure}
    \hspace{2ex}
    \begin{subfigure}[b]{0.54\columnwidth}
        \centering
        \includegraphics[width=\columnwidth]{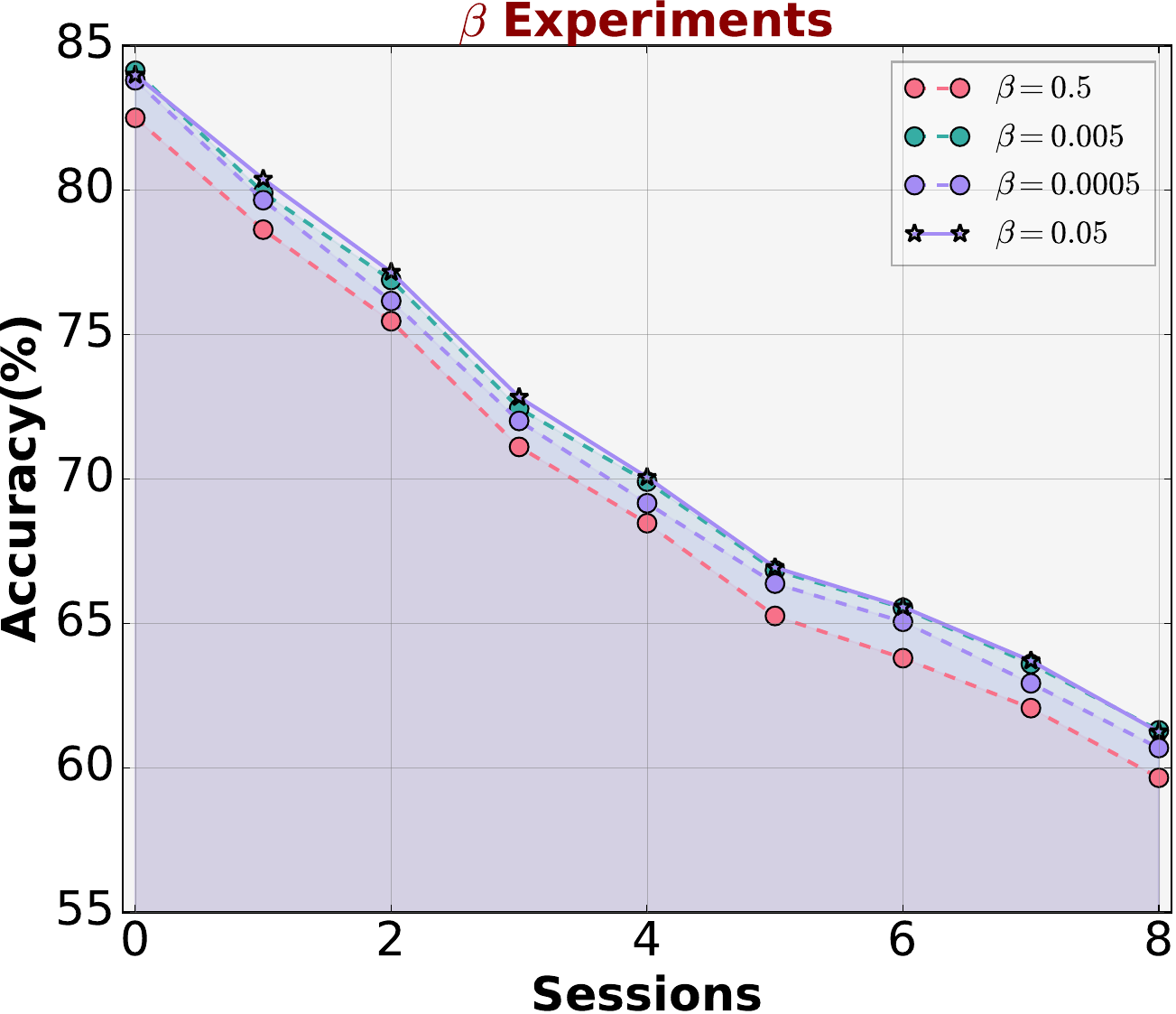}
        \label{supp:meta_cifar100_beta}
    \end{subfigure}
    \hspace{2ex}
    \begin{subfigure}[b]{0.54\columnwidth}
        \centering
        \includegraphics[width=\columnwidth]{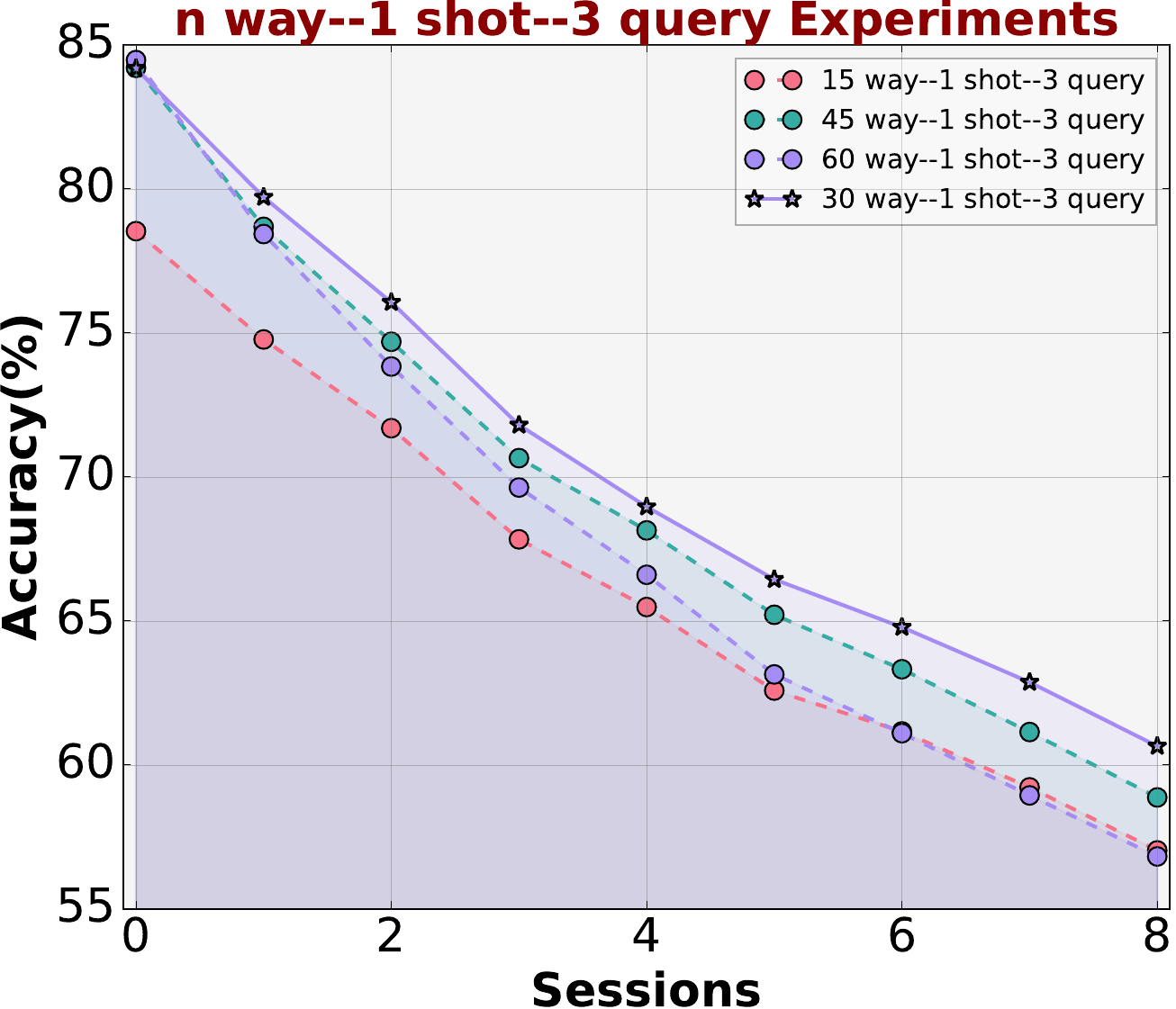}
        \label{supp:meta_cifar100_n}
    \end{subfigure}
    \hspace{2ex}
    \begin{subfigure}[b]{0.54\columnwidth}
        \centering
        \includegraphics[width=\columnwidth]{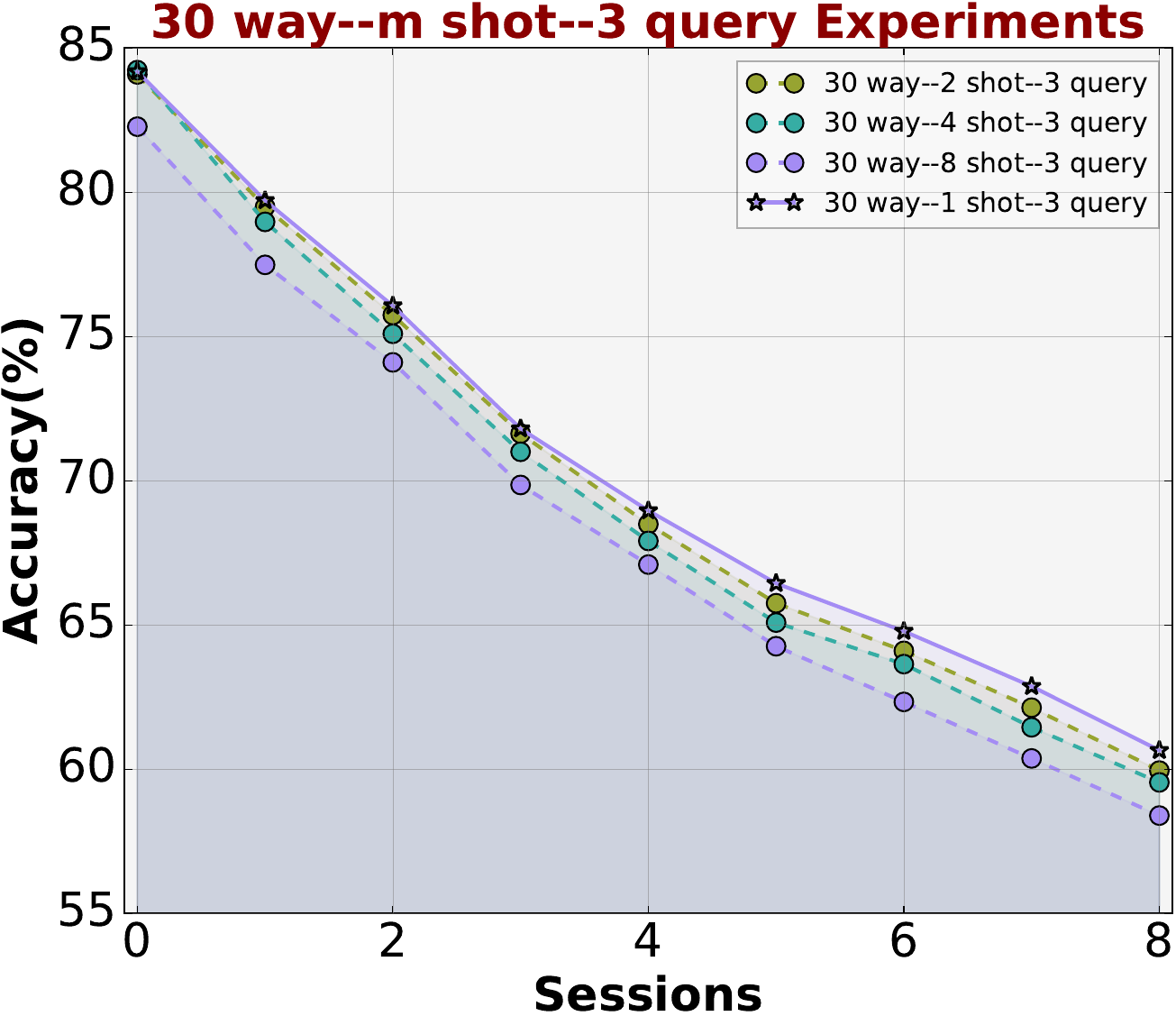}
        \label{supp:meta_cifar100_m}
    \end{subfigure}
    \hspace{2ex}
    \begin{subfigure}[b]{0.54\columnwidth}
        \centering
        \includegraphics[width=\columnwidth]{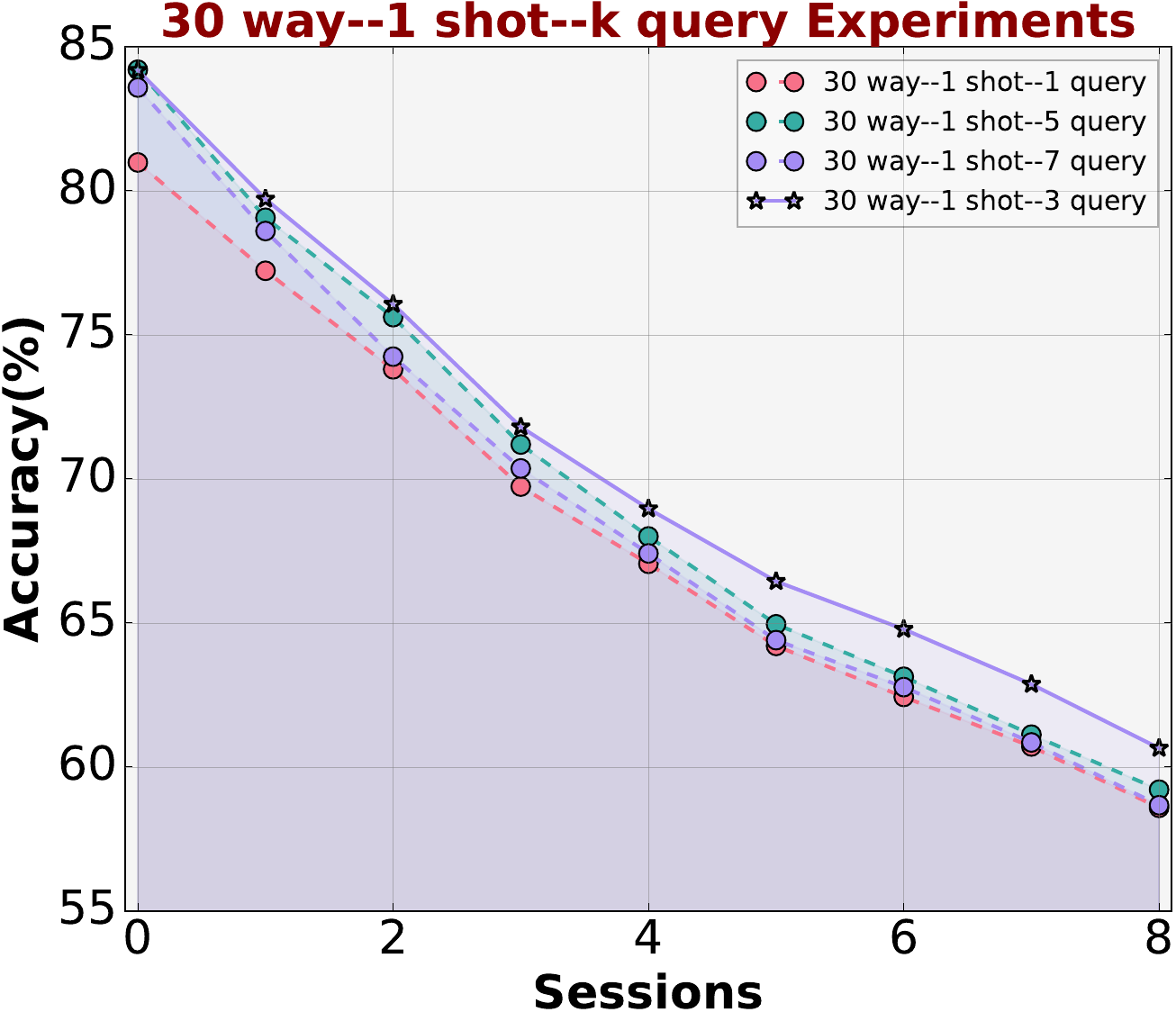}
        \label{supp:meta_cifar100_k}
    \end{subfigure}
    \caption{Top-1 accuracy for base and incremental sessions, assessed under diverse parameters on \textbf{CIFAR100}.}
    \label{supp:meta_results_ciffar100}
\end{figure*}

\begin{figure*}[t]
    \centering
    \begin{subfigure}[b]{0.54\columnwidth}
        \centering
        \includegraphics[width=\columnwidth]{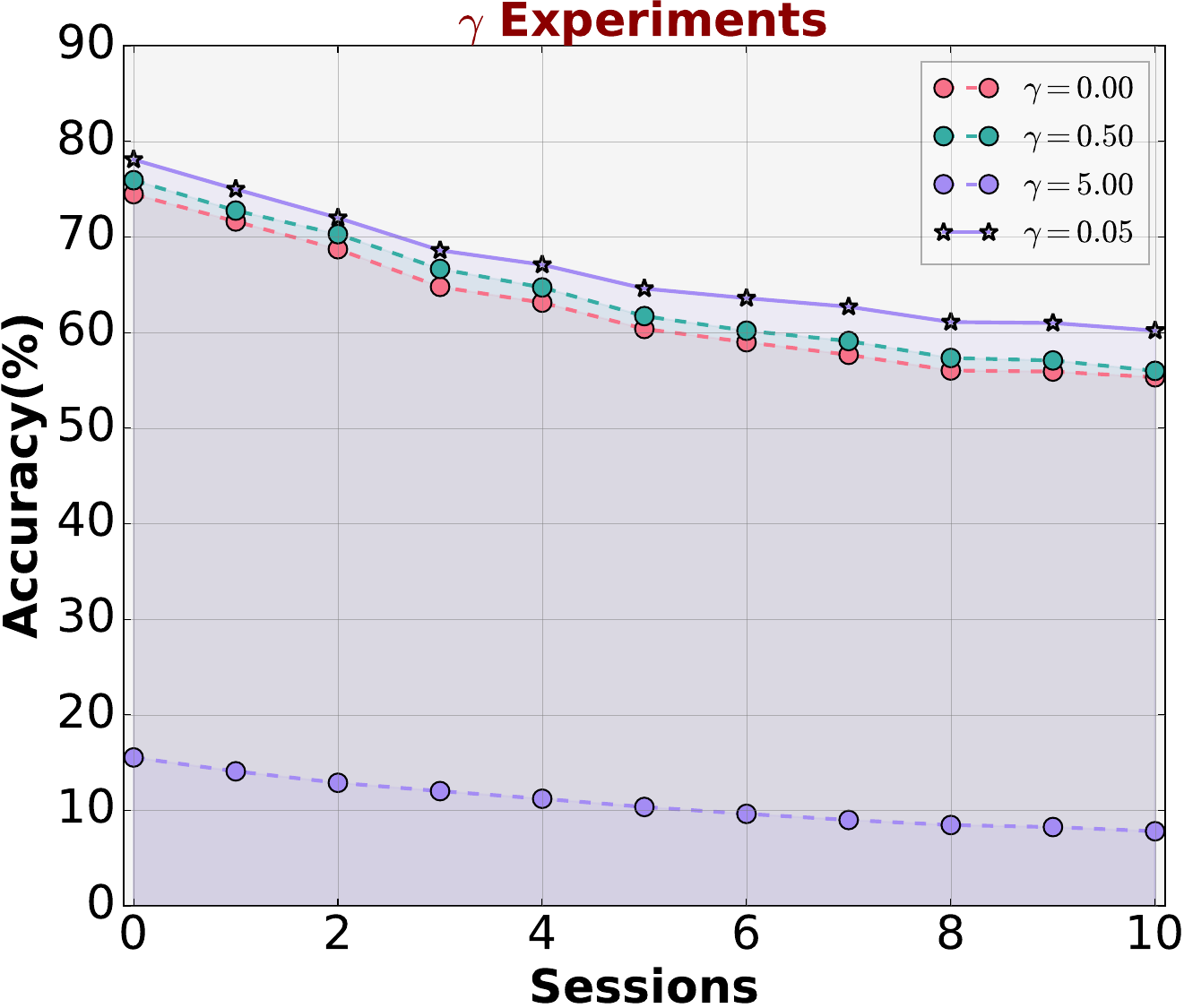}
        \label{supp:meta_cub200_gamma}
    \end{subfigure}
    \hspace{2ex}
    \begin{subfigure}[b]{0.54\columnwidth}
        \centering
        \includegraphics[width=\columnwidth]{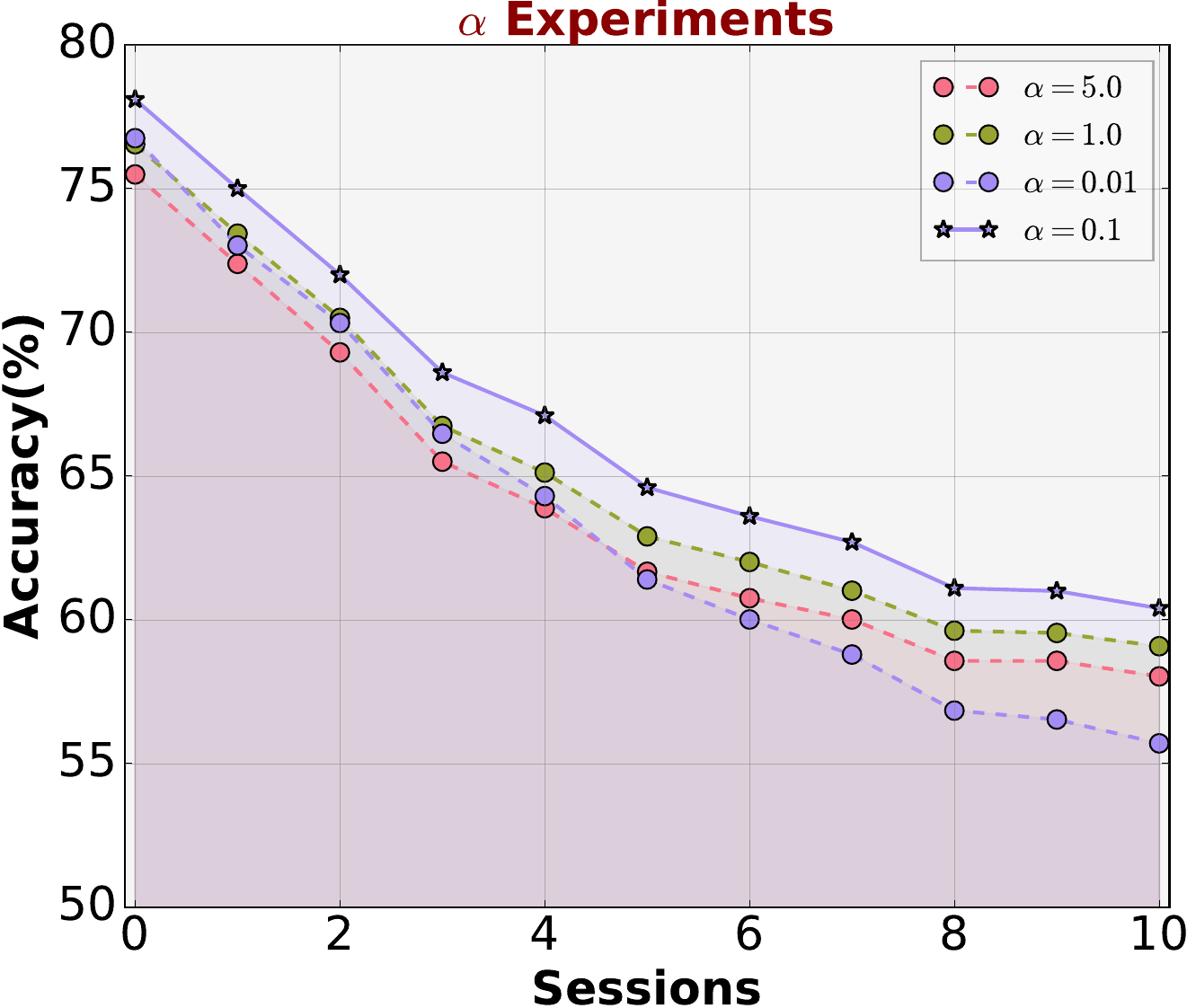}
        \label{supp:meta_cub200_alpha}
    \end{subfigure}
    \hspace{2ex}
    \begin{subfigure}[b]{0.54\columnwidth}
        \centering
        \includegraphics[width=\columnwidth]{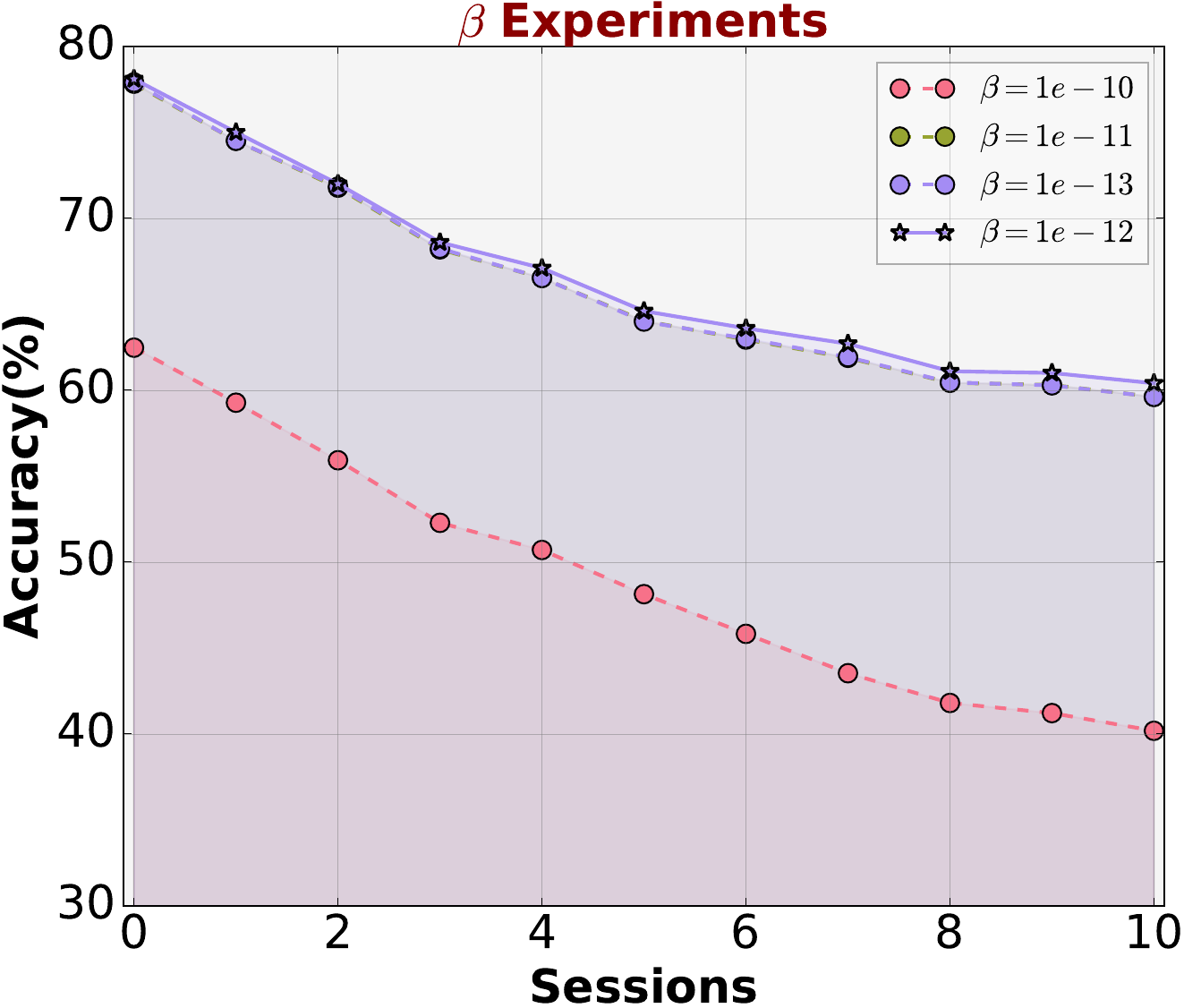}
        \label{supp:meta_cub200_beta}
    \end{subfigure}
    \begin{subfigure}[b]{0.54\columnwidth}
        \centering
        \includegraphics[width=\columnwidth]{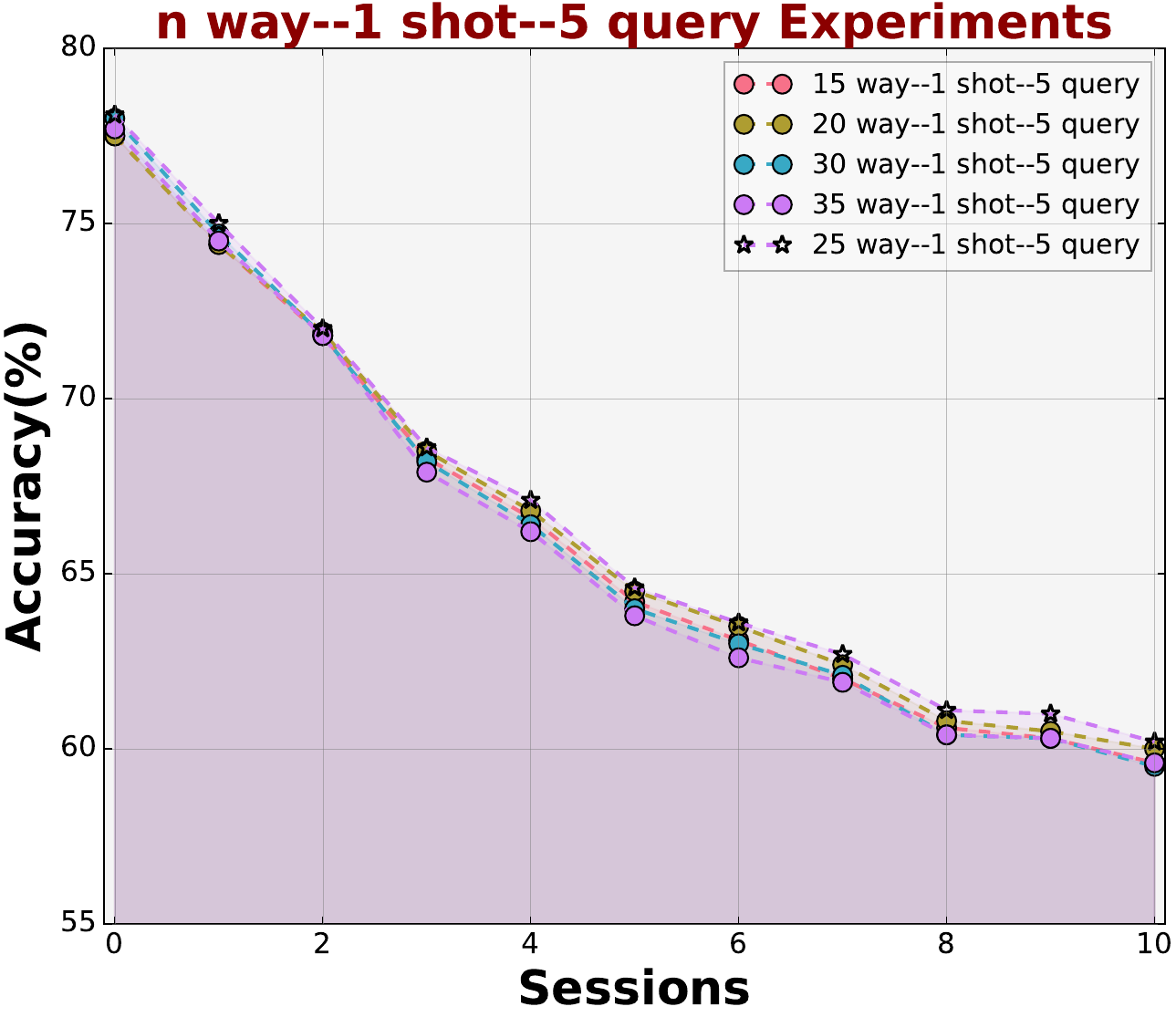}
        \label{supp:meta_cub200_n}
    \end{subfigure}
    \hspace{2ex}
    \begin{subfigure}[b]{0.54\columnwidth}
        \centering
        \includegraphics[width=\columnwidth]{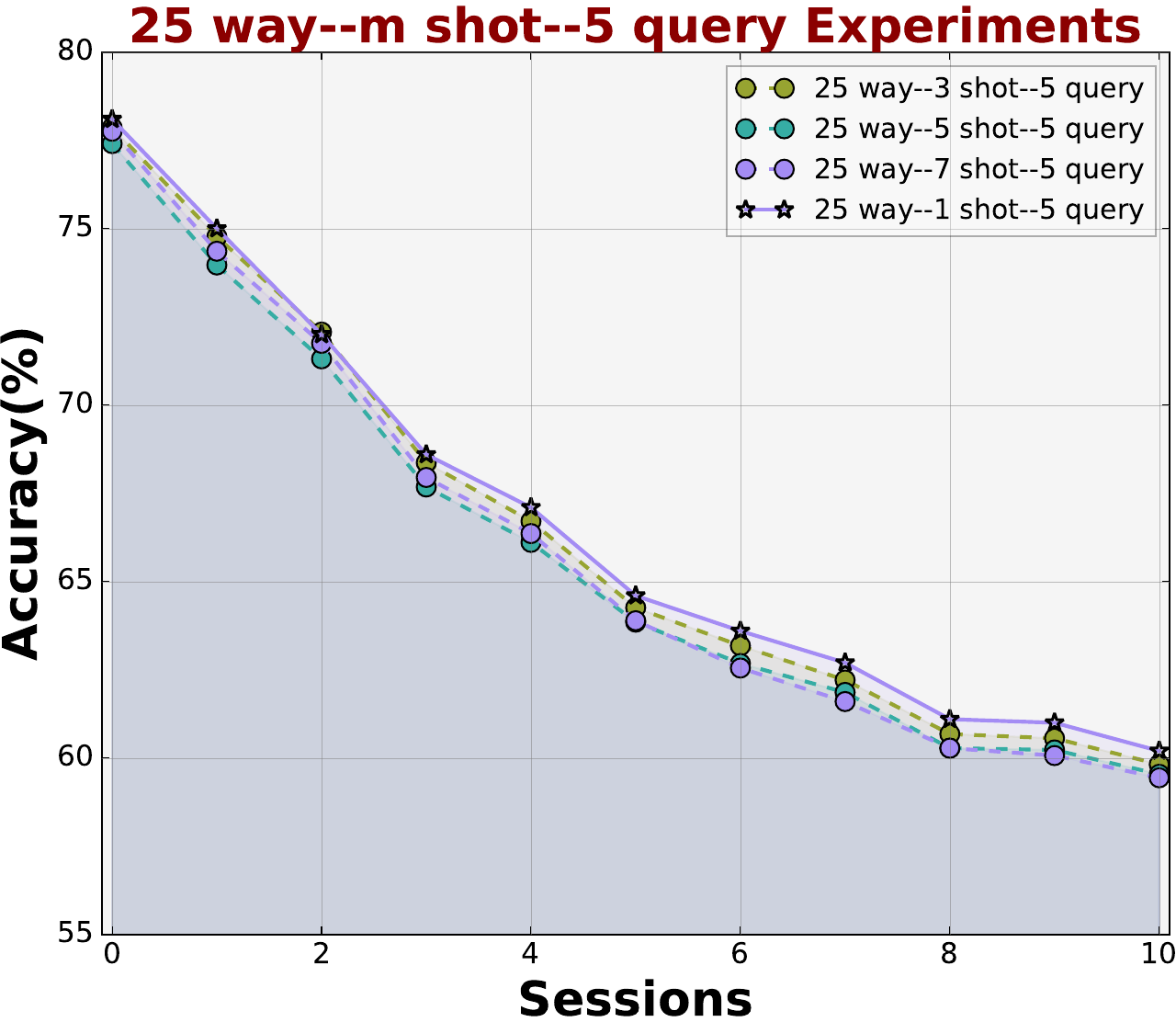}
        \label{supp:meta_cub200_m}
    \end{subfigure}
    \hspace{2ex}
    \begin{subfigure}[b]{0.54\columnwidth}
        \centering
        \includegraphics[width=\columnwidth]{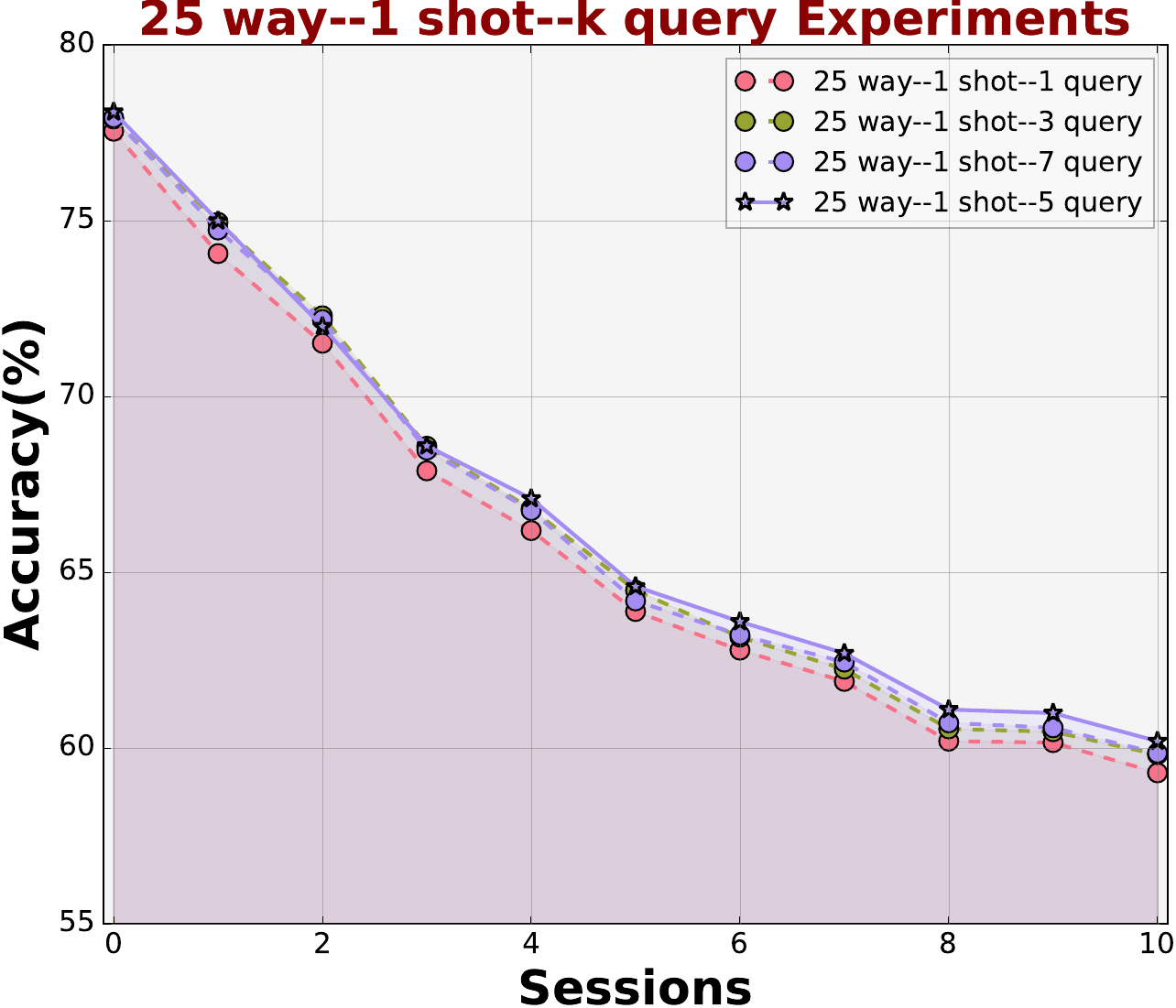}
        \label{supp:meta_cub200_k}
    \end{subfigure}
    \caption{Top-1 accuracy for base and incremental sessions, assessed under diverse parameters on \textbf{CUB200}.}
    \label{supp:meta_results_cub200}
\end{figure*}

\begin{figure}[ht]
\centering
\includegraphics[width=0.38\textwidth]{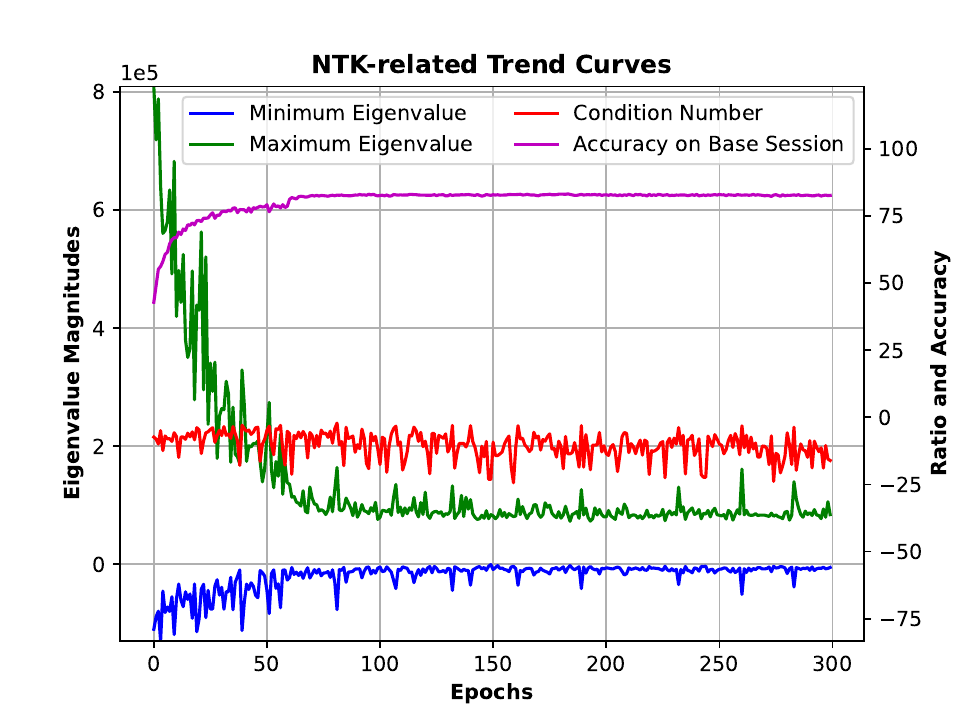}
\caption{NTK-related metrics and accuracy on the base session.}
\label{fig:NTK values}
\end{figure}

\begin{figure*}[htbp]
    \centering
    \begin{subfigure}{0.31\textwidth}
        \includegraphics[width=\linewidth]{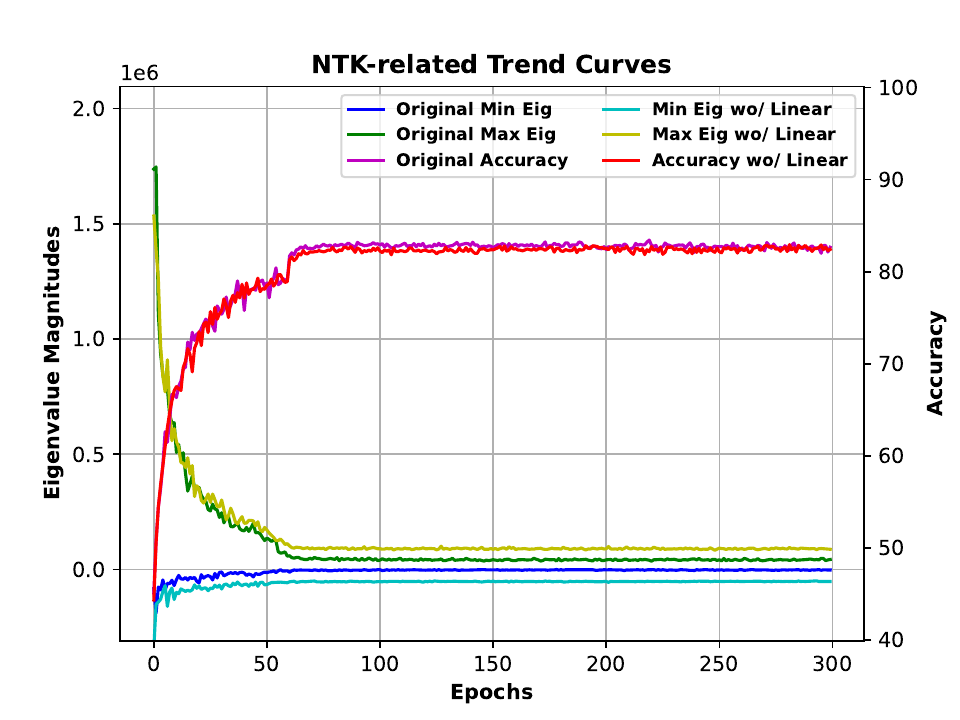}
        \caption{\footnotesize Linear NTK Regularization}
    \label{NTK_without_linear}
    \end{subfigure}
    \hfill
    \begin{subfigure}{0.31\textwidth}
        \includegraphics[width=\linewidth]{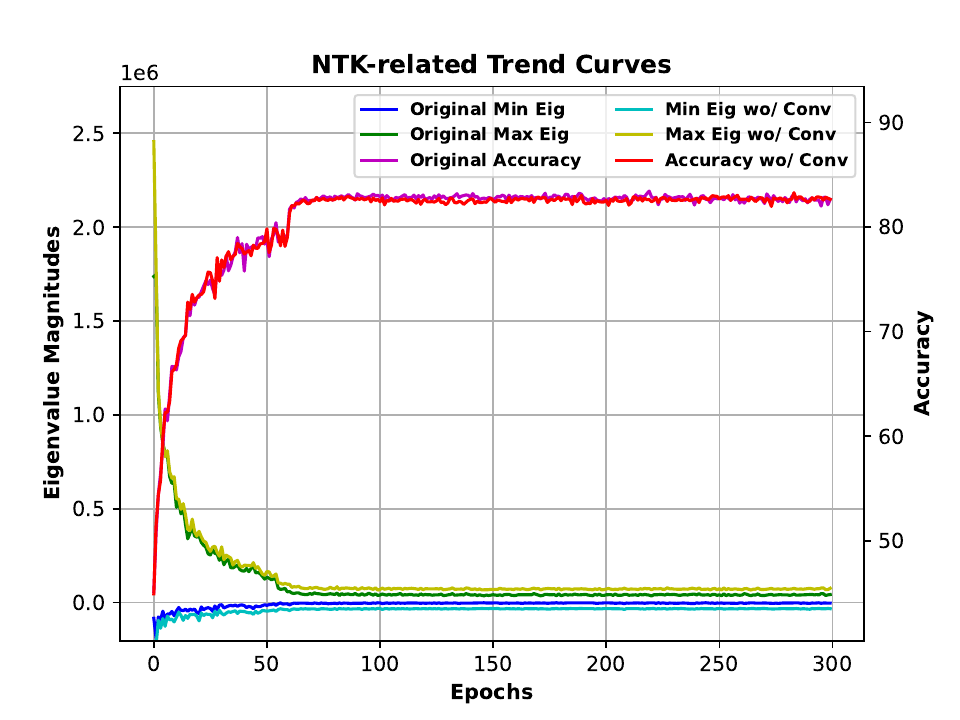}
        \caption{\footnotesize Convolutional NTK Regularization}
    \label{NTK_without_conv}
    \end{subfigure}
    \hfill
    \begin{subfigure}{0.31\textwidth}
        \includegraphics[width=\linewidth]{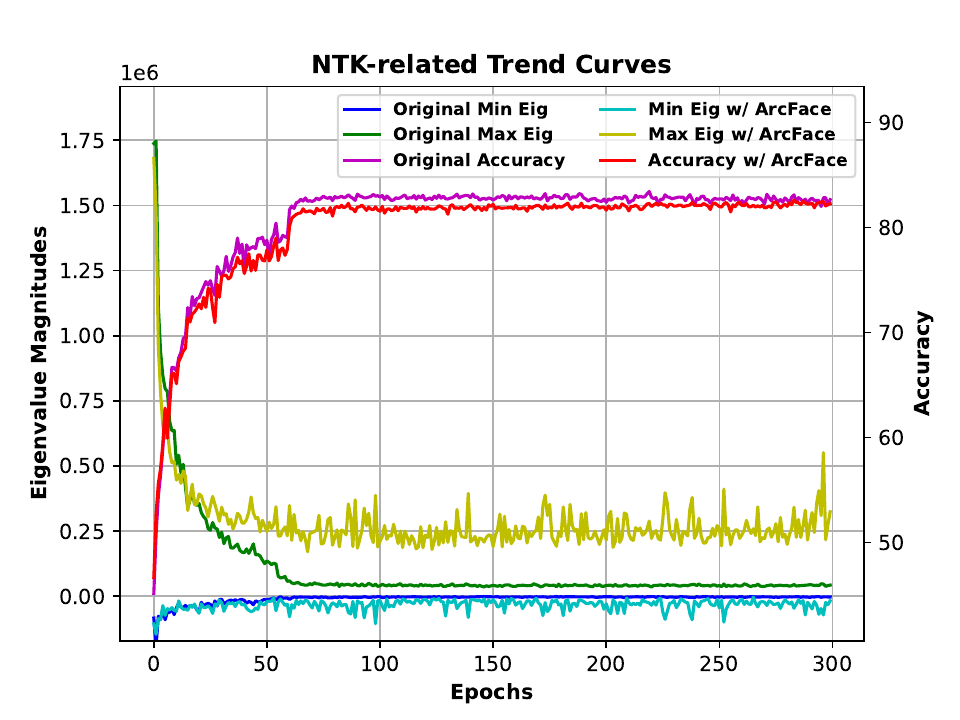}
        \caption{\footnotesize Curricular Alignment}
    \label{NTK_with_arcface}
    \end{subfigure}
    
    \begin{subfigure}{0.31\textwidth}
        \includegraphics[width=\linewidth]{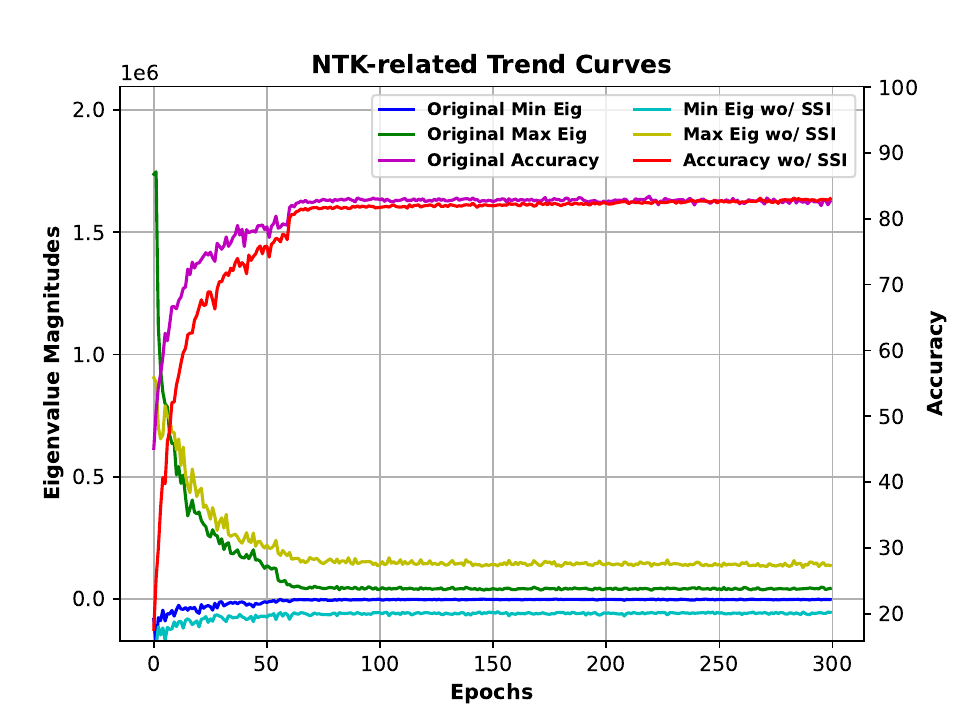}
        \caption{\footnotesize Self-supervised Pre-training}
    \label{NTK_without_SSI}
    \end{subfigure}
    \hfill
    \begin{subfigure}{0.31\textwidth}
        \includegraphics[width=\linewidth]{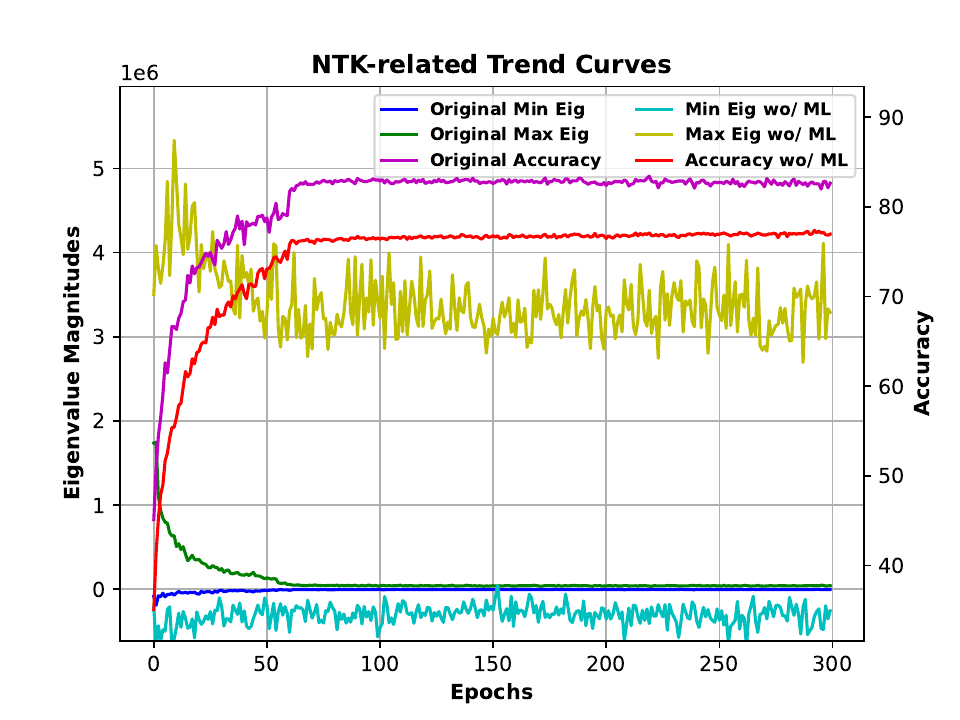}
        \caption{\footnotesize Meta-Learning}
    \label{NTK_without_ml}
    \end{subfigure}
    \hfill
    \begin{subfigure}{0.31\textwidth}
        \includegraphics[width=\linewidth]{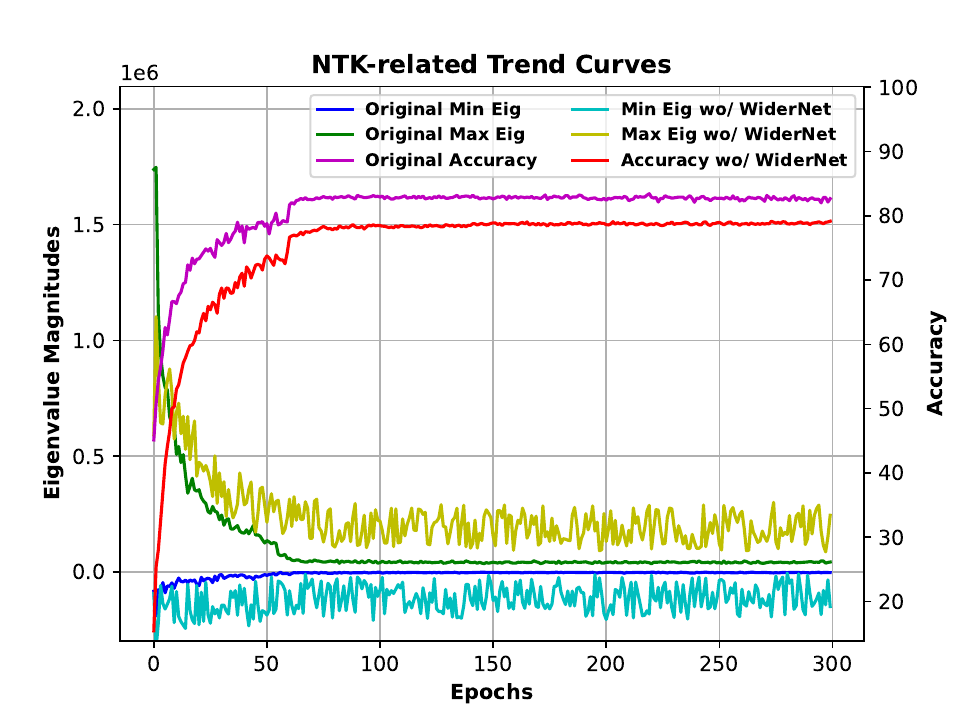}
        \caption{\footnotesize Wider ConvNet}
    \label{NTK_without_widernet}
    \end{subfigure}
    \caption{Investigate the NTK dynamics post-elimination of Linear NTK Regularization, Convolutional NTK Regularization, Curricular Alignment, Self-supervised Pre-training, Meta-Learning and Wider ConvNet strategies.}
    \label{supp:all_ntk}
\end{figure*}

\subsection{Ablation Study}
To demonstrate the effectiveness of NTK-FSCIL's components, we conduct ablation experiments on CIFAR100 and CUB200, detailed in Table~\ref{ablation study} and Table~\ref{ablation study cub}. Each experiment utilizes curricular alignment, the details of which are outlined in \cref{alignment}. For additional insights into alignment strategies, refer to \cref{supp:classification_loss}. As shown in Table~\ref{ablation study}, Table~\ref{ablation study cub}, \cref{wider convnets}, \cref{supp:self-supervised pre-training}, \cref{supp:classification_loss}, and \cref{supp:mix_ups}, the components of our designed NTK-FSCIL demonstrate exceptional performance on CIFAR100 and CUB200. On CUB200, our meticulously designed meta-learning strategy achieves a 23.8\% improvement in end-session accuracy compared to the traditional supervised paradigm. On CIFAR100, the network width expansion strategy yields an approximate 5.5\% improvement in end-session accuracy, with self-supervised pre-training further enhancing this by 2.8\%. Additionally, implementing linear and convolutional NTK regularization provides positive gains on both CIFAR100 and CUB200, with improvements of approximately 3.4\% and 2.0\%, respectively, highlighting the importance of regularization in continual learning scenarios \cite{sun2023regularizing}. In summary, these representative ablation studies have demonstrated that our components are critically important for enhancing performance in FSCIL scenarios, both theoretically and practically.

\subsection{Hyperparameters Tuning}
In this subsection, to identify the optimal values for $\gamma$, $\alpha$, $\beta$, $n$, $m$, and $k$ within our NTK-FSCIL framework, we conduct detailed tuning of these hyperparameters. The fine-tuning is primarily performed on the CIFAR100 and CUB200 datasets. The parameters for CIFAR100 are shared with \emph{mini}ImageNet and ImageNet100. Detailed results are presented in \cref{supp:meta_results_ciffar100} and \cref{supp:meta_results_cub200}. According to the statistical data, the optimal hyperparameters for CIFAR100 are $\gamma$=0.30, $\alpha$=0.1, $\beta$=0.05, $n$=30, $m$=1, and $k$=3, while the optimal hyperparameters for CUB200 are $\gamma$=0.05, $\alpha$=0.1, $\beta$=1e-12, $n$=25, $m$=1, and $k$=5. It is particularly noteworthy that for CUB200, which uses a pre-trained model, $\beta$ must be set to an extremely small value to suppress the loss associated with Linear NTK Regularization, preventing numerical imbalance.

\subsection{NTK Contributions} \label{supp:NTK_properties}
To measure model generalization during optimization, we randomly sample images from both base and incremental sessions. For GPU efficiency, we only compute the NTK matrix solely for the final nested linear layers.
This allows us to scrutinize the linear operations, thereby providing insight into how the ultimate linear layers contribute to the model generalization. Following the construction of the NTK matrix, we monitor them in real time by evaluating the evolution of eigenvalues and the accuracy of the base session. The condition number of the NTK matrix is then computed as an indicator of its stability, in a manner analogous to \cref{eq:linear_ntk3}, defined as the maximum-to-minimum eigenvalue ratio. 

The detailed evolutionary trajectories, including changes in the minimum and maximum eigenvalues, condition number, and accuracy, are illustrated in \cref{fig:NTK values}.
Interestingly, we observe a strong positive correlation between the eigenvalues of the NTK matrix and the accuracy. The NTK's eigenvalues gradually converge, maintaining a state of equilibrium, mirroring the trend of ascending and then stabilizing accuracy. Moreover, the condition number remains relatively stable, illustrating the gradual convergence and stability of the NTK matrix, thereby facilitating improved generalization.

To probe the impact of widening ConvNet, meta-learning, self-supervised pre-training, curricular alignment and dual NTK regularization on NTK properties, we conduct a comprehensive analysis on CIFAR100. We systematically omit each module one by one to observe the resulting shifts in the NTK's eigenvalue distribution, as depicted in \cref{supp:all_ntk}.

From \cref{supp:all_ntk}\subref{NTK_without_linear}, \cref{supp:all_ntk}\subref{NTK_without_conv}, \cref{supp:all_ntk}\subref{NTK_with_arcface} and \cref{supp:all_ntk}\subref{NTK_without_SSI}, it's evident that Linear NTK Regularization, Convolutional NTK Regularization, Curricular Alignment and Self-supervised Pre-training play vital roles in ensuring NTK stability during optimization. Omitting or altering any of these components leads to a markedly erratic NTK's eigenvalue distribution, characterized by significant variations between the highest and lowest eigenvalues. Such instability negatively impacts the base session accuracy and hampers model generalization. Similarly, as shown in \cref{supp:all_ntk}\subref{NTK_without_ml} and \cref{supp:all_ntk}\subref{NTK_without_widernet}, Meta-Learning and a Wider ConvNet are essential for NTK convergence. Their absence results in unpredictable fluctuations in the NTK eigenvalue distribution, with the maximum and minimum eigenvalues lacking a consistent path during optimization, further affecting model generalization.

In summary, each component within our NTK-FSCIL plays a critical role in ensuring the NTK's convergence and stability during optimization. Their combined efforts enable our FSCIL learner to exhibit strong generalization capabilities at the NTK theoretical level, leading to impressive performance in incremental sessions. Consequently, it not only improves overall FSCIL performance but also effectively mitigates the catastrophic forgetting problem.

\section{Exploring Pre-training Weights}\label{supp:ntk_pre_training}
As highlighted by \cite{wei2022more}, utilizing extensive data during pre-training can significantly bolster a model's NTK stability and enhance its classification efficacy. In this subsection, we critically evaluate the efficacy of our NTK-FSCIL framework across varied pre-trained weights, with a focus on the CUB200 dataset. As inferred from Table~\ref{pre-trained_weights_CUB200}, when there is no image scale discrepancy between pre-training and FSCIL training, most pre-trained weights significantly enhance FSCIL performance. Among these, the ConvNeXt series achieves the best performance among ConvNets, while the Swin Transformer series stands out among Transformer-based architectures. Both significantly outperform current FSCIL methods on CUB200.

A noteworthy observation pertains to the CLIP series' visual encoders, which benefit from training on colossal datasets. Contrary to expectations, their performance does not consistently exceed those trained solely on ImageNet-1K and requires a significantly lower learning rate for optimal convergence. For example, while standard FSCIL approaches on CUB200 typically operate best at a learning rate of 5e-4, leveraging CLIP pre-trained weights necessitates a learning rate of less than 1e-6. This discrepancy may stem from CLIP's multimodal training, where using a single modality might result in information attenuation. Additionally, the expansive training data could render the visual feature space more compact, requiring a tempered learning rate for optimal convergence. Results from both CLIP-RN50 and CLIP-RN50$\times$4 reinforce the notion that widening the model architecture can stabilize NTK's eigenvalue distribution, enhance model generalization, and effectively improve its anti-amnesia capability.

Furthermore, we explore the performance of self-supervised pre-training weights, as shown in Table~\ref{self-pre-trained_weights_CUB200}. The current self-supervised pre-trained weights, despite being supported by extensive training data, still fall short of their supervised counterparts in overall FSCIL performance. However, these self-supervised models exhibit notable strengths. Specifically, in terms of the PD metric, the weights from self-supervised pre-training, particularly MoCo-v3, demonstrate competitive performance, matching or even surpassing some supervised methods. This suggests that self-supervised techniques have significant untapped potential and could evolve to match or exceed the performance of supervised pre-training in the FSCIL context.

\begin{table*}[!t]
    \centering
    \setlength{\tabcolsep}{2.8mm}
    \renewcommand{\arraystretch}{1.0}
    \resizebox{0.8\linewidth}{!}{
        \begin{tabular}{cccccccccccccc}
            \toprule
            \multirow{2}{*}{Backbone} & \multicolumn{12}{c}{Acc. in each session (\%) ↑} & \multicolumn{1}{l}{\multirow{2}{*}{PD $\downarrow$}} \\ \cline{2-13} & Dataset & 0 & 1 & 2 & 3 & 4 & 5 & 6 & 7 & 8 & 9 & 10
            & \multicolumn{1}{l}{} \\ \hline
            ResNet-18 & ImageNet-1K & 78.1 & 75.0 & 72.0 & 68.6 & 67.1 & 64.6 & 63.6 & 62.7 & 61.1 & 61.0 & 60.2 & 17.9 \\
            ResNet-50 & ImageNet-1K & 80.2 & 77.3 & 74.9 & 71.2 & 70.3 & 67.7 & 66.1 & 65.2 & 63.5 & 63.3 & 62.9 & 17.3 \\
            ResNet-50 $\times$ 2 & ImageNet-1K & 82.9 & 80.2 & 78.5 & 75.2 & 73.3 & 70.9 & 69.7 & 68.8 & 67.4 & 67.3 & 66.8 & 16.1 \\
            ResNet-101 & ImageNet-1K & 83.1 & 80.8 & 78.7 & 75.1 & 73.5 & 71.2 & 69.8 & 68.6 & 66.7 & 66.6 & 66.1 & 17.0 \\
            ResNet-101 $\times$ 2 & ImageNet-1K & 83.1 & 80.3 & 78.1 & 75.2 & 73.7 & 71.2 & 70.5 & 69.7 &  67.8 & 68.1 & 67.8 & 12.5 \\
            ConvNeXt-Tiny & ImageNet-1K & 81.5 & 79.0 & 77.1 & 74.2 & 72.8 & 70.6 & 69.6 & 69.0 & 67.6 & 67.3 & 67.0 & 14.5 \\
            ConvNeXt-Small & ImageNet-1K & 83.5 & 81.0 & \textcolor[rgb]{0,0,1}{\textbf{79.5}} & 77.3 & 76.5 & \textcolor[rgb]{0,0,1}{\textbf{74.6}} & \textcolor[rgb]{0,0,1}{\textbf{74.1}} & 73.9 & 73.2 & 73.0 & 72.9 & 10.6 \\
            ConvNeXt-Base & ImageNet-1K & 83.3 & \textcolor[rgb]{0,0,1}{\textbf{81.3}} & \textcolor[rgb]{1,0,0}{\textbf{80.1}} & \textcolor[rgb]{1,0,0}{\textbf{78.0}} & \textcolor[rgb]{1,0,0}{\textbf{77.2}} & \textcolor[rgb]{1,0,0}{\textbf{74.9}} & \textcolor[rgb]{1,0,0}{\textbf{74.6}} & \textcolor[rgb]{1,0,0}{\textbf{74.5}} & \textcolor[rgb]{1,0,0}{\textbf{74.3}} & \textcolor[rgb]{1,0,0}{\textbf{74.2}} & \textcolor[rgb]{1,0,0}{\textbf{74.0}} & \textcolor[rgb]{1,0,0}{\textbf{9.3}} \\
            ViT-Base16 & ImageNet-1K & 81.4 & 78.6 & 76.3 & 73.1 & 72.1 & 69.9 & 68.8 & 68.4 & 67.5 & 67.3 & 67.2 & 14.2 \\
            ViT-Base32 & ImageNet-1K & 75.9 & 72.9 & 70.1 & 67.3 & 65.7 & 63.4 & 62.6 & 61.8 & 60.9 & 60.7 & 60.6 & 15.3 \\
            Swin-Tiny & ImageNet-1K & 82.8 & 80.5 & 78.5 & 76.5 & 75.5 & 73.6 & 73.3 & 73.2 & 72.6 & 72.4 & 71.8 & 11.0 \\
            Swin-Small & ImageNet-1K & \textcolor[rgb]{0,0,1}{\textbf{84.0}} & \textcolor[rgb]{1,0,0}{\textbf{81.8}} & \textcolor[rgb]{1,0,0}{\textbf{80.1}} & \textcolor[rgb]{0,0,1}{\textbf{77.4}} & \textcolor[rgb]{0,0,1}{\textbf{76.6}} & \textcolor[rgb]{0,0,1}{\textbf{74.6}} & \textcolor[rgb]{1,0,0}{\textbf{74.6}} & \textcolor[rgb]{0,0,1}{\textbf{74.4}} & \textcolor[rgb]{0,0,1}{\textbf{74.0}} & \textcolor[rgb]{0,0,1}{\textbf{73.9}} & \textcolor[rgb]{0,0,1}{\textbf{73.6}} & \textcolor[rgb]{0,0,1}{\textbf{10.4}} \\
            Swin-Base & ImageNet-1K & 83.2 & 80.7 & 78.6 & 76.3 & 75.5 & 73.8 & 73.6 & 73.4 & 72.8 & 72.7 & 72.3 & 10.9 \\
            Swin-v2-Tiny & ImageNet-1K & 82.6 & 80.4 & 78.6 & 75.9 & 74.8 & 72.8 & 72.3 & 72.2 & 71.2 & 71.1 & 71.0 & 11.6 \\
            Swin-v2-Small & ImageNet-1K & 82.8 & 80.4 & 78.7 & 75.6 & 74.5 & 72.6 & 72.1 & 71.9 & 70.6 & 70.4 & 70.3 & 12.5 \\
            Swin-v2-Base & ImageNet-1K & 81.8 & 79.8 & 78.1 & 75.5 & 74.6 & 72.4 & 71.7 & 71.5 & 70.8 & 70.7 & 70.5 & 11.3 \\
            CLIP-RN50 & WIT & 74.2 & 66.6 & 63.0 & 58.5 & 56.8 & 54.1 & 52.0 & 51.1 & 49.0 & 48.3 & 47.4 & 26.8 \\
            CLIP-RN50$\times$4 & WIT & 83.3 & 80.2 & 77.7 & 73.9 & 72.7 & 70.4 & 69.5 & 68.0 & 66.0 & 65.9 & 65.1 & 18.2 \\
            CLIP-RN101 & WIT & 81.9 & 78.4 & 75.7 & 72.1 & 70.6 & 67.3 & 65.5 & 64.1 & 62.0 & 61.4 & 60.3 & 21.6 \\
            CLIP-ViT-Base16 & WIT & \textcolor[rgb]{1,0,0}{\textbf{84.9}} & 80.9 & 77.9 & 73.6 & 71.4 & 68.2 & 66.5 & 64.5 & 62.7 & 62.1 & 61.0 & 23.9 \\
            CLIP-ViT-Base32 & WIT & 79.9 & 76.0 & 72.2 & 67.8 & 65.0 & 61.9 & 60.0 & 58.5 & 56.4 & 55.5 & 54.3 & 25.6 \\
            \bottomrule
        \end{tabular}
    }
\caption{Effectiveness of pre-trained weights from various large datasets within our NTK-FSCIL framework for the \textbf{CUB200} dataset. Red-bolded segments represent the optimal results, while blue-bolded segments represent suboptimal results.}
\label{pre-trained_weights_CUB200}
\end{table*}

\begin{table*}[!t]
    \centering
    \setlength{\tabcolsep}{2.8mm}
    \renewcommand{\arraystretch}{1.0}
    \resizebox{0.8\linewidth}{!}{
        \begin{tabular}{cccccccccccccc}
            \toprule
            \multirow{2}{*}{Backbone} & \multicolumn{12}{c}{Acc. in each session (\%) ↑} & \multicolumn{1}{l}{\multirow{2}{*}{PD $\downarrow$}} \\ \cline{2-13} & Dataset & 0 & 1 & 2 & 3 & 4 & 5 & 6 & 7 & 8 & 9 & 10
            & \multicolumn{1}{l}{} \\ \hline
            ResNet-50+Dino & ImageNet-1K & \textcolor[rgb]{0,0,1}{\textbf{69.2}} & \textcolor[rgb]{0,0,1}{\textbf{67.5}} & \textcolor[rgb]{0,0,1}{\textbf{64.2}} & \textcolor[rgb]{0,0,1}{\textbf{61.5}} & \textcolor[rgb]{0,0,1}{\textbf{59.4}} & \textcolor[rgb]{0,0,1}{\textbf{57.0}} & \textcolor[rgb]{0,0,1}{\textbf{54.8}} & \textcolor[rgb]{0,0,1}{\textbf{53.2}} & \textcolor[rgb]{0,0,1}{\textbf{51.8}} & \textcolor[rgb]{0,0,1}{\textbf{51.6}} & \textcolor[rgb]{0,0,1}{\textbf{51.3}} & 17.9\\
            ResNet-50+BYOL & ImageNet-1K & 68.7 & 66.3 & 63.3 & 60.6 & 58.2 & 55.8 & 53.9 & 52.2 & 50.5 & 50.4 & 50.1 & 18.6 \\
            ResNet-50+MoCo-v3 & ImageNet-1K & \textcolor[rgb]{1,0,0}{\textbf{73.0}} & \textcolor[rgb]{1,0,0}{\textbf{70.3}} & \textcolor[rgb]{1,0,0}{\textbf{67.9}} & \textcolor[rgb]{1,0,0}{\textbf{65.3}} & \textcolor[rgb]{1,0,0}{\textbf{62.9}} & \textcolor[rgb]{1,0,0}{\textbf{60.3}} & \textcolor[rgb]{1,0,0}{\textbf{58.7}} & \textcolor[rgb]{1,0,0}{\textbf{57.1}} & \textcolor[rgb]{1,0,0}{\textbf{55.6}} & \textcolor[rgb]{1,0,0}{\textbf{55.5}} & \textcolor[rgb]{1,0,0}{\textbf{55.2}} & \textcolor[rgb]{0,0,1}{\textbf{17.8}} \\
            ResNet-50+SparK & ImageNet-1K & 52.6 & 49.0 & 46.1 & 43.1 & 41.3 & 39.1 & 37.2 & 35.6 & 34.0 & 33.5 & 32.6 & 20.0 \\
            ResNet-50+simCLR & ImageNet-1K & 43.1 & 41.0 & 38.2 & 36.0 & 34.0 & 32.5 & 30.6 & 29.0 & 27.8 & 27.3 & 26.9 & \textcolor[rgb]{1,0,0}{\textbf{16.2}}\\
            ResNet-50$\times$2+simCLR & ImageNet-1K & 48.7 & 46.6 & 43.4 & 41.3 & 39.1 & 37.0 & 35.2 & 33.2 & 31.5 & 31.1 & 30.5 & 18.2 \\
            ResNet-50$\times$4+simCLR & ImageNet-1K & 54.0 & 51.2 & 47.8 & 45.0 & 42.4 & 40.2 & 38.0 & 35.9 & 34.3 & 33.8 & 33.3 & 20.7 \\
            \bottomrule
        \end{tabular}
    }
\caption{Effectiveness of self-supervised pre-trained weights from large datasets within our NTK-FSCIL framework for \textbf{CUB200}. Red-bolded segments represent optimal results, while blue-bolded segments represent suboptimal results.}
\label{self-pre-trained_weights_CUB200}
\end{table*}

\subsection{Visualization}
\begin{figure}[ht]
\centering
\begin{subfigure}{0.22\textwidth}
    \includegraphics[width=\textwidth]{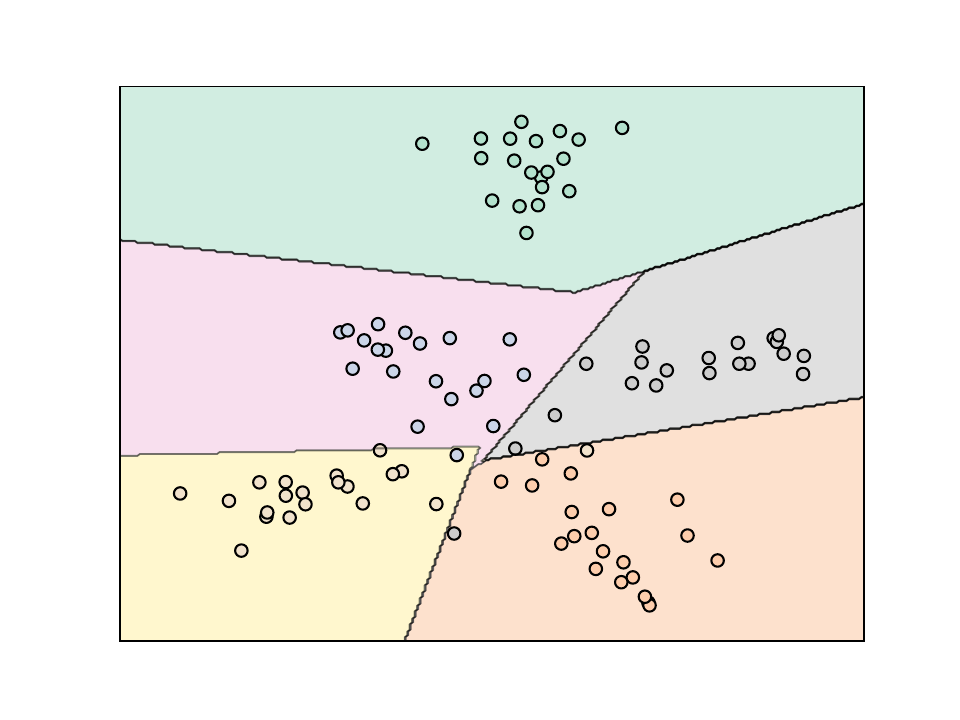}
    \caption{\scriptsize Five classes in base session}
    \label{fig:old_tsne}
\end{subfigure}
\hfill
\begin{subfigure}{0.22\textwidth}
    \includegraphics[width=\textwidth]{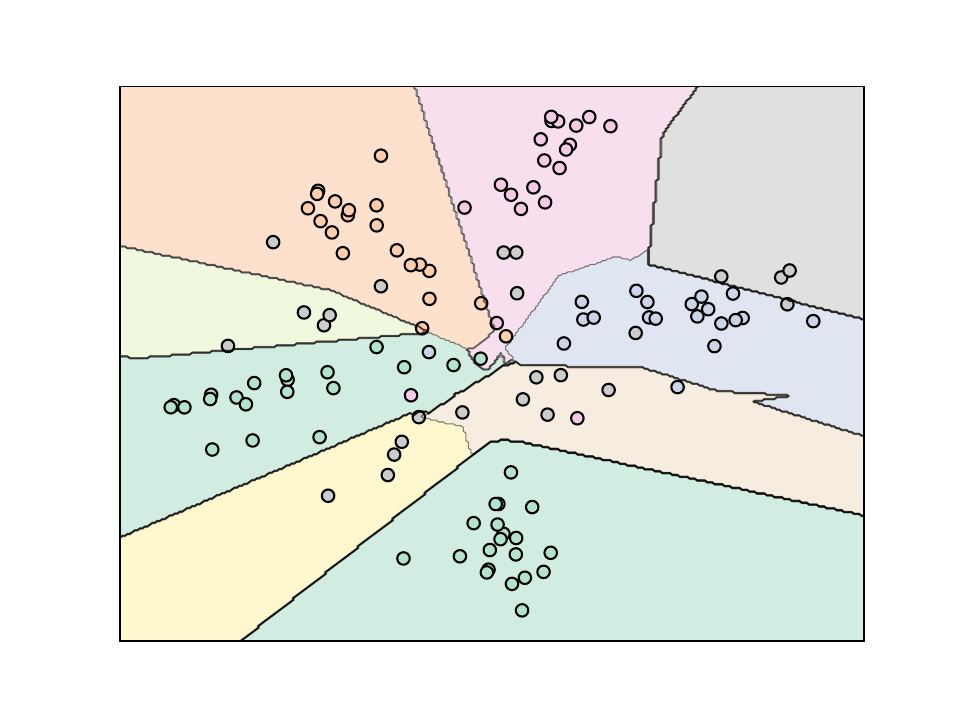}
    \caption{\scriptsize Nine classes in both sessions}
    \label{fig:new_tsne}
\end{subfigure}
\caption{t-SNE visualization of embeddings from randomly sampled instances in the base session and incremental sessions.}
\label{fig:embeddings_tsne}
\end{figure}

To compellingly showcase the advantages of our NTK-FSCIL methodology, we employ t-SNE visualization to depict the embeddings generated by our optimized model, as illustrated in \cref{fig:embeddings_tsne}. Initially, we select five categories from the base session, with each category comprising 20 samples, as displayed in \cref{fig:embeddings_tsne}\subref{fig:old_tsne}. Following this, we enrich this dataset by adding four additional categories from subsequent incremental sessions, with each new category contributing 5 samples. This augmentation leads to a more detailed and insightful t-SNE visualization, as exhibited in \cref{fig:embeddings_tsne}\subref{fig:new_tsne}, which offers a clearer perspective on the optimized embedding space.

As demonstrated in \cref{fig:embeddings_tsne}, our NTK-FSCIL method distinctively segregates samples from both base and incremental sessions, showcasing enhanced separability and a clear testament to the model's generalization capabilities. This visualization not only highlights the effectiveness of our approach in creating distinct representations for different categories but also underscores its ability to adapt and maintain performance across incremental learning phases.

\section{Conclusion}
Our research applies the Neural Tangents Kernel (NTK) theory to the FSCIL paradigm, establishing a robust correlation between model generalization capabilities and NTK convergence and NTK-related generalization loss. Through both theoretical and empirical analyses, we validate the significance of NTK in FSCIL generalization. Our NTK-FSCIL approach, emphasizing NTK matrix optimization, attains exceptional results on well-established FSCIL benchmarks.

{
    \small
    \bibliographystyle{IEEEtran}
    \bibliography{main}

\begin{thebibliography}{10}
\providecommand{\url}[1]{#1}
\csname url@samestyle\endcsname
\providecommand{\newblock}{\relax}
\providecommand{\bibinfo}[2]{#2}
\providecommand{\BIBentrySTDinterwordspacing}{\spaceskip=0pt\relax}
\providecommand{\BIBentryALTinterwordstretchfactor}{4}
\providecommand{\BIBentryALTinterwordspacing}{\spaceskip=\fontdimen2\font plus
\BIBentryALTinterwordstretchfactor\fontdimen3\font minus \fontdimen4\font\relax}
\providecommand{\BIBforeignlanguage}[2]{{%
\expandafter\ifx\csname l@#1\endcsname\relax
\typeout{** WARNING: IEEEtran.bst: No hyphenation pattern has been}%
\typeout{** loaded for the language `#1'. Using the pattern for}%
\typeout{** the default language instead.}%
\else
\language=\csname l@#1\endcsname
\fi
#2}}
\providecommand{\BIBdecl}{\relax}
\BIBdecl

\bibitem{rebuffi2017icarl}
S.-A. Rebuffi, A.~Kolesnikov, G.~Sperl, and C.~H. Lampert, ``icarl: Incremental classifier and representation learning,'' in \emph{Proceedings of the IEEE/CVF Conference on Computer Vision and Pattern Recognition}, 2017, pp. 2001--2010.

\bibitem{yan2021dynamically}
S.~Yan, J.~Xie, and X.~He, ``Der: Dynamically expandable representation for class incremental learning,'' in \emph{Proceedings of the IEEE/CVF Conference on Computer Vision and Pattern Recognition}, 2021, pp. 3014--3023.

\bibitem{douillard2022dytox}
A.~Douillard, A.~Ram{\'e}, G.~Couairon, and M.~Cord, ``Dytox: Transformers for continual learning with dynamic token expansion,'' in \emph{Proceedings of the IEEE/CVF Conference on Computer Vision and Pattern Recognition}, 2022, pp. 9285--9295.

\bibitem{zhang2021few}
C.~Zhang, N.~Song, G.~Lin, Y.~Zheng, P.~Pan, and Y.~Xu, ``Few-shot incremental learning with continually evolved classifiers,'' in \emph{Proceedings of the IEEE/CVF Conference on Computer Vision and Pattern Recognition}, 2021, pp. 12\,455--12\,464.

\bibitem{zhou2022few}
D.-W. Zhou, H.-J. Ye, L.~Ma, D.~Xie, S.~Pu, and D.-C. Zhan, ``Few-shot class-incremental learning by sampling multi-phase tasks,'' \emph{IEEE Transactions on Pattern Analysis and Machine Intelligence}, 2022.

\bibitem{yoon2023soft}
J.~Yoon, S.~Madjid, S.~J. Hwang, C.-D. Yoo \emph{et~al.}, ``On the soft-subnetwork for few-shot class incremental learning,'' in \emph{International Conference on Learning Representations}.\hskip 1em plus 0.5em minus 0.4em\relax International Conference on Learning Representations, 2023.

\bibitem{yang2023neural}
Y.~Yang, H.~Yuan, X.~Li, Z.~Lin, P.~Torr, and D.~Tao, ``Neural collapse inspired feature-classifier alignment for few-shot class-incremental learning,'' in \emph{International Conference on Learning Representations}, 2023.

\bibitem{novak2022fast}
R.~Novak, J.~Sohl-Dickstein, and S.~S. Schoenholz, ``Fast finite width neural tangent kernel,'' in \emph{International Conference on Machine Learning}.\hskip 1em plus 0.5em minus 0.4em\relax PMLR, 2022, pp. 17\,018--17\,044.

\bibitem{Cagnetta2022WhatCB}
F.~Cagnetta, A.~Favero, and M.~Wyart, ``What can be learnt with wide convolutional neural networks?'' in \emph{International Conference on Machine Learning}.\hskip 1em plus 0.5em minus 0.4em\relax PMLR, 2022.

\bibitem{jacot2018neural}
A.~Jacot, F.~Gabriel, and C.~Hongler, ``Neural tangent kernel: Convergence and generalization in neural networks,'' \emph{Advances in Neural Information Processing Systems}, vol.~31, 2018.

\bibitem{wang2022global}
H.~Wang, Y.~Wang, R.~Sun, and B.~Li, ``Global convergence of maml and theory-inspired neural architecture search for few-shot learning,'' in \emph{Proceedings of the IEEE/CVF Conference on Computer Vision and Pattern Recognition}, 2022, pp. 9797--9808.

\bibitem{caron2023over}
F.~Caron, F.~Ayed, P.~Jung, H.~Lee, J.~Lee, and H.~Yang, ``Over-parameterised shallow neural networks with asymmetrical node scaling: Global convergence guarantees and feature learning,'' \emph{arXiv preprint arXiv:2302.01002}, 2023.

\bibitem{du2018gradient}
S.~S. Du, X.~Zhai, B.~Poczos, and A.~Singh, ``Gradient descent provably optimizes over-parameterized neural networks,'' \emph{arXiv preprint arXiv:1810.02054}, 2018.

\bibitem{zou2018stochastic}
D.~Zou, Y.~Cao, D.~Zhou, and Q.~Gu, ``Stochastic gradient descent optimizes over-parameterized deep relu networks,'' \emph{arXiv preprint arXiv:1811.08888}, 2018.

\bibitem{allen2019convergence}
Z.~Allen-Zhu, Y.~Li, and Z.~Song, ``A convergence theory for deep learning via over-parameterization,'' in \emph{International Conference on Machine Learning}.\hskip 1em plus 0.5em minus 0.4em\relax PMLR, 2019, pp. 242--252.

\bibitem{wei2022more}
A.~Wei, W.~Hu, and J.~Steinhardt, ``More than a toy: Random matrix models predict how real-world neural representations generalize,'' in \emph{International Conference on Machine Learning}.\hskip 1em plus 0.5em minus 0.4em\relax PMLR, 2022, pp. 23\,549--23\,588.

\bibitem{richards2021stability}
D.~Richards and I.~Kuzborskij, ``Stability \& generalisation of gradient descent for shallow neural networks without the neural tangent kernel,'' in \emph{Advances in Neural Information Processing Systems}, A.~Beygelzimer, Y.~Dauphin, P.~Liang, and J.~W. Vaughan, Eds., 2021.

\bibitem{bombari2023stability}
S.~Bombari and M.~Mondelli, ``Stability, generalization and privacy: Precise analysis for random and ntk features,'' \emph{arXiv preprint arXiv:2305.12100}, 2023.

\bibitem{taheri2023generalization}
H.~Taheri and C.~Thrampoulidis, ``Generalization and stability of interpolating neural networks with minimal width,'' \emph{arXiv preprint arXiv:2302.09235}, 2023.

\bibitem{neal1996priors}
R.~M. Neal and R.~M. Neal, ``Priors for infinite networks,'' \emph{Bayesian Learning for Neural Networks}, pp. 29--53, 1996.

\bibitem{le2007continuous}
N.~Le~Roux and Y.~Bengio, ``Continuous neural networks,'' in \emph{Artificial Intelligence and Statistics}.\hskip 1em plus 0.5em minus 0.4em\relax PMLR, 2007, pp. 404--411.

\bibitem{hazan2015steps}
T.~Hazan and T.~Jaakkola, ``Steps toward deep kernel methods from infinite neural networks,'' \emph{arXiv preprint arXiv:1508.05133}, 2015.

\bibitem{du2019width}
S.~Du and W.~Hu, ``Width provably matters in optimization for deep linear neural networks,'' in \emph{International Conference on Machine Learning}.\hskip 1em plus 0.5em minus 0.4em\relax PMLR, 2019, pp. 1655--1664.

\bibitem{lee2020finite}
J.~Lee, S.~Schoenholz, J.~Pennington, B.~Adlam, L.~Xiao, R.~Novak, and J.~Sohl-Dickstein, ``Finite versus infinite neural networks: an empirical study,'' \emph{Advances in Neural Information Processing Systems}, vol.~33, pp. 15\,156--15\,172, 2020.

\bibitem{xiang2022tkil}
J.~Xiang and E.~Shlizerman, ``Tkil: tangent kernel approach for class balanced incremental learning,'' \emph{arXiv preprint arXiv:2206.08492}, 2022.

\bibitem{tao2020few}
X.~Tao, X.~Hong, X.~Chang, S.~Dong, X.~Wei, and Y.~Gong, ``Few-shot class-incremental learning,'' in \emph{Proceedings of the IEEE/CVF Conference on Computer Vision and Pattern Recognition}, 2020, pp. 12\,183--12\,192.

\bibitem{shi2021overcoming}
G.~Shi, J.~Chen, W.~Zhang, L.-M. Zhan, and X.-M. Wu, ``Overcoming catastrophic forgetting in incremental few-shot learning by finding flat minima,'' \emph{Advances in Neural Information Processing Systems}, vol.~34, pp. 6747--6761, 2021.

\bibitem{hersche2022constrained}
M.~Hersche, G.~Karunaratne, G.~Cherubini, L.~Benini, A.~Sebastian, and A.~Rahimi, ``Constrained few-shot class-incremental learning,'' in \emph{Proceedings of the IEEE/CVF Conference on Computer Vision and Pattern Recognition}, 2022, pp. 9057--9067.

\bibitem{ji2023memorizing}
Z.~Ji, Z.~Hou, X.~Liu, Y.~Pang, and X.~Li, ``Memorizing complementation network for few-shot class-incremental learning,'' in \emph{IEEE Transactions on Image Processing}, 2023, pp. 937--948.

\bibitem{zhou2022forward}
D.-W. Zhou, F.-Y. Wang, H.-J. Ye, L.~Ma, S.~Pu, and D.-C. Zhan, ``Forward compatible few-shot class-incremental learning,'' in \emph{Proceedings of the IEEE/CVF Conference on Computer Vision and Pattern Recognition}, 2022, pp. 9046--9056.

\bibitem{peng2022few}
C.~Peng, K.~Zhao, T.~Wang, M.~Li, and B.~C. Lovell, ``Few-shot class-incremental learning from an open-set perspective,'' in \emph{European Conference on Computer Vision}.\hskip 1em plus 0.5em minus 0.4em\relax Springer, 2022, pp. 382--397.

\bibitem{lee2019wide}
J.~Lee, L.~Xiao, S.~Schoenholz, Y.~Bahri, R.~Novak, J.~Sohl-Dickstein, and J.~Pennington, ``Wide neural networks of any depth evolve as linear models under gradient descent,'' \emph{Advances in Neural Information Processing Systems}, vol.~32, 2019.

\bibitem{yang2020tensor}
G.~Yang, ``Tensor programs ii: Neural tangent kernel for any architecture,'' \emph{arXiv preprint arXiv:2006.14548}, 2020.

\bibitem{arora2019fine}
S.~Arora, S.~Du, W.~Hu, Z.~Li, and R.~Wang, ``Fine-grained analysis of optimization and generalization for overparameterized two-layer neural networks,'' in \emph{International Conference on Machine Learning}.\hskip 1em plus 0.5em minus 0.4em\relax PMLR, 2019, pp. 322--332.

\bibitem{chizat2019lazy}
L.~Chizat, E.~Oyallon, and F.~Bach, ``On lazy training in differentiable programming,'' \emph{Advances in neural information processing systems}, vol.~32, 2019.

\bibitem{daniely2017sgd}
A.~Daniely, ``Sgd learns the conjugate kernel class of the network,'' \emph{Advances in neural information processing systems}, vol.~30, 2017.

\bibitem{neyshabur2018towards}
B.~Neyshabur, Z.~Li, S.~Bhojanapalli, Y.~LeCun, and N.~Srebro, ``Towards understanding the role of over-parametrization in generalization of neural networks,'' \emph{arXiv preprint arXiv:1805.12076}, 2018.

\bibitem{baratin2021implicit}
A.~Baratin, T.~George, C.~Laurent, R.~D. Hjelm, G.~Lajoie, P.~Vincent, and S.~Lacoste-Julien, ``Implicit regularization via neural feature alignment,'' in \emph{International Conference on Artificial Intelligence and Statistics}.\hskip 1em plus 0.5em minus 0.4em\relax PMLR, 2021, pp. 2269--2277.

\bibitem{li2019exponential}
Z.~Li and S.~Arora, ``An exponential learning rate schedule for deep learning,'' \emph{arXiv preprint arXiv:1910.07454}, 2019.

\bibitem{novak2018bayesian}
R.~Novak, L.~Xiao, J.~Lee, Y.~Bahri, G.~Yang, J.~Hron, D.~A. Abolafia, J.~Pennington, and J.~Sohl-Dickstein, ``Bayesian deep convolutional networks with many channels are gaussian processes,'' \emph{arXiv preprint arXiv:1810.05148}, 2018.

\bibitem{fort2019deep}
S.~Fort, H.~Hu, and B.~Lakshminarayanan, ``Deep ensembles: A loss landscape perspective,'' \emph{arXiv preprint arXiv:1912.02757}, 2019.

\bibitem{canatar2021spectral}
A.~Canatar, B.~Bordelon, and C.~Pehlevan, ``Spectral bias and task-model alignment explain generalization in kernel regression and infinitely wide neural networks,'' \emph{Nature communications}, vol.~12, no.~1, p. 2914, 2021.

\bibitem{doan2021theoretical}
T.~Doan, M.~A. Bennani, B.~Mazoure, G.~Rabusseau, and P.~Alquier, ``A theoretical analysis of catastrophic forgetting through the ntk overlap matrix,'' in \emph{International Conference on Artificial Intelligence and Statistics}.\hskip 1em plus 0.5em minus 0.4em\relax PMLR, 2021, pp. 1072--1080.

\bibitem{chai2009generalization}
K.~Chai, ``Generalization errors and learning curves for regression with multi-task gaussian processes,'' \emph{Advances in Neural Information Processing Systems}, vol.~22, 2009.

\bibitem{bennani2020generalisation}
M.~A. Bennani, T.~Doan, and M.~Sugiyama, ``Generalisation guarantees for continual learning with orthogonal gradient descent,'' \emph{arXiv preprint arXiv:2006.11942}, 2020.

\bibitem{nguyen2021tight}
Q.~Nguyen, M.~Mondelli, and G.~F. Montufar, ``Tight bounds on the smallest eigenvalue of the neural tangent kernel for deep relu networks,'' in \emph{International Conference on Machine Learning}.\hskip 1em plus 0.5em minus 0.4em\relax PMLR, 2021, pp. 8119--8129.

\bibitem{murray2022characterizing}
M.~Murray, H.~Jin, B.~Bowman, and G.~Montufar, ``Characterizing the spectrum of the ntk via a power series expansion,'' \emph{arXiv preprint arXiv:2211.07844}, 2022.

\bibitem{chi2022metafscil}
Z.~Chi, L.~Gu, H.~Liu, Y.~Wang, Y.~Yu, and J.~Tang, ``Metafscil: a meta-learning approach for few-shot class incremental learning,'' in \emph{Proceedings of the IEEE/CVF Conference on Computer Vision and Pattern Recognition}, 2022, pp. 14\,166--14\,175.

\bibitem{wei2019regularization}
C.~Wei, J.~D. Lee, Q.~Liu, and T.~Ma, ``Regularization matters: Generalization and optimization of neural nets vs their induced kernel,'' \emph{Advances in Neural Information Processing Systems}, vol.~32, 2019.

\bibitem{liu2023few}
S.~Liu and X.~Wang, ``Few-shot dataset distillation via translative pre-training,'' in \emph{Proceedings of the IEEE/CVF International Conference on Computer Vision}, 2023, pp. 18\,654--18\,664.

\bibitem{caron2020unsupervised}
M.~Caron, I.~Misra, J.~Mairal, P.~Goyal, P.~Bojanowski, and A.~Joulin, ``Unsupervised learning of visual features by contrasting cluster assignments,'' \emph{Advances in Neural Information Processing Systems}, vol.~33, pp. 9912--9924, 2020.

\bibitem{caron2021emerging}
M.~Caron, H.~Touvron, I.~Misra, H.~J{\'e}gou, J.~Mairal, P.~Bojanowski, and A.~Joulin, ``Emerging properties in self-supervised vision transformers,'' in \emph{Proceedings of the IEEE/CVF International Conference on Computer Vision}, 2021, pp. 9650--9660.

\bibitem{chen2020simple}
T.~Chen, S.~Kornblith, M.~Norouzi, and G.~Hinton, ``A simple framework for contrastive learning of visual representations,'' in \emph{International Conference on Machine Learning}.\hskip 1em plus 0.5em minus 0.4em\relax PMLR, 2020, pp. 1597--1607.

\bibitem{chen2021exploring}
X.~Chen and K.~He, ``Exploring simple siamese representation learning,'' in \emph{Proceedings of the IEEE/CVF Conference on Computer Vision and Pattern Recognition}, 2021, pp. 15\,750--15\,758.

\bibitem{he2022masked}
K.~He, X.~Chen, S.~Xie, Y.~Li, P.~Doll{\'a}r, and R.~Girshick, ``Masked autoencoders are scalable vision learners,'' in \emph{Proceedings of the IEEE/CVF Conference on Computer Vision and Pattern Recognition}, 2022, pp. 16\,000--16\,009.

\bibitem{chen2021mocov3}
X.~Chen*, S.~Xie*, and K.~He, ``An empirical study of training self-supervised vision transformers,'' \emph{arXiv preprint arXiv:2104.02057}, 2021.

\bibitem{grill2020bootstrap}
J.-B. Grill, F.~Strub, F.~Altch{\'e}, C.~Tallec, P.~Richemond, E.~Buchatskaya, C.~Doersch, B.~Avila~Pires, Z.~Guo, M.~Gheshlaghi~Azar \emph{et~al.}, ``Bootstrap your own latent-a new approach to self-supervised learning,'' \emph{Advances in Neural Information Processing Systems}, vol.~33, pp. 21\,271--21\,284, 2020.

\bibitem{tian2023designing}
\BIBentryALTinterwordspacing
K.~Tian, Y.~Jiang, qishuai diao, C.~Lin, L.~Wang, and Z.~Yuan, ``Designing {BERT} for convolutional networks: Sparse and hierarchical masked modeling,'' in \emph{International Conference on Learning Representations}, 2023. [Online]. Available: \url{https://openreview.net/forum?id=NRxydtWup1S}
\BIBentrySTDinterwordspacing

\bibitem{zou2022marginbased}
Y.~Zou, S.~Zhang, Y.~Li, and R.~Li, ``Margin-based few-shot class-incremental learning with class-level overfitting mitigation,'' in \emph{Advances in Neural Information Processing Systems}, A.~H. Oh, A.~Agarwal, D.~Belgrave, and K.~Cho, Eds., 2022.

\bibitem{huang2020curricularface}
Y.~Huang, Y.~Wang, Y.~Tai, X.~Liu, P.~Shen, S.~Li, J.~Li, and F.~Huang, ``Curricularface: adaptive curriculum learning loss for deep face recognition,'' in \emph{Proceedings of the IEEE/CVF Conference on Computer Vision and Pattern Recognition}, 2020, pp. 5901--5910.

\bibitem{krizhevsky2009learning}
A.~Krizhevsky, G.~Hinton \emph{et~al.}, ``Learning multiple layers of features from tiny images,'' 2009.

\bibitem{wah2011caltech}
C.~Wah, S.~Branson, P.~Welinder, P.~Perona, and S.~Belongie, ``The caltech-ucsd birds-200-2011 dataset,'' 2011.

\bibitem{ILSVRC15}
O.~Russakovsky, J.~Deng, H.~Su, J.~Krause, S.~Satheesh, S.~Ma, Z.~Huang, A.~Karpathy, A.~Khosla, M.~Bernstein, A.~C. Berg, and L.~Fei-Fei, ``{ImageNet Large Scale Visual Recognition Challenge},'' \emph{International Journal of Computer Vision (IJCV)}, vol. 115, no.~3, pp. 211--252, 2015.

\bibitem{wu2019large}
Y.~Wu, Y.~Chen, L.~Wang, Y.~Ye, Z.~Liu, Y.~Guo, and Y.~Fu, ``Large scale incremental learning,'' in \emph{Proceedings of the IEEE/CVF Conference on Computer Vision and Pattern Recognition}, 2019, pp. 374--382.

\bibitem{deng2019arcface}
J.~Deng, J.~Guo, N.~Xue, and S.~Zafeiriou, ``Arcface: Additive angular margin loss for deep face recognition,'' in \emph{Proceedings of the IEEE/CVF Conference on Computer Vision and Pattern Recognition}, 2019, pp. 4690--4699.

\bibitem{wang2018cosface}
H.~Wang, Y.~Wang, Z.~Zhou, X.~Ji, D.~Gong, J.~Zhou, Z.~Li, and W.~Liu, ``Cosface: Large margin cosine loss for deep face recognition,'' in \emph{Proceedings of the IEEE/CVF Conference on Computer Vision and Pattern Recognition}, 2018, pp. 5265--5274.

\bibitem{zhang2018mixup}
\BIBentryALTinterwordspacing
H.~Zhang, M.~Cisse, Y.~N. Dauphin, and D.~Lopez-Paz, ``mixup: Beyond empirical risk minimization,'' in \emph{International Conference on Learning Representations}, 2018. [Online]. Available: \url{https://openreview.net/forum?id=r1Ddp1-Rb}
\BIBentrySTDinterwordspacing

\bibitem{yun2019cutmix}
S.~Yun, D.~Han, S.~J. Oh, S.~Chun, J.~Choe, and Y.~Yoo, ``Cutmix: Regularization strategy to train strong classifiers with localizable features,'' in \emph{Proceedings of the IEEE/CVF International Conference on Computer Vision}, 2019, pp. 6023--6032.

\bibitem{2020augmix}
\BIBentryALTinterwordspacing
D.~Hendrycks*, N.~Mu*, E.~D. Cubuk, B.~Zoph, J.~Gilmer, and B.~Lakshminarayanan, ``Augmix: A simple method to improve robustness and uncertainty under data shift,'' in \emph{International Conference on Learning Representations}, 2020. [Online]. Available: \url{https://openreview.net/forum?id=S1gmrxHFvB}
\BIBentrySTDinterwordspacing

\bibitem{kim2020puzzle}
J.-H. Kim, W.~Choo, and H.~O. Song, ``Puzzle mix: Exploiting saliency and local statistics for optimal mixup,'' in \emph{International Conference on Machine Learning}.\hskip 1em plus 0.5em minus 0.4em\relax PMLR, 2020, pp. 5275--5285.

\bibitem{zhuang2023gkeal}
H.~Zhuang, Z.~Weng, R.~He, Z.~Lin, and Z.~Zeng, ``Gkeal: Gaussian kernel embedded analytic learning for few-shot class incremental task,'' in \emph{Proceedings of the IEEE/CVF Conference on Computer Vision and Pattern Recognition}, 2023, pp. 7746--7755.

\bibitem{kim2023warping}
\BIBentryALTinterwordspacing
D.-Y. Kim, D.-J. Han, J.~Seo, and J.~Moon, ``Warping the space: Weight space rotation for class-incremental few-shot learning,'' in \emph{International Conference on Learning Representations}, 2023. [Online]. Available: \url{https://openreview.net/forum?id=kPLzOfPfA2l}
\BIBentrySTDinterwordspacing

\bibitem{liu2021swin}
Z.~Liu, Y.~Lin, Y.~Cao, H.~Hu, Y.~Wei, Z.~Zhang, S.~Lin, and B.~Guo, ``Swin transformer: Hierarchical vision transformer using shifted windows,'' in \emph{Proceedings of the IEEE/CVF international conference on computer vision}, 2021, pp. 10\,012--10\,022.

\bibitem{liu2022convnet}
Z.~Liu, H.~Mao, C.-Y. Wu, C.~Feichtenhofer, T.~Darrell, and S.~Xie, ``A convnet for the 2020s,'' in \emph{Proceedings of the IEEE/CVF conference on computer vision and pattern recognition}, 2022, pp. 11\,976--11\,986.

\bibitem{sun2023regularizing}
Z.~Sun, Y.~Mu, and G.~Hua, ``Regularizing second-order influences for continual learning,'' in \emph{Proceedings of the IEEE/CVF Conference on Computer Vision and Pattern Recognition}, 2023, pp. 20\,166--20\,175.

\end{thebibliography}
}
\vspace{-2ex}
\begin{IEEEbiography}[{\includegraphics[width=1in,height=1.25in,clip,keepaspectratio]{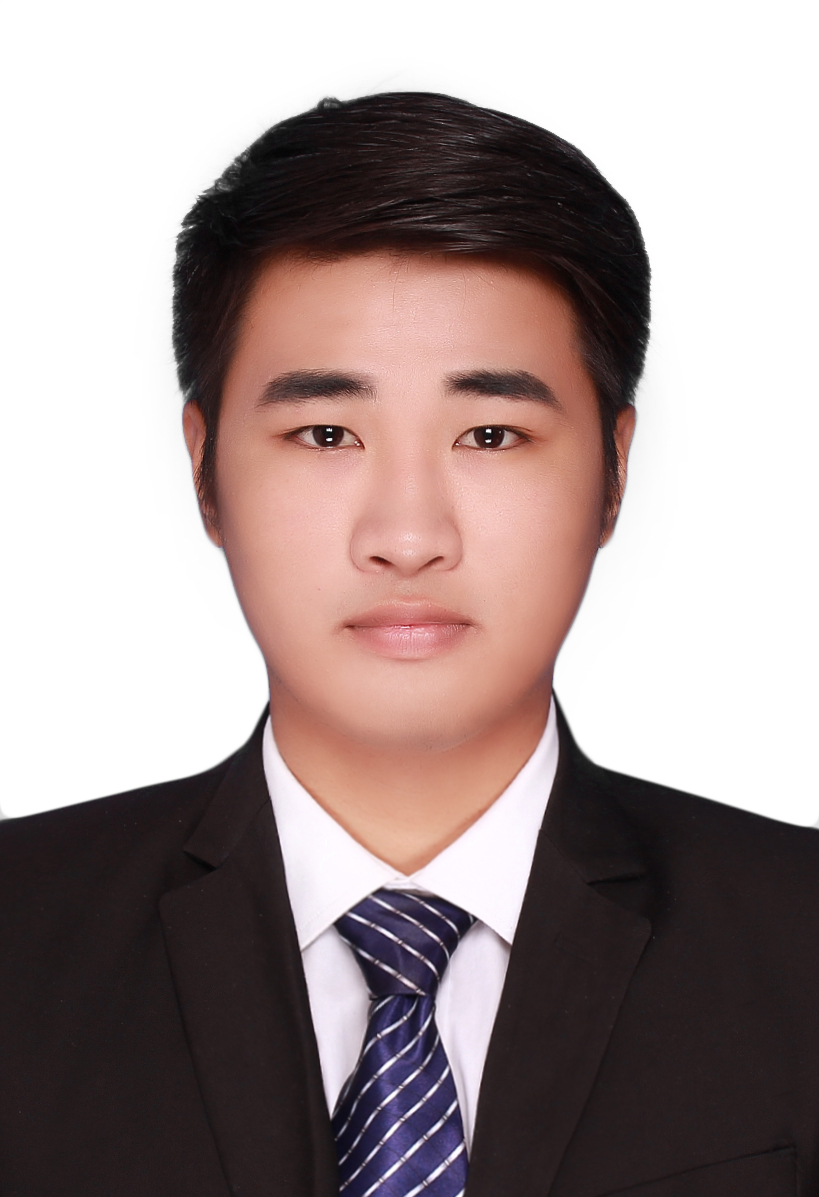}}]{Jingren Liu} received the B.S. degree in Computer Science and Technology from Nanjing University of Finance and Economy, Nanjing, China, in 2019, and is currently working toward the PhD degree in the School of Electrical and Information Engineering, Tianjin University, Tianjin, China. His current research interests include continual learning, few shot leanring, and prompt learning.
\end{IEEEbiography}

\vspace{-2ex}
\begin{IEEEbiography}[{\includegraphics[width=1in,height=1.25in,clip,keepaspectratio]{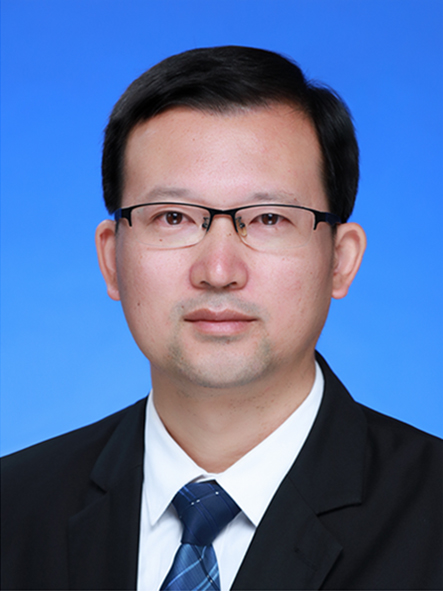}}]{Zhong Ji} received the Ph.D. degree in signal and information processing from Tianjin University, Tianjin, China, in 2008. He is currently a Professor with the School of Electrical and Information Engineering, Tianjin University. He has authored over 100 technical articles in refereed journals and proceedings. His current research interests include continual learning, few shot leanring, and cross-modal analysis.
\end{IEEEbiography}

\vspace{-2ex}
\begin{IEEEbiography}[{\includegraphics[width=1in,height=1.25in,clip,keepaspectratio]{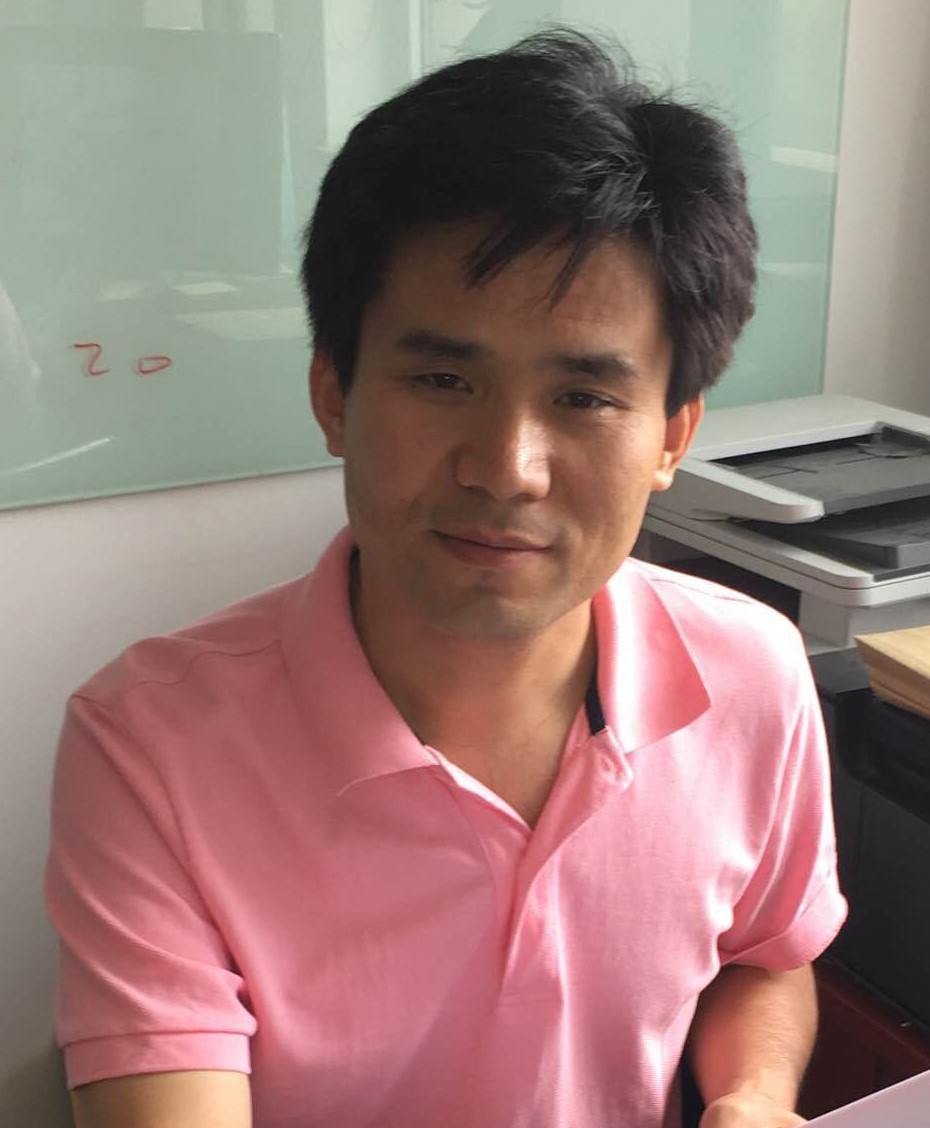}}]{YanWei Pang} received the Ph.D. degree in electronic engineering from the University of Science and Technology of China, Hefei, China, in 2004. He is currently a Professor with the School of Electrical and Information Engineering, Tianjin University, Tianjin, China. He has authored over 200 scientific papers. His current research interests include object detection and recognition, vision in bad weather, and computer vision.
\end{IEEEbiography}

\vspace{-2ex}
\begin{IEEEbiography}[{\includegraphics[width=1in,height=1.25in,clip,keepaspectratio]{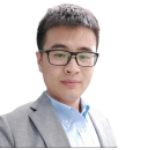}}]{YunLong Yu} received the Ph.D. degree in information and communication engineering from Tianjin University, Tianjin, China, in 2019. He is currently a Distinguished Associate Researcher with the College of Information Science and Electronic Engineering, Zhejiang University, Hangzhou, China. His current research interests include machine learning and computer vision.
\end{IEEEbiography}
\vfill
\end{document}